%% file: main.tex
\documentclass[10pt,journal,compsoc]{IEEEtran}
\ifCLASSOPTIONcompsoc
  \usepackage[nocompress]{cite}
\else
  \usepackage{cite}
\fi
\ifCLASSINFOpdf
   \usepackage[pdftex]{graphicx}
   \DeclareGraphicsExtensions{.pdf,.jpeg,.png}
\else
\fi
\usepackage{amsmath}
\usepackage{array}

\ifCLASSOPTIONcompsoc
 \usepackage[caption=false,font=footnotesize,labelfont=sf,textfont=sf]{subfig}
\else
 \usepackage[caption=false,font=footnotesize]{subfig}
\fi
\usepackage{url}

\usepackage[export]{adjustbox}
\usepackage{booktabs}
\usepackage{amssymb}
\usepackage{mathtools}
\usepackage{multirow}
\usepackage{enumitem}
\usepackage{tikz}
\usepackage{siunitx}
\usetikzlibrary{positioning,shapes,arrows,fit}

\DeclareMathOperator{\Beta}{Beta}
\DeclareMathOperator{\Bernoulli}{Bernoulli}

\DeclareMathOperator{\MI}{MI}
\DeclareMathOperator{\entropy}{H}

\DeclareMathOperator*{\argmax}{arg\,max}

\newcolumntype{C}[1]{>{\centering\let\newline\\\arraybackslash\hspace{0pt}}m{#1}}

\allowdisplaybreaks

\definecolor{color_sync}{rgb}{0.95,1,0.9}
\definecolor{color_feat}{rgb}{0.9,0.95,1}

\begin{document}
%
\title{Compositional Scene Representation Learning via Reconstruction: A Survey}
%
%
%
%

\author{Jinyang Yuan,
        Tonglin Chen,
        Bin Li,
        and Xiangyang Xue
\IEEEcompsocitemizethanks{\IEEEcompsocthanksitem The authors are with the Shanghai Key Laboratory of Intelligent Information Processing
and the School of Computer Science, Fudan University, Shanghai
200433, China.\protect\\
E-mail: \{yuanjinyang, tlchen18, libin, xyxue\}@fudan.edu.cn}
\thanks{Manuscript received Month Day, Year; revised Month Day, Year.\\
(Corresponding author: Bin Li.)}}

%
%

\markboth{Journal of \LaTeX\ Class Files,~Vol.~14, No.~8, August~2015}%
{Shell \MakeLowercase{\textit{et al.}}: Bare Demo of IEEEtran.cls for Computer Society Journals}
%


\IEEEpubid{\begin{minipage}{\textwidth}\ \\[12pt]
	\copyright~2023 IEEE. Personal use of this material is permitted. Permission from IEEE must be obtained for all other uses, including reprinting/republishing this material for advertising or promotional purposes, collecting new collected works for resale or redistribution to servers or lists, or reuse of any copyrighted component of this work in other works. This work has been submitted to the IEEE for possible publication. Copyright may be transferred without notice, after which this version may no longer be accessible.
\end{minipage}}


\IEEEtitleabstractindextext{%
\begin{abstract}
Visual scenes are composed of visual concepts and have the property of combinatorial explosion. An important reason for humans to efficiently learn from diverse visual scenes is the ability of compositional perception, and it is desirable for artificial intelligence to have similar abilities. Compositional scene representation learning is a task that enables such abilities. In recent years, various methods have been proposed to apply deep neural networks, which have been proven to be advantageous in representation learning, to learn compositional scene representations via reconstruction, advancing this research direction into the deep learning era. Learning via reconstruction is advantageous because it may utilize massive unlabeled data and avoid costly and laborious data annotation. In this survey, we first outline the current progress on reconstruction-based compositional scene representation learning with deep neural networks, including development history and categorizations of existing methods from the perspectives of the modeling of visual scenes and the inference of scene representations; then provide benchmarks, including an open source toolbox to reproduce the benchmark experiments, of representative methods that consider the most extensively studied problem setting and form the foundation for other methods; and finally discuss the limitations of existing methods and future directions of this research topic.
\end{abstract}

}

\maketitle

\IEEEdisplaynontitleabstractindextext

%
\IEEEpeerreviewmaketitle

\IEEEraisesectionheading{\section{Introduction}\label{sec:introduction}}

%
%
%
%


\IEEEPARstart{V}{isual} scene representation learning is an important research problem in computer vision. If more suitable representations can be learned, the performance of artificial intelligence systems on computer vision tasks can be improved \cite{Ciresan2011committee,Krizhevsky2012ImageNet,He2016Deep}. Visual scenes are composed of visual concepts (objects or backgrounds or their basic parts), and the combination of visual concepts has the property of combinatorial explosion. Even with only a few types of objects, infinite visual scenes with rich diversity can be created. Therefore, for complex visual scene understanding tasks such as visual question answering (VQA), learning a single representation for the entire scene is not advantageous as all visual information is entangled in this highly complex representation, making it hard to correctly extract information such as relations between objects \cite{Santoro2017simple}. Humans can efficiently learn from visual signals and effectively understand visual scenes. One important ingredient of this remarkable ability is to perceive the world in a \emph{compositional} way \cite{Lake2017Building}. To mimic human behavior and learn representations more suitable for visual scene understanding, additional inductive bias can be added by explicitly considering the compositionality of visual scenes and extracting \emph{compositional scene representations} that decompose visual scenes into several regions (different regions correspond to different visual concepts) and represent each region separately. In this way, the diversity caused by the combination of visual concepts could be better dealt with, and performing visual scene understanding based on the learned representations could be simpler and better understood by humans because the concepts of objects have already been abstracted in a \emph{neural-symbolic} way \cite{Yi2018Neural,Mao2019Neuro}.

The compositionality of visual scenes has been studied for a long time in the fields of computer vision and artificial intelligence. In the pioneering work of Geman et al. \cite{Geman2002Composition}, compositionality has been mathematically formulated according to Rissanen's Minimum Description Length (MDL) principle \cite{Rissanen1998Stochastic}, and this formulation has been successfully applied to recognize online uppercase characters based on hierarchical representations of objects in the object library \cite{Potter1999Compositional,Huang2001Compositional}. This work has verified that the correct decomposition of a visual scene into visual concepts leads to more \emph{compact} representations than alternative decompositions. The fact that compositionality can be learned via \emph{information bottlenecks} forms the cornerstone of reconstruction-based compositional scene representation learning. Besides the MDL principle, it has been shown that compositional scene representations can be learned in other ways, e.g., based on local inhibition \cite{Fidler2007Towards} or hierarchical clustering \cite{Zhu2008Unsupervised} in the unsupervised setting and based on maximum likelihood estimation \cite{Zhu2007Stochastic} when supervisions of parse graphs are available. With the development of this research direction, compositional scene representations have shown utility in real-world tasks, e.g., pattern recognition \cite{Kortylewski2016Probabilistic}, object classification \cite{Ommer2010Learning,Wu2010Learning}, object detection \cite{Ommer2010Learning,Wu2010Learning,Zhu2010Part}, and image parsing \cite{Zhu2008Unsupervised,Zhu2007Stochastic}.

The above-mentioned compositional scene representation learning methods, however, do not directly learn from RGB values of pixels. Instead, these methods operate on alternative formats like locations of sampled points \cite{Geman2002Composition} or rely on hand-crafted features such as outputs of Gabor filters \cite{Fidler2007Towards,Wu2010Learning,Kortylewski2016Probabilistic}, outputs of predefined spatial filters \cite{Zhu2010Part}, local edge and color histograms \cite{Ommer2010Learning}, oriented edge features \cite{Zhu2008Unsupervised}, and instances detected by algorithms like the Hough transform \cite{Zhu2007Stochastic}. With the advent of the era of big data and the rise of deep learning, it has been shown that given sufficient computing power and a large number of annotated images, features learned automatically from RGB values of pixels usually lead to better performance than hand-crafted features. Therefore, deep neural networks have become mainstream in computer vision. Given the excellent representation learning ability of deep neural networks, it is desirable to develop a ``deep'' version of compositional scene representation learning.

\setcounter{footnote}{-1}

\begin{figure}[t]
	\centering
	\includegraphics[width=0.97\columnwidth]{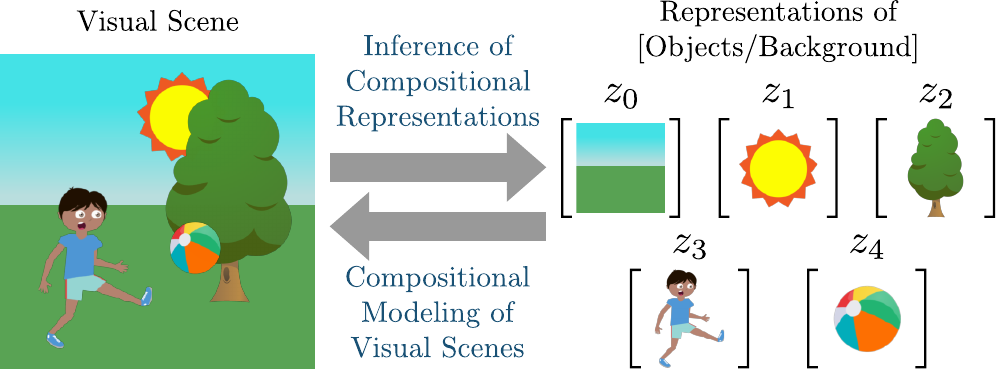}
	\caption{The general framework of learning compositional scene representations via reconstruction.\protect\footnotemark}
	\label{fig:intro}
\end{figure}

\footnotetext{This figure is modified based on Figure 4 in \cite{Yuan2021Knowledge}. The cartoon assets are from the Abstract Scene Dataset \cite{Zitnick2013Bringing}.}

When object-level supervisions such as bounding boxes, segmentation masks, and parse graphs are available, learning compositional scene representations based on deep neural networks could be straightforward. One viable way is to first learn object detection, image segmentation, or image parsing in the supervised setting and then learn separate representations for each bounding box or segmented region. This simple but effective scheme has been successfully applied to complex computer vision tasks such as visual question answering \cite{Yi2018Neural,Chen2021Grounding} and visual concept learning \cite{Mao2019Neuro,Han2019Visual}. However, manual labeling of images is expensive and laborious. Compared to all the accessible images, images annotated with bounding boxes or segmentation masks only occupy a small proportion. Therefore, it would be beneficial to find a way to apply deep neural networks to learn compositional scene representations without object-level annotations, such that the massive unlabeled images can be better utilized by learning in the weakly supervised, semi-supervised, or even unsupervised setting.

Autoencoding is a common approach to unsupervised learning of representations for the entire image using deep neural networks. Regularized by the explicitly defined regularization terms or the information bottlenecks provided by autoencoders, representations with desirable properties can be learned by minimizing reconstruction errors. The same idea can be applied\footnote{The experimental results in the Supplementary Material show that, compared with representing the entire scene with a single vector, compositional scene representations with approximately the same overall length usually lead to better reconstruction quality. This finding illustrates the superiority of compositional scene representations in terms of informativeness and verifies that compositional scene representations can be learned via reconstruction when information bottlenecks exist.} to compositional scene representation learning if combined with \emph{compositional modeling} of visual scenes, i.e., defining how to transform compositional scene representations into images of individual visual concepts and how to composite these images to form the entire scene.

This survey focuses on the problem of applying \emph{deep neural networks} to learn \emph{compositional representations} of visual scenes, with \emph{image reconstruction} as the main objective (\emph{not} using \emph{any} supervision or \emph{only} using \emph{scene-level} annotations like the viewpoints of visual scenes). Figure \ref{fig:intro} illustrates the general learning framework that consists of two parts, i.e., encoding (inference of compositional scene representations) and decoding (compositional modeling of visual scenes). The considered problem has gained increasing attention over the past few years, and various methods have been proposed. Depending on the datasets used in the experiments and the naming conventions, compositional scene representation learning is often referred to as \emph{perceptual grouping}, \emph{object-based representation learning}, \emph{object-centric learning}, or \emph{object-oriented learning}. These terms will be used interchangeably in this survey. Although the effectiveness of most existing methods has only been verified on synthetic visual scenes, the core components of these methods, i.e., \emph{compositional modeling and inference}, are not developed based on the assumption of synthetic scenes and can thus serve as the foundation for designing more advanced methods capable of learning from complex real-world visual scenes. For example, largely developed based on Slot Attention \cite{Locatello2020Object}, the recently proposed methods BO-QSA \cite{Jia2023Unsupervised} and DINOSAUR \cite{Seitzer2023Bridging} have achieved encouraging results on \emph{real-world images}, demonstrating the great potential of this promising research topic in practical applications.

Because multiple design choices need to be considered in compositional modeling and inference, the categorization of existing methods is not straightforward. In addition, different methods usually use different sets of datasets and evaluation metrics to conduct experiments, which makes direct comparisons of these methods difficult. Furthermore, despite the growing research interest in reconstruction-based compositional scene representation learning with deep neural networks in recent years, the research on this topic is still limited to a relatively small scope. Therefore, there is a need to \emph{categorize and compare representative methods systematically} and \emph{summarize potential future directions that may spark broader research interest}, which motivates the writing of this survey.

This survey is organized as follows: Section 2 provides an overview of reconstruction-based compositional scene representation learning with deep neural networks; Sections 3 and 4 categorize existing methods from the perspectives of modeling of visual scenes and inference of scene representations, respectively; Section 5 provides benchmarks of representative methods that consider the most extensively studied problem setting in this research topic; Section 6 discusses limitations of existing methods; Section 7 looks forward to several directions for future research; Section 8 concludes the whole survey.

\section{Overview}

In this section, we will first describe the most widely adopted problem setting of reconstruction-based compositional scene representation learning with deep neural networks and the notations used in the paper, then introduce the development history of this research topic, and finally give an overview of categorizations of existing methods.

\subsection{Problem Setting and Notations}

\input{table_notation}

Under the considered problem setting, each visual scene is modeled as the composition of multiple layers (e.g., RGBA images) of visual concepts. For example, the visual scene in Fig. \ref{fig:intro} can be obtained by pasting the scaled and translated objects onto the background in the correct order. Each layer is associated with a series of representations that \emph{fully characterize} the corresponding visual concept (i.e., contain all the information needed to generate the RGBA image of the layer), and the collection of representations of all the layers form the compositional representations of the visual scene\footnote{In general, compositional scene representations are hierarchical, i.e., objects are composed of object parts, and coarser object parts are further composed of finer object parts. However, only very few of the methods surveyed in this paper (e.g., GSGN \cite{Deng2021Generative}) consider the hierarchy structure. Therefore, we omit the hierarchy for simplicity.}. The goal is to learn compositional scene representations under the autoencoding framework (i.e., using reconstruction error minimization as the main objective), with encoders and decoders implemented by deep neural networks.

A \emph{bonus feature} brought about by solving this problem is \emph{unsupervised panoptic segmentation} (also amodal if complete shapes of objects are considered in the compositional modeling of visual scenes). More specifically, by encoding the image of a visual scene into compositional scene representations and decoding the representations of individual visual concepts with the learned neural networks, segmentation results can be automatically obtained from the decoded images describing shapes of visual concepts.

The notations used throughout the paper are summarized in Table \ref{tab:notation}. $\boldsymbol{x} \in \mathbb{R}^{N \times C}$ and $\tilde{\boldsymbol{x}} \in \mathbb{R}^{N \times C}$ denote the observed and reconstructed images of the visual scene, respectively. $N$ denotes the number of pixels in each image. $C$ denotes the number of image channels. $K$ denotes the number of layers modeling objects in the visual scene (each visual scene is assumed to contain at most $K$ objects). The index of the background layer is $0$, and the indexes of $K$ object layers are between $1$ and $K$. $\boldsymbol{a}_{k} \in \mathbb{R}^{N \times C}$, $\boldsymbol{s}_{k} \in \mathbb{R}^{N}$, $\boldsymbol{\pi}_{k} \in \mathbb{R}^{N}$, and $o_{k} \in \mathbb{R}_{+}$ denote the appearance, complete shape (or logit of perceived shape if not modeling complete shape), perceived shape (may be incomplete due to occlusion), and the \emph{optional} variable describing the depth of the $k$th visual concept, respectively. $\boldsymbol{z}_k$ denotes the representation or the collection of representations of the background ($k = 0$) or the $k$th object ($1 \!\leq\! k \!\leq\! K$). $f_{\text{bck}}$ and $f_{\text{obj}}$ are decoders that take representations of the background and objects as inputs, respectively, i.e., $[\boldsymbol{a}_k, \boldsymbol{s}_k, o_{k}] = f_{\text{bck}}(\boldsymbol{z}_k)$ for $k = 0$ and $[\boldsymbol{a}_k, \boldsymbol{s}_k, o_{k}] = f_{\text{obj}}(\boldsymbol{z}_k)$ for $1 \!\leq\! k \!\leq\! K$. Perceived shapes $\boldsymbol{\pi}_{0:K}$ are computed based on complete shapes $\boldsymbol{s}_{0:K}$ and the optional variables $\boldsymbol{o}_{0:K}$ containing depth information. The reconstructed image $\tilde{\boldsymbol{x}}$ can be obtained by compositing the appearances $\boldsymbol{a}_{0:K}$ and perceived shapes $\boldsymbol{\pi}_{0:K}$.

It is worth mentioning that the number of object layers $K$ is an assumed upper bound and is \emph{not} necessarily equal to the actual number of objects in the visual scene. Because different visual scenes may contain different numbers of objects, the actual number of objects in each visual scene is assumed to be \emph{unknown} during learning. In addition, the separate modeling of the background with $\boldsymbol{z}_0$ and $f_{\text{bck}}$ is \emph{optional}, and the index of the separately modeled background is chosen to be $0$ for notational convenience only. Methods like RC \cite{Greff2016Binding} and N-EM \cite{Greff2017Neural} do not consider the background layer, resulting in complete information about the background being contained in every layer. Methods like IODINE \cite{Greff2019Multi} and Slot Attention \cite{Locatello2020Object} model background identically to objects, resulting in no natural way to distinguish between the background and objects.

\subsection{Development History}

\begin{figure*}[!t]
	\centering
	\input{fig_history}
	\caption{Development history of reconstruction-based compositional scene representation learning with deep neural networks. Nodes in \textit{green} and \textit{blue} colors are methods mainly based on parallel refinement and sequential attention, respectively.}
	\label{fig:history}
\end{figure*}
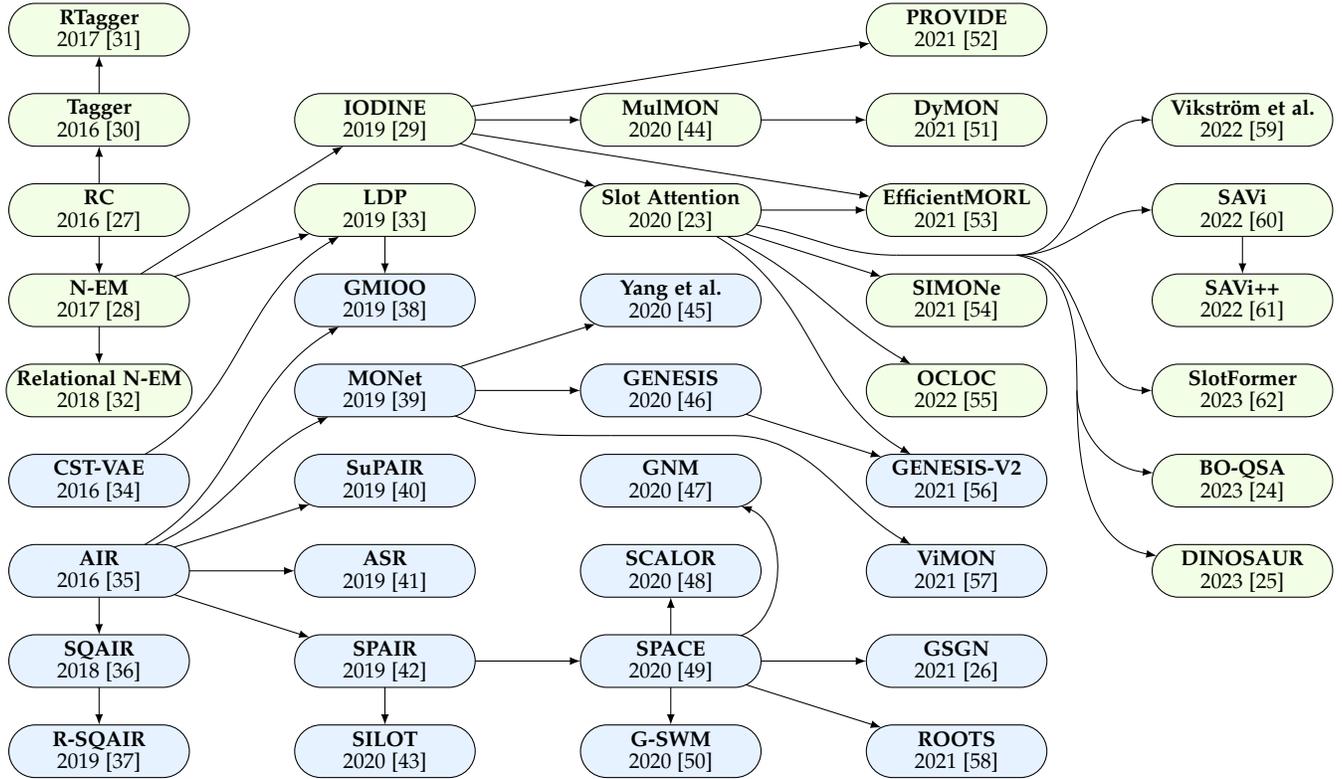

The development of reconstruction-based compositional scene representation learning with deep neural networks can be roughly divided into two streams, i.e., parallel refinement and sequential attention. Representative methods in these two streams are shown in Fig. \ref{fig:history}. In the early days of research on this topic, methods in the two streams are almost developed independently. As the research progress, some methods that mainly fall in one stream borrow ideas from the other stream. Currently, the slot attention mechanism \cite{Locatello2020Object} (a type of parallel refinement) is widely adopted mainly because of its simplicity, efficiency, and effectiveness.

\subsubsection{Parallel Refinement}

The main characteristic of methods in this stream is that the representations corresponding to different layers are randomly initialized and then iteratively refined in parallel. Training neural networks to perform iterative optimization is similar to the idea proposed by Andrychowicz et al. \cite{Andrychowicz2016Learning}. This strategy makes it possible to use different maximum numbers of layers for different visual scenes, an important property considered by all the methods surveyed in this paper because it makes the learned models generalize more easily to visual scenes containing more or fewer objects than the ones used for training. Given an image, a well-trained encoder network is expected to iteratively transform a set of noises randomly sampled from a particular distribution to the compositional representations of that image. The randomness in the initialization of compositional scene representations increases the difficulty of image reconstruction, leading to the amplification of the information bottleneck that encourages the encoder network to decompose visual scenes in the desired way. Early methods in this stream, such as RC \cite{Greff2016Binding}, Tagger \cite{Greff2016Tagger}, N-EM \cite{Greff2017Neural}, LDP \cite{Yuan2019Spatial}, and IODINE \cite{Greff2019Multi}, iteratively update compositional scene representations based on information in the pixel space. More specifically, at each iteration, compositional scene representations are decoded into reconstructions of decomposed visual concepts, which subsequently participate in the computation of inputs of the encoder network. Considering the high computational cost of converting between pixel space and representation space at each iteration, several methods proposed in recent years, such as Slot Attention \cite{Locatello2020Object}, EfficientMORL \cite{Emami2021Efficient}, SIMONe \cite{Kabra2021SIMONe}, and OCLOC \cite{Yuan2022Unsupervised}, perform parallel refinements based on information in the representation space, e.g., similarities between feature maps of the image and representations of visual concepts. Because iterative updates in the representation space do not involve expensive feature extraction and image generation operations, both computational complexity and memory consumption are reduced.

Developed based on the above methods, some methods additionally consider the modeling of object motions, relationships among objects, and the problem of learning from multiple viewpoints. For example, RTagger \cite{PremontSchwarz2017Recurrent} employs Recurrent Ladder Network that can model temporal relationships, extending Tagger to learn compositional representations from videos. Relational N-EM \cite{Steenkiste2018Relational} incorporates in N-EM a type of graph neural network (GNN) \cite{Gilmer2017Neural} with the attention mechanism \cite{Duan2017One}, thereby being able to model the relationships among objects. PROVIDE \cite{Zablotskaia2021PROVIDE} extends the iterative amortized inference used by IODINE to temporal data and can thus be applied to learning from videos. MulMON \cite{Li2020Learning} applies the idea of IODINE to images observed from multiple viewpoints, thereby capable of synthesizing images from novel viewpoints. DyMON \cite{Nanbo2021Object} extends MulMON to multiple-viewpoint videos by decoupling the influence of viewpoint change and object motion.

\subsubsection{Sequential Attention}

Besides parallel refinements, compositional scene representation can also be learned by sequentially attending to the local region of each visual concept. By adjusting the number of attention steps, this strategy also supports using different numbers of layers for different visual scenes. Early methods mainly designed based on sequential attention, such as CST-VAE \cite{Huang2016Efficient}, AIR \cite{Eslami2016Attend}, ASR \cite{Xu2019Multi}, and SPAIR \cite{Crawford2019Spatially}, use bounding boxes of objects as the attention mechanism and estimate bounding boxes based on the entire image. To achieve better spatial invariance during the inference of scene representations, SPAIR \cite{Crawford2019Spatially} combines the main idea of YOLO \cite{Redmon2016You} with compositional scene representation learning and estimates bounding boxes of objects based on local features. This strategy has inspired various methods, such as SPAIR \cite{Crawford2019Spatially} and SPACE \cite{Lin2020SPACE}, and is one of the current mainstream practices. The other current mainstream practice, which begins with MONet \cite{Burgess2019MONet} and is adopted by methods like Yang et al. \cite{Yang2020Learning} and GENESIS \cite{Engelcke2020GENESIS}, is to employ arbitrary-shaped masks that sequentially attend to regions of different visual concepts.

Some methods in this stream adopt strategies similar to the ones used in methods mainly based on parallel refinements. For example, GMIOO \cite{Yuan2019Generative} first sequentially initializes compositional scene representations with bounding boxes as the attention mechanism and then iteratively updates these representations to obtain better inference results. GENESIS-V2 \cite{Engelcke2021GENESIS} adopts the idea of instance coloring previously used in supervised instance segmentation, resembling Slot Attention \cite{Locatello2020Object} in that compositional scene representations are inferred based on weighted averages of feature maps, where the weights are computed based on similarities of local features. Some methods consider the modeling of object motions, relationships among objects, and the problem of learning from multiple viewpoints. For example, GNM \cite{Jiang2020Generative} models relationships among objects with hierarchical latent variables and infers latent variables similarly to SPACE. SQAIR \cite{Kosiorek2018Sequential} and ViMON \cite{Weis2021Benchmarking} are developed on AIR and MONet, respectively, and additionally consider the modeling of object motions. R-SQAIR \cite{Stanic2019R} integrates SQAIR with graph neural networks \cite{Steenkiste2018Relational,Santoro2018Relational} and can thus model interactions among objects. SILOT \cite{Crawford2020Exploiting} and SCALOR \cite{Jiang2020SCALOR} extend the ideas of SPAIR and SPACE to object tracking, respectively, thereby being able to learn compositional scene representations from videos. G-SWM \cite{Lin2020Improving} considers the modeling of both object relationships and object motions, thereby capable of effectively learning from videos with multimodal uncertainty. GSGN \cite{Deng2021Generative} uses hierarchical latent variables with a tree structure to model visual scenes and can hierarchically decompose objects. ROOTS \cite{Chen2021ROOTS} extends the idea of SPAIR to three-dimensional space and is thus able to learn compositional scene representations from multiple viewpoints.

\subsection{Categorizations of Methods}

Existing methods can be roughly categorized from two perspectives, i.e., the modeling of visual scenes and the inference of scene representations. There are several aspects to consider within each perspective, and Fig. \ref{fig:compare} makes comparisons from the most important ones.

\input{fig_compare}

\subsubsection{Modeling of Visual Scenes}
\label{sec:summary_modeling}

The modeling of visual scenes can be mainly categorized in three ways, i.e., the composition of layers, the modeling of shapes, and the representation of objects. Moreover, the modeling of extra properties, such as the number of objects, the layouts of scenes, the multiple viewpoints of scenes, and the motions of objects, are considered by some methods.

\textbf{Composition of Layers:} In compositional scene representation learning methods, a visual scene image is modeled as the composition of layers corresponding to visual concepts. These layers are usually composited in two ways, i.e., using spatial mixture models and weighted summations. \textbf{1)} Methods using spatial mixture models treat appearances of visual concepts as parameters of mixture components and perceived shapes of visual concepts as mixture weights. At each pixel of the visual scene image, the index of the observed layer is first sampled according to mixture weights, and then the color or intensity of the pixel is sampled from the corresponding mixture component. Representative methods using spatial mixture models include Tagger \cite{Greff2016Tagger}, RC \cite{Greff2016Binding}, N-EM \cite{Greff2017Neural}, IODINE \cite{Greff2019Multi}, MONet \cite{Burgess2019MONet}, GMIOO \cite{Yuan2019Generative}, and GENESIS \cite{Engelcke2020GENESIS}. \textbf{2)} Methods using weighted summations assume that the color or intensity of each pixel is sampled from a unimodal distribution, whose parameters are computed based on the weighted summation of appearances of visual concepts. Representative methods using weighted summations include AIR \cite{Eslami2016Attend}, SPAIR \cite{Crawford2019Spatially}, SuPAIR \cite{Stelzner2019Faster}, SQAIR \cite{Kosiorek2018Sequential}, SPACE \cite{Lin2020SPACE}, and Slot Attention \cite{Locatello2020Object}.

\textbf{Modeling of Shapes:} The perceived visual concepts in the visual scenes are often incomplete due to occlusions. Therefore, the modeling of shapes of visual concepts is relatively difficult. There are three main modeling approaches. \textbf{1)} The first approach is to directly normalize the outputs of neural networks into perceived shapes of visual concepts with the softmax function (the sum of perceived shapes of all visual concepts equals $1$ at every pixel). In this way, the complete shapes of visual concepts are not modeled. Representative methods include MONet \cite{Burgess2019MONet}, IODINE \cite{Greff2019Multi}, and Slot Attention \cite{Locatello2020Object}. \textbf{2)} The second approach computes perceived shapes by normalizing complete shapes and an extra type of variables characterizing the depths of objects. Objects with larger depths are occluded by objects with smaller depths. Representative methods include SPAIR \cite{Crawford2019Spatially}, SPACE \cite{Lin2020SPACE}, SCALOR \cite{Jiang2020SCALOR}, and GNM \cite{Jiang2020Generative}. \textbf{3)} The third approach transforms complete shapes into perceived shapes in a way similar to the stick-breaking process \cite{Teh2007Stick}. Objects with larger indexes are occluded by objects with smaller indexes. Representative methods include CST-VAE \cite{Huang2016Efficient}, LDP \cite{Yuan2019Spatial}, GMIOO \cite{Yuan2019Generative}, and ECON \cite{Kuegelgen2020Towards}. It is worth mentioning that a few methods, such as AIR \cite{Eslami2016Attend} and SQAIR \cite{Kosiorek2018Sequential}, directly add the appearances of all the layers to obtain the visual scene image without modeling the shapes of visual concepts.

\textbf{Representation of Objects:}
Objects in the visual scenes are usually represented in two ways. One is to use vectors that may take any values in the real vector space as representations of objects, and the other is to represent objects based on a finite number of prototypes. \textbf{1)} Methods representing objects with vectors in the real vector space can be further divided into two categories, i.e., using latent variables and embeddings. Methods using latent variables to represent objects include AIR \cite{Eslami2016Attend}, SPAIR \cite{Crawford2019Spatially}, IODINE \cite{Greff2019Multi}, GMIOO \cite{Yuan2019Generative}, MONet \cite{Burgess2019MONet}, GENESIS \cite{Engelcke2020GENESIS}, and EfficientMORL \cite{Emami2021Efficient}. Methods using embeddings include Taggers \cite{Greff2016Tagger}, N-EM \cite{Greff2017Neural}, LDP \cite{Yuan2019Spatial}, and Slot Attention \cite{Locatello2020Object}. The main difference between latent variables and embeddings is that the former ones are assumed to be sampled from prior distributions in the generative model, while the latter ones do not have prior distributions. \textbf{2)} Methods representing objects based on a finite number of prototypes can also be further divided into two categories. The first one is to directly use prototype images that can be transformed into layers of objects with interpretable transformations. Representative methods include DTI-Sprites \cite{Monnier2021Unsupervised} and PCDNet \cite{VillarCorrales2022Unsupervised}. The second one is to use prototype vectors that can be transformed into prototype images using neural networks. Representative methods include GSGN \cite{Deng2021Generative} and MarioNette \cite{Smirnov2021MarioNette}.

\textbf{Additional Modelings:} The above aspects are essential parts of the compositional modeling of visual scenes. Other properties, including \textbf{1)} the number of objects, \textbf{2)} the layouts of scenes, \textbf{3)} the multiple viewpoints of scenes, and \textbf{4)} the motions of objects, can be additionally modeled to extend the abilities of compositional scene representation learning methods. For example, some methods, such as AIR \cite{Eslami2016Attend}, GMIOO \cite{Yuan2019Generative}, SPAIR \cite{Crawford2019Spatially}, and SPACE \cite{Lin2020SPACE}, model the number of objects and can thus count objects in the visual scene image by inferring compositional scene representations. Some methods, such as GENESIS \cite{Engelcke2020GENESIS}, GENESIS-V2 \cite{Engelcke2021GENESIS}, GNM \cite{Jiang2020Generative} and GSGN \cite{Deng2021Generative}, model the layouts of scenes, thereby capable of generating more reasonable images under the unconditional setting. Methods like MulMON \cite{Li2020Learning}, ROOTS \cite{Chen2021ROOTS}, DyMON \cite{Nanbo2021Object}, SIMONe \cite{Kabra2021SIMONe}, and OCLOC \cite{Yuan2022Unsupervised} model the multiple viewpoints of scenes. Methods like SQAIR \cite{Kosiorek2018Sequential}, R-SQAIR \cite{Stanic2019R}, SILOT \cite{Crawford2020Exploiting}, SCALOR \cite{Jiang2020SCALOR}, and G-SWM \cite{Lin2020Improving} model the motions of objects.

\subsubsection{Inference of Scene Representations}
\label{sec:simple_inference}

The inference of scene representations can be categorized either from the inference frameworks or the attention mechanisms used during inference.

\textbf{Inference Frameworks:} The frameworks for inferring compositional scene representations are usually \emph{determined} by the modeling of visual scenes and can be divided into three categories. \textbf{1)} For methods that use latent variables as representations, e.g., AIR \cite{Eslami2016Attend}, GMIOO \cite{Yuan2019Generative}, IODINE \cite{Greff2019Multi}, GENESIS \cite{Engelcke2020GENESIS}, and GENESIS-V2 \cite{Engelcke2021GENESIS}, compositional scene representations are inferred based on \emph{amortized variational inference}, and parameters of approximate posterior distributions are estimated with neural networks. \textbf{2)} For methods that model visual scenes with spatial mixture models and explicitly infer indicating variables, e.g., N-EM \cite{Greff2017Neural}, Relational N-EM \cite{Steenkiste2018Relational}, and LDP \cite{Yuan2019Spatial}, the inference is performed in a way inspired by the \emph{Expectation Maximization} algorithm \cite{Dempster1977Maximum}. \textbf{3)} For methods that do not use any latent variables when modeling the visual scene, e.g., Tagger \cite{Greff2016Tagger}, RTagger \cite{PremontSchwarz2017Recurrent}, and Slot Attention \cite{Locatello2020Object}, the inference network is learned by minimizing the \emph{reconstruction errors} between the observed and reconstructed images of visual scenes.

\textbf{Attention Mechanisms:} To learn compositional scene representations, it is necessary to distinguish between different visual concepts in the same visual scene. The way to do this can be regarded as some kind of attention mechanism, and there are two main ways to design attention. \textbf{1)} One is to use rectangular attention, which first estimates bounding boxes of objects and then extracts representations of objects based on the contents in the bounding boxes. Representative methods include AIR \cite{Eslami2016Attend}, SQAIR \cite{Kosiorek2018Sequential}, GMIOO \cite{Yuan2019Generative}, SPAIR \cite{Crawford2019Spatially}, and SPACE \cite{Lin2020SPACE}. \textbf{2)} The other is to use \emph{arbitrary-shaped attention}, i.e., attention masks that have values between $0$ and $1$ and are of the same size as the image. Representative methods include N-EM \cite{Greff2017Neural}, IODINE, \cite{Greff2019Multi}, MONet \cite{Burgess2019MONet}, Slot Attention \cite{Locatello2020Object}, and GENESIS \cite{Engelcke2020GENESIS}.

\section{Modeling of Visual Scenes}
\label{sec:modeling}

This section describes the design choices for the modeling of visual scenes in detail, including the composition of layers, the modeling of shapes, the representation of objects, and the modeling of extra properties of visual scenes.

\subsection{Composition of Layers [C]}

\begin{figure}[t]
	\centering
	\subfloat[Spatial mixture models]{\includegraphics[width=0.99\columnwidth]{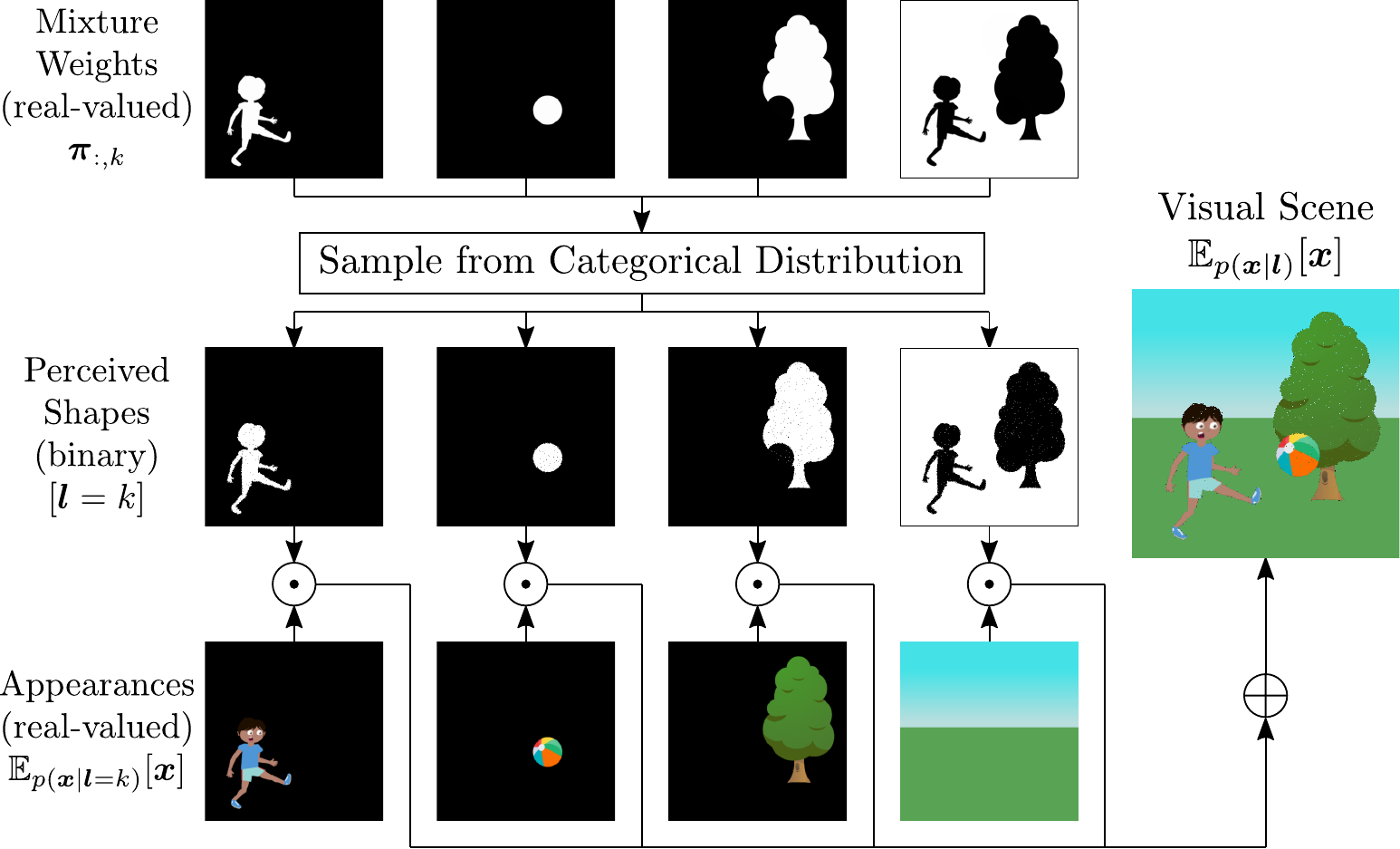}%
		\label{fig:compositing_mixture}}
	\\
	\subfloat[Weighted summations]{\includegraphics[width=0.99\columnwidth]{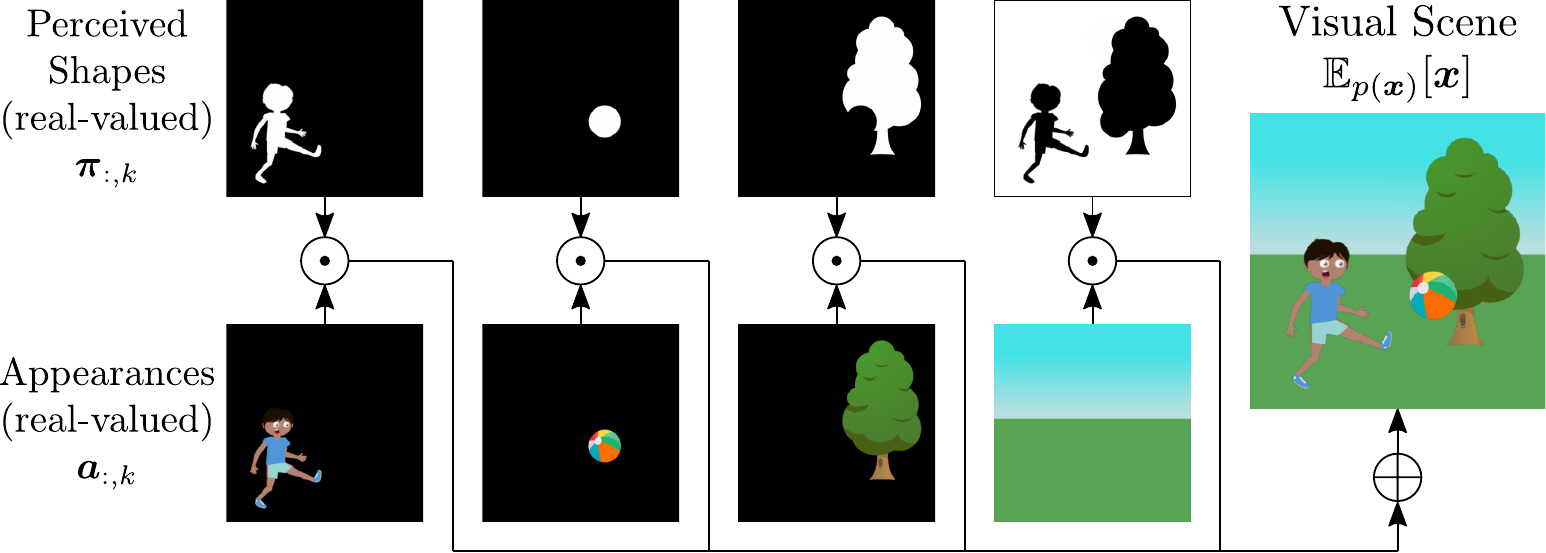}%
		\label{fig:compositing_summation}}
	\caption{Comparison of different types of layer composition.}
	\label{fig:composition_of_layers}
\end{figure}

Layers of visual concepts can be composited using either spatial mixture models or weighted summations. As shown in Fig. \ref{fig:composition_of_layers}, the main difference between these two choices lies in the ``softness'' of occlusions. Methods using spatial mixture models assume that only one layer of visual concept is observed at each pixel, and the color or intensity of each pixel is chosen to be the same as the observed layer. Methods using weighted summations assume that visual concepts are occluded softly, and the color or intensity of each pixel is the weighted summation of all the layers of visual concepts.

\subsubsection{Spatial Mixture Models (M)}
\label{sec:spatial_mixture_models}

When using spatial mixture models, it is assumed that all pixels of the visual scene image are independent. Each pixel $\boldsymbol{x}_n$ is associated with a variable $l_n$ indicating which layer is observed at that pixel. The joint probability distribution of all the $N$ pixels in the visual scene image is factorized as
\begin{equation}
    \label{equ:spatial_mixture}
    p(\boldsymbol{x}) = \prod_{n=1}^{N}{p(\boldsymbol{x}_n)} = \prod_{n=1}^{N}{\sum_{k=0}^{K}{\underbrace{p(l_n \!=\! k)}_{\pi_{k,n}} p(\boldsymbol{x}_n|l_n \!=\! k)}}
\end{equation}

In Eq. \eqref{equ:spatial_mixture}, $p(l_n)$ is the mixture weight describing the prior distribution of the indicating variable $l_n$, and $p(\boldsymbol{x}_n|l_n)$ is the mixture component describing the conditional distribution of the $n$th pixel, given that the $l_n$th layer is observed at that pixel. Although pixels are assumed to be independent, spatial dependencies of pixels can still be captured by modeling dependencies of mixture weights $p(l_n)$, mixture components $p(\boldsymbol{x}_n|l_n)$, or both of them with neural networks. Mixture weights $p(l_n)$ can be interpreted as the perceived shapes of visual concepts (may be incomplete due to occlusions) and are usually modeled by categorical distributions. For mixture components $p(\boldsymbol{x}_n|l_n)$, which are parameterized by the appearances of visual concepts, Bernoulli distributions (for binary images) or normal distributions (for grayscale or color images) are usually chosen.

The advantage of spatial mixture models is the better separation of different visual concepts. More specifically, the operation of sampling from categorical distributions requires that each pixel be reconstructed well by a single layer instead of a weighted summation of multiple layers. Therefore, the usage of spatial mixture models usually leads to sharper (closer to binary values) shapes of visual concepts. On the other hand, spatial mixture models have their drawbacks. The most significant one is that the difficulty of model parameter optimization is increased due to the extra discrete indicating variables $\boldsymbol{l}$. A simple trick to alleviate this problem is to use the weighted average of loss functions of spatial mixture models and weighted summations as the training objective. By gradually shifting from weighted summations to spatial mixture models during the training, undesired local optima that cause difficulties for spatial mixture models are more likely to be avoided.

\subsubsection{Weighted Summations (S)}

As with methods using spatial mixture models, methods using weighted summations also assume independence among all the pixels of the visual scene image, i.e., $p(\boldsymbol{x}) = \prod_{n=1}^{N}{p(\boldsymbol{x}_n)}$. The main difference between these two types of methods is that the latter ones do not model the distribution of each pixel $p(\boldsymbol{x}_n)$ with a mixture model but instead model $p(\boldsymbol{x}_n)$ with a unimodal distribution such as Bernoulli distribution (for binary images) or normal distribution (for grayscale or color images). The expected value of each pixel $\boldsymbol{x}_n$ is used as the parameter of the distribution $p(\boldsymbol{x}_n)$ and modeled as a weighted summation of appearances of all the visual concepts using the following expression.
\begin{equation}
    \label{equ:weighted_summation}
    \mathbb{E}_{\boldsymbol{x}_n \sim p(\boldsymbol{x}_n)}[\boldsymbol{x}_n] = \sum_{k=0}^{K}{\pi_{k,n} \boldsymbol{a}_{k,n}}
\end{equation}

In Eq. \eqref{equ:weighted_summation}, $\pi_{k,n}$ and $\boldsymbol{a}_{k,n}$ are the weight and appearance of the $k$th layer of visual concept at the $n$th pixel, respectively. If the constraint $(\forall n) \sum_{k=0}^{K}{\pi_{k,n}} \!=\! 1$ is imposed, the weights $\boldsymbol{\pi}_{k}$ of the $k$th layer can be interpreted as the perceived shape of the $k$th visual concept. Despite the assumption of independence of all the pixels, the weights $\boldsymbol{\pi}$, the appearances $\boldsymbol{a}$, or both can be assumed to be spatially dependent for modeling the spatial relationships of pixels.

As stated in the descriptions of spatial mixture models, the main advantage of using weighted summations is easier training of models brought by the simpler loss function. This advantage is more noticeable on relatively hard datasets where models converge slowly in the early stage of training. The main downside of choosing weighted summations is that the learned representations corresponding to one visual concept are more likely to contain information about other visual concepts, resulting in more artifacts in regions where multiple objects overlap when images are generated in the unconditional setting. If the complete disentanglement of different visual concepts and the unconditional image generation quality are not of vital importance, it is suggested to composite layers with weighted summations for simplicity. Otherwise, adopting a more complex strategy that combines spatial mixture models and weighted summations (e.g., the trick mentioned at the end of Section \ref{sec:spatial_mixture_models}) is preferred.

\subsection{Modeling of Shapes [S]}

\begin{figure}[t]
	\centering
	\subfloat[Directly normalize outputs of neural networks]{\includegraphics[width=0.99\columnwidth]{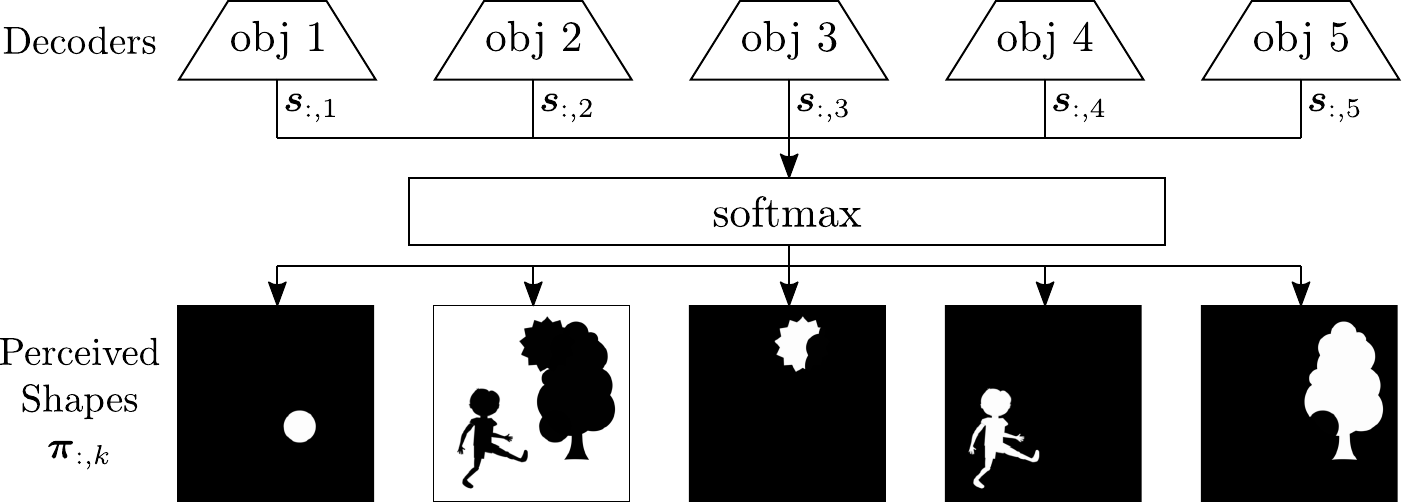}%
		\label{fig:shape_softmax}}
	\\
	\subfloat[Composite complete shapes and an extra type of variables $\boldsymbol{o}_{1:K}$]{\includegraphics[width=0.99\columnwidth]{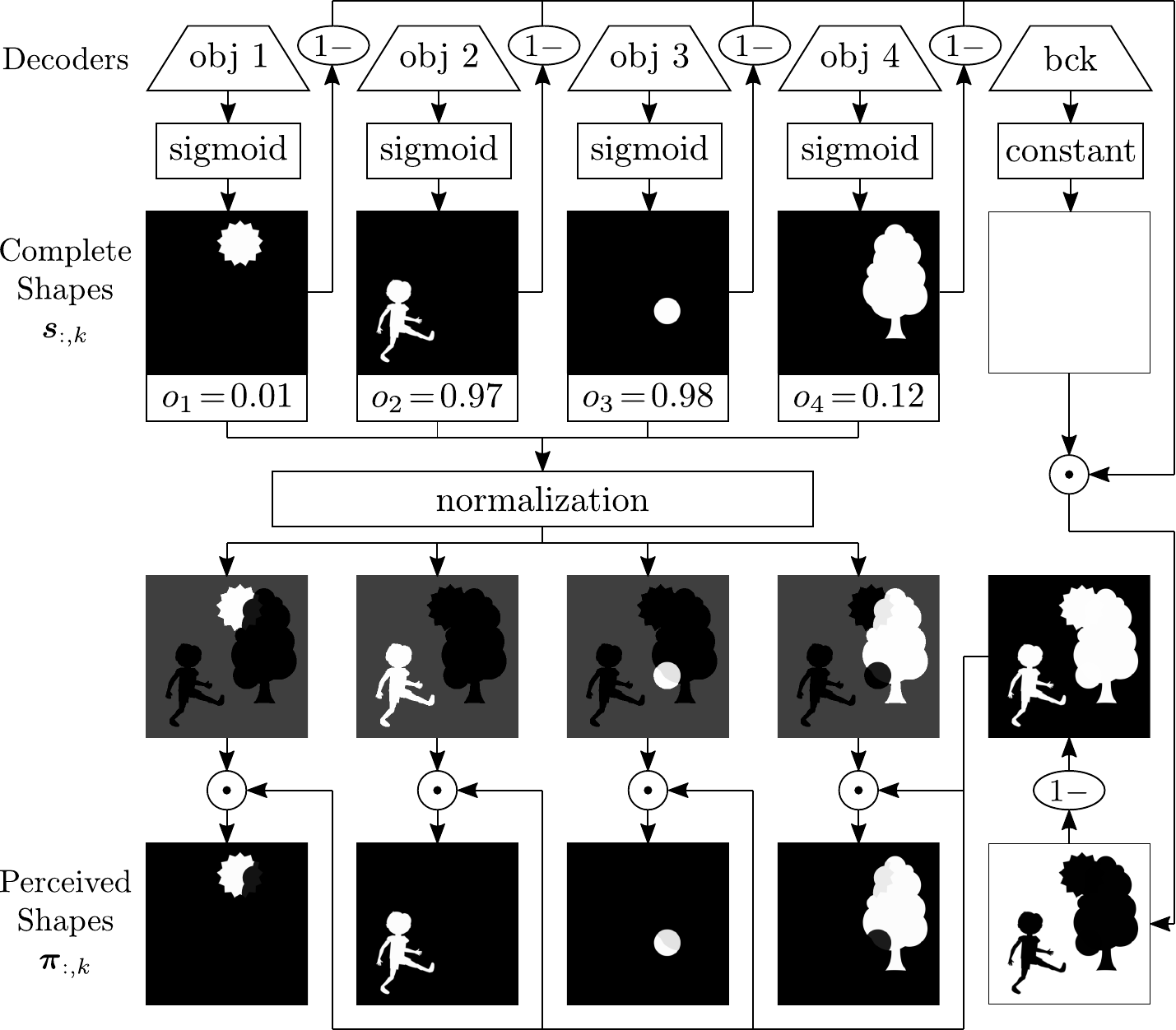}%
		\label{fig:shape_complete_soft}}
	\\
	\subfloat[Composite complete shapes similarly to the stick-breaking process]{\includegraphics[width=0.99\columnwidth]{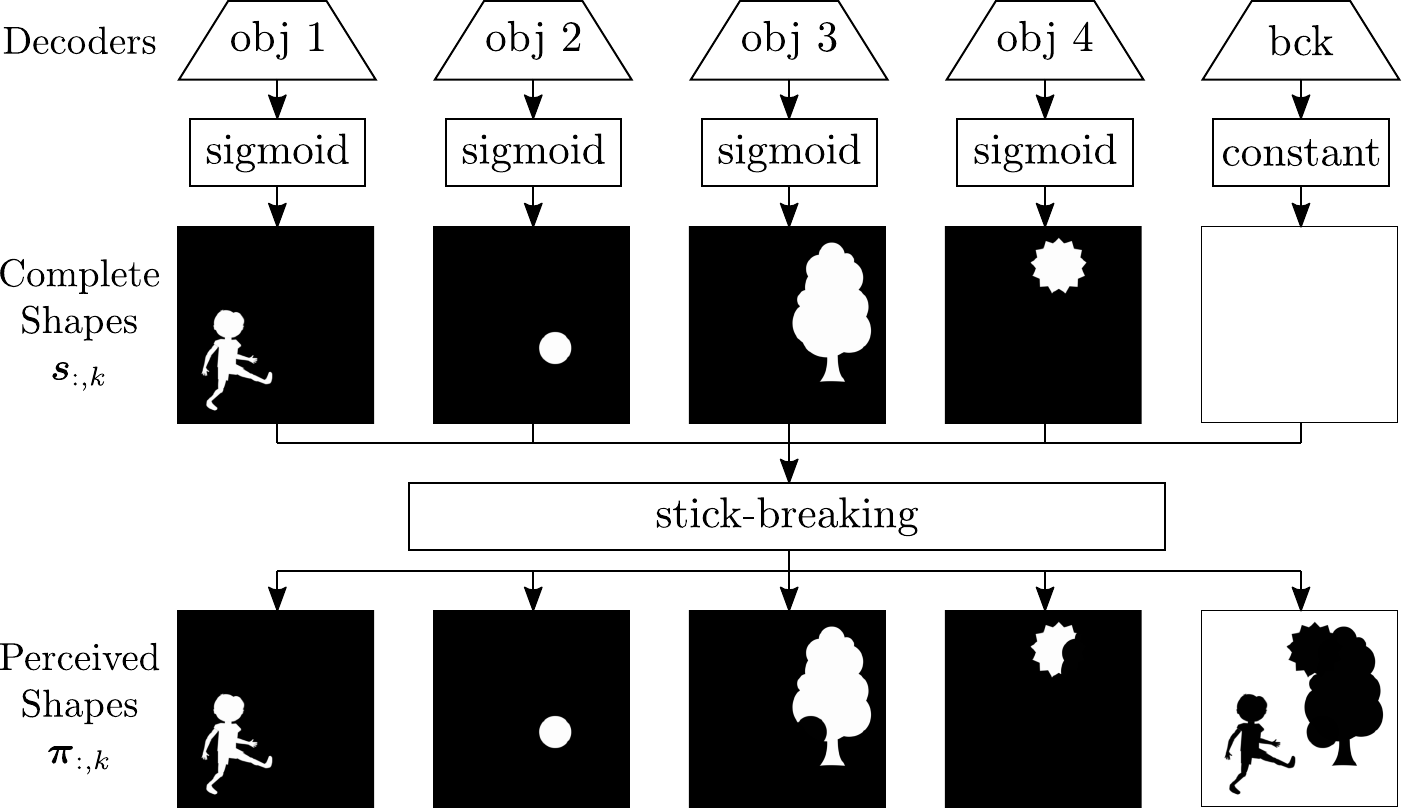}%
		\label{fig:shape_complete_hard}}
	\caption{Comparison of different types of shape modeling.}
	\label{fig:modeling_of_shapes}
\end{figure}

Except for very few methods, such as AIR \cite{Eslami2016Attend} and SQAIR \cite{Kosiorek2018Sequential}, which composite layers into visual scene images using weighted summations that do not constrain the weights to sum up to $1$ at each pixel, modeling the shapes of visual concepts is a necessary part of compositional scene representation learning. As shown in Fig. \ref{fig:modeling_of_shapes}, existing methods that model shapes of visual concepts can be classified into three categories. The first category has the advantage of simplicity. The second and third categories are advantageous for naturally supporting the estimations of complete shapes and depth ordering of objects without heuristics.

\subsubsection{Direct Normalization of Network Outputs (N)}

A prerequisite for computing perceived shapes of visual concepts by directly normalizing outputs of neural networks is to model the object and the background identically. Therefore, the perceived shape of the $0$th layer corresponding to the specially modeled background is $0$ at all pixels by assumption. The representation of each visual concept $\boldsymbol{z}_k$ is transformed into a real-valued variable $\boldsymbol{s}_{k}$ containing the shape information of the $k$th layer with a neural network, with the range of the value that each entry $s_{k,n}$ can take unconstrained. The mixture weights of the spatial mixture models or the weights used in the weighted summations $\boldsymbol{\pi}$ are computed using the softmax function.
\begin{equation}
    \pi_{k,n} = 
    \begin{dcases}
        0, & k = 0 \\
        \frac{\exp(s_{k,n})}{\sum_{k'=1}^{K}{\exp(s_{k',n})}}, & 1 \leq k \leq K
    \end{dcases}
\end{equation}

The most significant advantage of modeling shapes with direct normalization of network outputs is simplicity. There is no need to distinguish between object and background, nor to model the complete shapes and depth ordering of objects, thereby leading to easier optimization of model parameters than the other two alternatives. The main drawback of this modeling choice, i.e., the impossibility of obtaining amodal segmentation results without complex and non-exhaustive heuristics, also comes from simplicity. If this crucial drawback is not an issue, then modeling shapes by directly normalizing network outputs is the primary choice.

\subsubsection{Normalization of Complete Shapes and Ordering (O)}

It is also possible to model shapes of visual concepts by first transforming outputs of neural networks into variables $\boldsymbol{s}_{0:K}$ and $\boldsymbol{o}_{1:K}$ that indicate the complete shapes and depth ordering of objects, respectively, and then normalizing these variables into perceived shapes. Values of $o_k$ are constrained to be greater than $0$ by using functions like the sigmoid function or the exponential function. Objects with larger $o_k$ are assumed to occlude those with smaller $o_k$, and the background is assumed to be occluded by all the objects. The ascending depth ordering of objects can be obtained by sorting variables $o_k$ in descending order. Perceived shapes $\boldsymbol{\pi}_{0:K}$ are computed using the following expression.
\begin{equation}
    \label{equ:masked_softmax}
    \pi_{k,n} = 
    \begin{dcases}
        \prod\nolimits_{k'=1}^{K}{(1 - s_{k',n})}, & k = 0 \\
        (1 - \pi_{n,0}) \frac{s_{k,n} \, o_k}{\sum_{k'=1}^{K}{s_{k',n} \, o_{k'}}}, & 1 \leq k \leq K
    \end{dcases}
\end{equation}

Because complete shapes are considered in this modeling choice, amodal segmentation results can be directly obtained by decoding and compositing the inferred compositional scene representations. Besides the built-in ability to perform amodal segmentation, another main characteristic of this modeling choice is approximating the discrete depth ordering of objects with continuous variables $\boldsymbol{o}$. On the one hand, because all the variables that need to be estimated are continuous, gradient-based optimization techniques that are well-suited for neural networks can be directly applied. On the other hand, because the inferred or generated $o_k$ for each layer usually does not differ significantly, the occlusions between objects are more or less transparent, resulting in artifacts in regions where multiple objects overlap. Therefore, when implementing this modeling choice, it is suggested to assign a temperature hyperparameter to variables $\boldsymbol{o}$ and gradually decrease the temperature during the training. If the ability of amodal segmentation is needed and the qualities of image reconstruction and image generation are not the main concerns, then this modeling choice is preferable.

\subsubsection{Stick-Breaking Composition of Complete Shapes (S)}

Another modeling choice is to compute perceived shapes of visual concepts in a way similar to the stick-breaking process \cite{Teh2007Stick}. Objects with smaller indexes are assumed to occlude objects with larger indexes, and the ascending depth ordering of objects is identical to the ascending index ordering. Complete shapes $\boldsymbol{s}_{0:K}$ are transformed into perceived shapes $\boldsymbol{\pi}_{0:K}$ using the following expression.
\begin{equation}
    \label{equ:stick_breaking}
    \pi_{k,n} = 
    \begin{dcases}
        \prod\nolimits_{k'=1}^{K}{(1 - s_{k',n})}, & k = 0 \\
        s_{k,n} \prod\nolimits_{k'=1}^{k-1}{(1 - s_{k',n})}, & 1 \leq k \leq K
    \end{dcases}
\end{equation}

As with the second modeling choice described above, the stick-breaking composition of complete shapes also enables amodal segmentation without heuristics. Unlike the above, this modeling choice uses discrete variables (i.e., indexes of layers) to describe the depth ordering of objects. The main advantage is that the modeling of depth ordering of objects does not introduce additional artifacts in the reconstructed or generated images, while the main disadvantage is that the optimization becomes more difficult due to discrete variables. If the computational cost is not a concern, a straightforward way is to enumerate all the $K!$ possible orderings ($K$ is the number of object layers) and choose the one that leads to the lowest reconstruction error. However, a much more computationally feasible solution is needed in real situations. One can either estimate the depth ordering with neural networks and apply discrete variable optimization methods (e.g., NVIL \cite{Mnih2014Neural} and VIMCO \cite{Mnih2016Variational}) or use heuristics to determine the depth ordering greedily. Well-designed heuristics may reduce the learning difficulty (provided that the estimated depth ordering is correct) because fewer variables need to be inferred. In cases where segmentation results are required to be amodal, this type of shape modeling is the first choice if such heuristics can be found. Otherwise, a more complex strategy that combines the second modeling choice and this one based on straight-through estimators \cite{Bengio2013Estimating} can be applied.

\subsection{Representation of Objects [R]}
\label{sec:object_representations}

Based on the possible values that representations can take, existing methods can be classified into two categories. Methods in the first category use representations in the real vector space and have the advantages of requiring simpler inference and training strategies (because there is no need to infer the discrete prototype indexes) and being able to handle huge variability (may be caused by variations of visual concepts themselves or variations of global factors like viewpoints and lighting effects). Methods in the second category represent objects based on a finite number of prototypes and are advantageous in that they have an additional ability to distinguish different types of objects. The representation choices used in these two categories are not mutually exclusive. One can combine the ideas of these two, e.g., using both prototypes (describe categories of objects) and variables in real vector space (encode intra-class variability), to increase the expressiveness of representations.

\subsubsection{Representations in Real Vector Space (R)}

Depending on whether defining the prior distributions of representations, methods using representations in real vector space can be further divided into two categories.

\textbf{Embeddings (E):} One choice is to use embeddings with no prior distribution placed on them. Methods employing this representation choice only define the conditional generative model of visual scenes, thereby not providing a natural way to generate images in the unconditional setting. A simple strategy that can overcome this problem is to learn the distribution of inferred embeddings after compositional scene representation learning, such that images can be generated based on embeddings sampled from this distribution.

Using embeddings to represent objects is advantageous in simplicity and relatively high reconstruction quality. Because no regularization is placed on these representations, models can focus more on minimizing the reconstruction error. If reconstruction quality is of vital importance, this representation choice is preferable.

\textbf{Latent Variables (L):} The other choice is to represent objects with latent variables that have prior distributions. A commonly used prior distribution is the standard normal distribution $\mathcal{N}(\boldsymbol{0}, \boldsymbol{I})$. Methods using this representation choice define the complete generative model of visual scenes and naturally support generating visual scene images similar to those used for training.

The prior distribution of latent variables can be seen as a regularization that encourages the inferred representations to follow a specific distribution. This regularization amplifies the information bottleneck by increasing the difficulty of image reconstruction and may help obtain better decomposition results. If a better decomposition of the visual scene is more important than a slight decrease in reconstruction quality, then representing objects with latent variables is preferred over embeddings.

\subsubsection{A Finite Number of Prototypes (P)}
\label{sec:prototypes}

In methods that represent objects based on a finite number of prototypes, a dictionary of prototypes shared across all the visual scenes is learned. Layers of visual concepts are generated by first selecting prototypes according to the variables indicating categories of visual concepts and then transforming the selected prototypes. According to the forms of prototypes, existing methods can be divided into two categories, i.e., using prototype images and low-dimensional prototype vectors. The main difference is that an extra step to transform prototype vectors into prototype images is needed for methods in the second category.

\textbf{Images (I):} When representing prototypes in the form of images, the transformations from prototypes to layers of visual concepts are all done in the image space. These transformations (e.g., geometric transformations and colorimetric transformations) are usually well interpretable, and the inference of scene representations involves the estimations of both categories of prototypes and parameters of transformations. Prototype images are randomly initialized at the beginning of the training and automatically learned from data as training progresses.

This representation choice is advantageous for its simplicity and higher interpretability. There is no need to train decoder networks that perform transformations from low-dimensional vector space to high-dimensional image space, and the learned prototypes can be understood by humans easily. The two most notable drawbacks are the relatively large space required to store the learned prototypes and the difficulty of optimization in high-dimensional image space. Therefore, besides the requirement that combinations of predefined transformations should completely cover intra-class variability, the applicability of this representation choice also requires that the number of object categories and the dimensionality of prototype images be relatively small. If the above requirements are satisfied, and the continual learning of prototypes is needed, then using prototype images is the first choice. This is because the learning of new object categories does not interfere with the previously learned prototype images, making the application of sophisticated continual learning techniques unnecessary.

\textbf{Vectors (V):} It is also possible to represent prototypes with low-dimensional vectors and apply an additional neural network to decode prototype vectors into images. The generated images can be converted into layers of visual concepts using similar transformations employed by methods representing prototypes as images. The learning of prototypes is also similar to the above methods. All the prototype vectors in the dictionary are first randomly initialized and then iteratively updated.

As with methods representing objects with prototype images, methods using prototype vectors also model the intra-class variability of visual concepts with transformations in the image space, thereby not suitable for cases where intra-class variability is complex enough that defining possible transformations is infeasible. Compared with prototype images, prototype vectors are advantageous in that both the size of the space needed to store prototypes and the dimensionality of the space in which the prototypes are optimized are reduced. However, because all the prototypes share the same decoder network that transforms them into images, learning one prototype will lead to changes in others, making the continual learning of new visual concepts difficult (i.e., the already learned visual concepts suffer from the problem of catastrophic forgetting). If the capability of continual learning is not required or an effective continual learning strategy can be found, then the use of prototype vectors is preferred over prototype images.

\subsection{Additional Modelings}
\label{sec:additional_modelings}

The aspects mentioned above are essential in the modeling of visual scenes. Besides these, several properties of visual scenes can be additionally considered to extend the abilities of compositional scene representation learning methods. For example, modeling the number of objects provides a natural way to count objects, modeling scene layouts improves the plausibility of images generated in the unconditional setting, modeling multiple viewpoints of visual scenes enables learning from additional viewpoints that provide complementary information, and modeling motions of objects enables learning dynamics of objects from videos and predicting the changes of visual scenes in the future.

\subsubsection{Number of Objects [N]}

The number of objects in the visual scene image is usually modeled as the sum of binary variables $\boldsymbol{z}_{1:\infty}^{\text{pres}}$ or $\boldsymbol{z}_{1:I,1:J}^{\text{pres}}$ that indicate the presence of objects in the visual scene. According to the modeling of these binary variables, existing methods can be divided into three categories:
\begin{itemize}[leftmargin=*]
	\item In methods like AIR \cite{Eslami2016Attend} and SQAIR \cite{Kosiorek2018Sequential}, each visual scene is assumed to contain possibly an infinite number of objects. Variables $\boldsymbol{z}_{1:\infty}^{\text{pres}}$ are generated as described below.
	\begin{align}
		z_{k}^{\text{cond}} & \sim \Bernoulli(\alpha), \mkern34mu k \geq 1 \\
		z_{k}^{\text{pres}} & = \prod\nolimits_{k'=1}^{k}{z_{k'}^{\text{cond}}}, \mkern38mu k \geq 1
	\end{align}
	In the above expressions, $\alpha$ is a hyperparameter. It is ensured that the starting $\sum_{k=1}^{\infty}{z_{k}^{\text{pres}}}$ entries of $\boldsymbol{z}_{1:\infty}^{\text{pres}}$ are all $1$ and the following entries are all $0$. Therefore, $\boldsymbol{z}_{1:\infty}^{\text{pres}}$ can be seen as the unary code for the number of objects. This modeling choice is advantageous in terms of simplicity. However, the inference of $\boldsymbol{z}_{1:\infty}^{\text{pres}}$ is relatively difficult due to the direct dependencies among these discrete variables.
	\item In methods like SPAIR \cite{Crawford2019Spatially} and SPACE \cite{Lin2020SPACE}, each visual scene image is partitioned into $I \!\times\! J$ regions, and each region is associated with a binary variable $z_{i,j}^{\text{pres}}$ indicating whether there is an object located in that region. These variables are modeled using the following expression.
	\begin{equation}
		z_{i,j}^{\text{pres}} \sim \Bernoulli(\alpha), \mkern30mu 1 \!\leq\! i \!\leq\! I, 1 \!\leq\! j \!\leq\! J
	\end{equation}
	Because all the variables $z_{i,j}^{\text{pres}}$ are independent, the difficulty of inference is reduced. However, the number of objects that can be modeled is constrained to be at most $I \times J$. It should be noted that it is possible to support more objects by either increasing the values of $I$ and $J$ or sampling multiple binary variables for each region.
	\item Methods like GMIOO \cite{Yuan2019Generative} assume that each visual scene may contain infinite objects and model the number of objects in a way inspired by the Indian Buffet Process \cite{Ghahramani2006Infinite}. The procedure to generate $\boldsymbol{z}_{1:\infty}^{\text{pres}}$ is shown below.
	\begin{align}
		\nu_{k} & \sim \Beta\big(\alpha, 1\big), \mkern107mu k \geq 1 \\
		z_{k}^{\text{pres}} & \sim \Bernoulli\big({\textstyle\prod}_{k'=1}^{k}{\nu_{k'}}\big), \mkern30mu k \geq 1
	\end{align}
	The introduction of continuous variables $\boldsymbol{\nu}$ breaks the direct dependencies among discrete variables $z_{k}^{\text{pres}}$, thus reducing the difficulty of inferring $\boldsymbol{z}_{1:\infty}^{\text{pres}}$. Although more variables need to be inferred, this modeling choice is usually preferred over the first one due to easier inference.
\end{itemize}

\subsubsection{Layouts of Scenes [L]}

The layouts of visual scenes can be modeled by considering relationships among objects in the modeling of visual scenes. A viable way is to model dependencies among object representations. According to how dependencies are modeled, existing methods can be classified into three categories.

\begin{itemize}[leftmargin=*]
	\item In methods like GENESIS \cite{Engelcke2020GENESIS} and GENESIS-V2 \cite{Engelcke2021GENESIS}, dependencies are modeled by factorizing the joint probability distribution $p(\boldsymbol{z}_{1:K})$ using the chain rule and modeling the conditional distributions using a recurrent neural network (RNN). The factorization of $p(\boldsymbol{z}_{1:K})$ is
	\begin{equation}
		\label{equ:layout_rnn}
		p(\boldsymbol{z}_{1:K}) = \prod_{k=1}^{K}{p(\boldsymbol{z}_{k}|\boldsymbol{z}_{1:k-1})} = \prod_{k=1}^{K}{p(\boldsymbol{z}_{k}|\boldsymbol{h}_{k})}
	\end{equation}
	In the above expression, $\boldsymbol{h}_{k} \!=\! f_{\text{RNN}}(\boldsymbol{z}_{k-1}, \boldsymbol{h}_{k-1})$ denotes the hidden states of the RNN $f_{\text{RNN}}$. The most significant advantages of this modeling choice are simplicity and versatility. Almost all existing methods can be augmented with it. The main disadvantage is that dependencies are modeled in a chain structure, requiring very powerful neural networks to be trained to learn dependencies well.
	\item In methods like GNM \cite{Jiang2020Generative}, each visual scene image is partitioned into $I \!\times\! J$ regions, and each region is associated with a set of latent variables $\boldsymbol{z}_{i,j}$ characterizing that region. The layout of the visual scene is described by an extra latent variable $\boldsymbol{z}^{\text{lyt}}$ on which variables $\boldsymbol{z}_{i,j}$ depend. The joint probability distribution of all latent variables is
	\begin{equation}
		p(\boldsymbol{z}^{\text{lyt}}, \boldsymbol{z}_{1:I,1:J}) = p(\boldsymbol{z}^{\text{lyt}}) \prod_{i=1}^{I}{\prod_{j=1}^{J}{p(\boldsymbol{z}_{i,j}|\boldsymbol{z}^{\text{lyt}})}}
	\end{equation}
	Compared to the first modeling choice, this choice can better handle complex dependencies because the assumed structure of dependencies is not restricted to a chain. However, the use of this modeling choice is more limited, as it requires partitioning visual scenes into local regions when modeling visual scenes.
	\item In methods like GSGN \cite{Deng2021Generative}, the layout of each visual scene is assumed to be hierarchical and represented in the form of a tree. The leaf nodes represent primitive entities of the visual scene (i.e., object parts or objects), and the edges represent transformations applied to lower-level entities to compose higher-level entities. Let $\mathcal{V}$ denote the set of all the nodes, and let $\boldsymbol{z}_{v}$ denote the representation of node $v$. The joint probability of all the node representations is factorized according to the structure of the tree.
	\begin{equation}
		p(\boldsymbol{z}_{\mathcal{V}}) = \prod_{v \in \mathcal{V}}{p(\boldsymbol{z}_{v}|pa(\boldsymbol{z}_{v}))}
	\end{equation}
	In the above expression, $pa(\boldsymbol{z}_v)$ denotes the parent node of node $v$. This modeling choice has the advantage of learning hierarchical dependencies and has the greatest potential as it naturally supports compositional scene representations with hierarchical structures. However, this choice is more difficult to employ due to the increased complexities in both modeling and inference.
\end{itemize}

\subsubsection{Multiple Viewpoints of Scenes [V]}

For visual scenes that may be observed from different viewpoints, it is necessary to model multiple viewpoints of the scene for learning viewpoint-independent representations of visual concepts. Let $M$ denote the number of viewpoints modeled. The representations of each visual scene are divided into two parts. The first part $\boldsymbol{z}_{1:M}^{\text{view}}$ contains information about the $M$ viewpoints. The second part $\boldsymbol{z}_{0:K}^{\text{attr}}$ encodes viewpoint-independent attributes of visual concepts. For each viewpoint, layers of visual concepts are computed by
\begin{equation}
    \label{equ:multi_view}
    [\boldsymbol{a}_{m,k}, \boldsymbol{s}_{m,k}, o_{m,k}] =
    \begin{cases}
        f_{\text{bck}}(\boldsymbol{z}_{m}^{\text{view}}, \boldsymbol{z}_{k}^{\text{attr}}), & k = 0 \\
        f_{\text{obj}}(\boldsymbol{z}_{m}^{\text{view}}, \boldsymbol{z}_{k}^{\text{attr}}), & 1 \leq k \leq K
    \end{cases}
\end{equation}
In Eq. \eqref{equ:multi_view}, $f_{\text{bck}}$ and $f_{\text{obj}}$ are neural networks that transform $\boldsymbol{z}_{m}^{\text{view}}$ and $\boldsymbol{z}_{k}^{\text{attr}}$ into layers of visual concepts. By distinguishing between viewpoints and viewpoint-independent representations, a well-trained model can synthesize images of the same visual scene from novel viewpoints by keeping viewpoint-independent representations $\boldsymbol{z}_{1:K}^{\text{attr}}$ unchanged and only modifying $\boldsymbol{z}_{m}^{\text{view}}$. According to the assumption of viewpoints, existing methods can be divided into the following two categories.

\begin{itemize}[leftmargin=*]
	\item In methods assuming the exact viewpoints from which to observe each visual scene are known, e.g., MulMON \cite{Li2020Learning}, DyMON \cite{Nanbo2021Object}, and ROOTS \cite{Chen2021ROOTS},  the ground truth viewpoints are used as $\boldsymbol{z}_{1:M}^{\text{view}}$ in both the training and testing. These methods have the advantage of being able to synthesize images observed from specific viewpoints. The major challenge for these methods is to include as many viewpoint-independent attributes and as little viewpoint information as possible in $\boldsymbol{z}_{0:K}^{\text{attr}}$. A common practice is to infer $\boldsymbol{z}_{0:K}^{\text{attr}}$ using images observed from some viewpoints and encourage the learned representations to be able to predict images observed from other viewpoints as accurately as possible during training.
	\item In methods considering a harder problem that compositional scene representations are learned from multiple viewpoints without knowing viewpoints, e.g., SIMONe \cite{Kabra2021SIMONe} and OCLOC \cite{Yuan2022Unsupervised}, both $\boldsymbol{z}_{1:M}^{\text{view}}$ and $\boldsymbol{z}_{0:K}^{\text{attr}}$ need to be inferred. SIMONe \cite{Kabra2021SIMONe} assumes that viewpoints are ordered and can thus utilize temporal information to assist the learning of compositional scene representations. OCLOC \cite{Yuan2022Unsupervised} assumes that viewpoints are unordered, thereby applicable to scenarios where relationships among viewpoints are unknown. Compared with methods that utilize viewpoint annotations, methods in this category have the advantage of learning from completely unlabeled data.
\end{itemize}

\subsubsection{Motions of Objects [M]}

To better learn from videos, the motions of objects need to be modeled because they are the main causes of temporal changes in video frames. Methods considering the motions of objects in the modeling of visual scenes usually distinguish objects in each frame into two parts. The first part consists of objects seen in the previous frames. The second part consists of objects that newly appear in the current frame. The modelings of objects in these two parts are usually termed \emph{propagation} and \emph{discovery}, respectively. Let $\boldsymbol{z}_{t,k}$ denote the representation of the $k$th object in the $t$th frame, and let $\tilde{K}_t$ denote the total number of distinct objects seen in the first $t$ frames. The joint probability distribution of representations of all the $\tilde{K}_{T}$ objects seen in $T$ frames is factorized in the way described below.
\begin{equation}
    \label{equ:object_motion}
    p(\boldsymbol{z}) \!=\!\! \prod_{t=1}^{T}{\underbrace{p^{\text{P}}(\boldsymbol{z}_{t,1:\tilde{K}_{t-1}}|\boldsymbol{z}_{1:t-1,1:\tilde{K}_{t-1}})}_{\text{propagation}} \underbrace{p^{\text{D}}(\boldsymbol{z}_{t,\tilde{K}_{t-1}+1:\tilde{K}_{t}})}_{\text{discovery}}}
\end{equation}
For simplicity, newly discovered objects are usually modeled as independent, i.e., the discovery part is factorized as $\prod_{k=\tilde{K}_{t-1}+1}^{\tilde{K}_{t}}{p(\boldsymbol{z}_{t,k})}$. According to whether considering the relationships among objects in the propagation part, existing methods can be divided into two categories.

\begin{itemize}[leftmargin=*]
	\item In methods not considering the relationships among objects, e.g., SQAIR \cite{Kosiorek2018Sequential} and SILOT \cite{Crawford2020Exploiting}, the motions of objects are independently modeled, and the propagation part is factorized as $\prod_{k=1}^{\tilde{K}_{t-1}}{p^{\text{P}}(\boldsymbol{z}_{t,k}|\boldsymbol{z}_{1:t-1,k})}$. A common practice is to apply a recurrent neural network (RNN) to summarize the information from all previous frames. The main advantage of this modeling choice is simplicity.
	\item In methods considering the relationships among objects, e.g., R-SQAIR \cite{Stanic2019R} and G-SWM \cite{Lin2020Improving}, the propagation part captures the interactions of objects. To model relationships among a variable number of objects while keeping the computational cost relatively low, it is often assumed that representations of all objects in the same frame are conditionally independent given the representations in all previous frames, i.e., the propagation part is factorized as $\prod_{k=1}^{\tilde{K}_{t-1}}\!\!{p^{\text{P}}(\boldsymbol{z}_{t,k}|\boldsymbol{z}_{1:t-1,1:\tilde{K}_{t-1}})}$.
	The conditional distribution $p^{\text{P}}(\boldsymbol{z}_{t,k}|\boldsymbol{z}_{1:t-1,1:\tilde{K}_{t-1}})$ can be modeled using the combination of a recurrent neural network (RNN) and a graph neural network (GNN). This modeling choice is advantageous for modeling motions involving interactions among objects and is usually preferred over the above one.
\end{itemize}

\section{Inference of Scene Representations}
\label{sec:inference}

Inferring scene representations is the inverse problem of modeling visual scenes. There are two main aspects to consider. The first is the choice of inference frameworks, and the second is the choice of attention mechanisms.

\subsection{Inference Frameworks [I]}

The inference framework is usually \emph{determined} by how visual scenes are modeled. Therefore, there is no need to struggle with the choice of this aspect. For methods representing visual concepts with latent variables, amortized variational inference is commonly used. For methods representing visual concepts with embeddings, the Expectation Maximization (EM) algorithm is adopted if layers are composited with spatial mixture models; otherwise, the parameters of the inference network are directly optimized by minimizing the reconstruction errors.

\subsubsection{Amortized Variational Inference (VI)}

The goal of amortized variational inference is to estimate the posterior distribution $p(\boldsymbol{z}_{0:K}|\boldsymbol{x})$ of latent variables $\boldsymbol{z}_{0:K}$ using a neural network that takes the observed visual scene image $\boldsymbol{x}$ as input. The estimated posterior distribution is denoted as $q(\boldsymbol{z}_{0:K}|\boldsymbol{x})$ and is usually referred to as the variational distribution. The encoder network used for computing parameters of $q(\boldsymbol{z}_{0:K}|\boldsymbol{x})$ and the decoder network used for decoding $\boldsymbol{z}_{0:K}$ are jointly trained by optimizing the following expression.
\begin{equation}
    \label{equ:elbo}
    \mathcal{L}_{\text{elbo}} \!=\! \mathbb{E}_{q(\boldsymbol{z}_{0:K}|\boldsymbol{x})}\big[\log{p(\boldsymbol{x}|\boldsymbol{z}_{0:K})}\big] - D_{\text{KL}}\big(q(\boldsymbol{z}_{0:K}|\boldsymbol{x})||p(\boldsymbol{z}_{0:K})\big)
\end{equation}
In Eq. \eqref{equ:elbo}, $\mathbb{E}_{q(\boldsymbol{z}_{0:K}|\boldsymbol{x})}\big[\log{p(\boldsymbol{x}|\boldsymbol{z}_{0:K})}\big]$ is the reconstruction term that encourages latent variables $\boldsymbol{z}_{0:K}$ of all visual concepts to lead to accurate reconstruction of the visual scene image $\boldsymbol{x}$, and $D_{\text{KL}}\big(q(\boldsymbol{z}_{0:K}|\boldsymbol{x})||p(\boldsymbol{z}_{0:K})\big)$ is the Kullback-Leibler (KL) divergence term that regularizes the aggregated posteriors $\mathbb{E}_{p_{\text{data}}(\boldsymbol{x})}\big[q(\boldsymbol{z}_{0:K}|\boldsymbol{x})\big]$ of all training images to be close to the prior distribution $p(\boldsymbol{z}_{0:K})$. The most common practice is to transform the observed image $\boldsymbol{x}$ into parameters of $q(\boldsymbol{z}_{0:K}|\boldsymbol{x})$ in a feed-forward way and directly optimize Eq. \eqref{equ:elbo}. It is also possible to apply iterative amortized inference \cite{Marino2018Iterative} to improve the inference accuracy or use generalized ELBO with constrained optimization (GECO) \cite{Rezende2018Taming} to improve the quality of unconditional image generation.

\subsubsection{Expectation Maximization (EM)}

In spatial mixture models where visual concepts are not represented with latent variables, compositional scene representations $\boldsymbol{z}_{0:K}$ can be seen as parameters of the distribution $p(\boldsymbol{x}; \boldsymbol{z}_{0:K}) \!=\! \prod_{n=1}^{N}{\sum_{k=0}^{K}{p(l_n \!=\! k; \boldsymbol{z}_{0:K}) p(\boldsymbol{x}_n|l_n \!=\! k; \boldsymbol{z}_{0:K})}}$. Due to the multimodality of $p(\boldsymbol{x}; \boldsymbol{z}_{0:K})$, it is hard to infer $\boldsymbol{z}_{0:K}$ by directly maximizing $\log{p(\boldsymbol{x}; \boldsymbol{z}_{0:K})}$. A commonly used solution is to iteratively maximize the Q-function $\mathcal{Q}(\boldsymbol{z}_{0:K}, \boldsymbol{z}_{0:K}^{\text{old}})$, which is computed by
\begin{equation}
    \label{equ:q_function}
    \mathcal{Q} = \sum_{n=1}^{N}{\sum_{k=0}^{K}{\underbrace{p(l_n \!=\! k|\boldsymbol{x}; \boldsymbol{z}_{0:K}^{\text{old}})}_{\gamma_{k,n}} \log{p(\boldsymbol{x}_n, l_n\!=\!k; \boldsymbol{z}_{0:K})}}}
\end{equation}
The optimization of Eq. \eqref{equ:q_function} consists of two alternating steps, i.e., the expectation step (E-step) that estimates the posteriors $\boldsymbol{\gamma}_{0:K}$ and the maximization step (M-step) that finds the representations $\boldsymbol{z}_{0:K}$ that maximizes the Q-function. Due to the non-linearities of the neural networks used to transform $\boldsymbol{z}_{0:K}$ into parameters of $p(\boldsymbol{x}_n, l_n; \boldsymbol{z}_{0:K})$, finding the optimal $\boldsymbol{z}_{0:K}$ in the M-step is usually intractable. N-EM \cite{Greff2017Neural} provides a solution that uses neural networks to update $\boldsymbol{z}_{0:K}$ based on the value of the Q-function in the M-step, and empirical results have verified its effectiveness.

\subsubsection{Reconstruction Error (RE)}

A commonly used metric to measure the reconstruction error is the mean squared error. In this case, the objective of the learning is to minimize the following expression.
\begin{equation}
    \label{equ:mse}
    \mathcal{L}_{\text{re}} = \frac{1}{N} \sum_{n=1}^{N}{\bigg\lVert \boldsymbol{x}_n - \sum_{k=0}^{K}{\pi_{k,n} \boldsymbol{a}_{k,n}} \bigg\rVert_2^2}
\end{equation}
The minimization of Eq. \eqref{equ:mse} also has a probabilistic interpretation. By treating compositional scene representations $\boldsymbol{z}_{0:K}$ as parameters of the probability $p(\boldsymbol{x}; \boldsymbol{z}_{0:K})$ and assuming that each pixel $\boldsymbol{x}_n$ is independently distributed according to a normal distribution with $\sum_{k=0}^{K}{\pi_{k,n} \boldsymbol{a}_{k,n}}$ as the mean vector and $\sigma_\text{x} \boldsymbol{I}$ as the constant covariance matrix, minimizing Eq. \eqref{equ:mse} is equivalent to maximizing $\log{p(\boldsymbol{x}; \boldsymbol{z}_{0:K})}$.

\subsection{Attention Mechanisms [A]}

Attention mechanisms can be roughly categorized as rectangular attention and arbitrary-shaped attention. The former is easier to estimate and can be naturally combined with prior knowledge of the distribution of positions and scales of objects. The latter can better suppress information contained in irrelevant regions if attention masks are correctly estimated and can be trained more easily because the support of the derivative with respect to attention masks is the entire image instead of local regions. In general, if the prior knowledge of positions and scales of objects is available and approximately correct, and visual scenes are simple enough to be learned well in relatively few training steps, then rectangular attention is preferred over arbitrary-shaped attention. Otherwise, the latter is preferable because they may benefit more from more training steps.

\subsubsection{Rectangular Attention (R)}

\begin{figure}[t]
	\centering
	\subfloat[Computing bounding boxes based on global features]{\includegraphics[width=0.98\columnwidth]{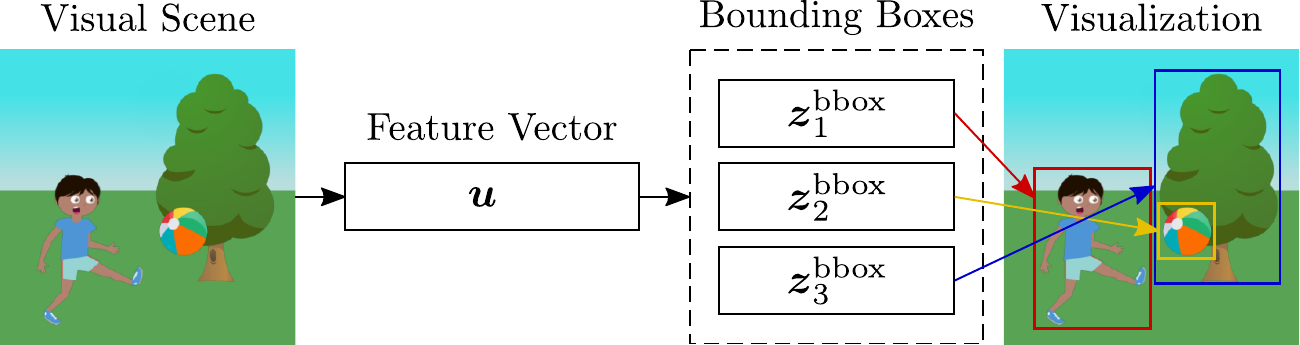}%
		\label{fig:attention_rg}}
	\\
	\subfloat[Computing bounding boxes based on local features]{\includegraphics[width=0.98\columnwidth]{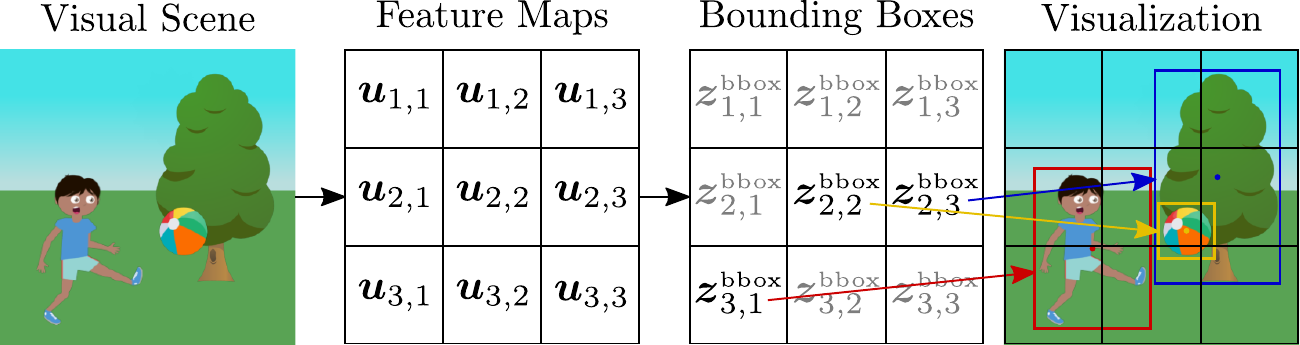}%
		\label{fig:attention_rl}}
	\caption{Two types of rectangular attention.}
	\label{fig:attention_rectangular}
\end{figure}

Methods employing rectangular attention explicitly model parameters of bounding boxes of objects with variables $\boldsymbol{z}_{1:K}^{\text{bbox}}$ or $\boldsymbol{z}_{1:I,1:J}^{\text{bbox}}$. For each object, the parameters of the bounding box are first inferred. Then, the local region in the bounding box is cropped and used to estimate the variable $\boldsymbol{z}_{k}^{\text{attr}}$ or $\boldsymbol{z}_{i,j}^{\text{attr}}$ that characterizes intrinsic attributes of the object (e.g., shape and appearance in the canonical coordinate). Existing methods can be further divided into two categories based on the features used to estimate bounding boxes. Methods of one type use features of the entire visual scenes, while methods of the other type divide a visual scene image into multiple local regions and estimate bounding boxes located in local regions based on local features. The main ideas of these two types of attention are shown in Fig. \ref{fig:attention_rectangular}.

\begin{figure}[t]
	\centering
	\subfloat[Computing attention masks based on network outputs]{\includegraphics[width=0.98\columnwidth]{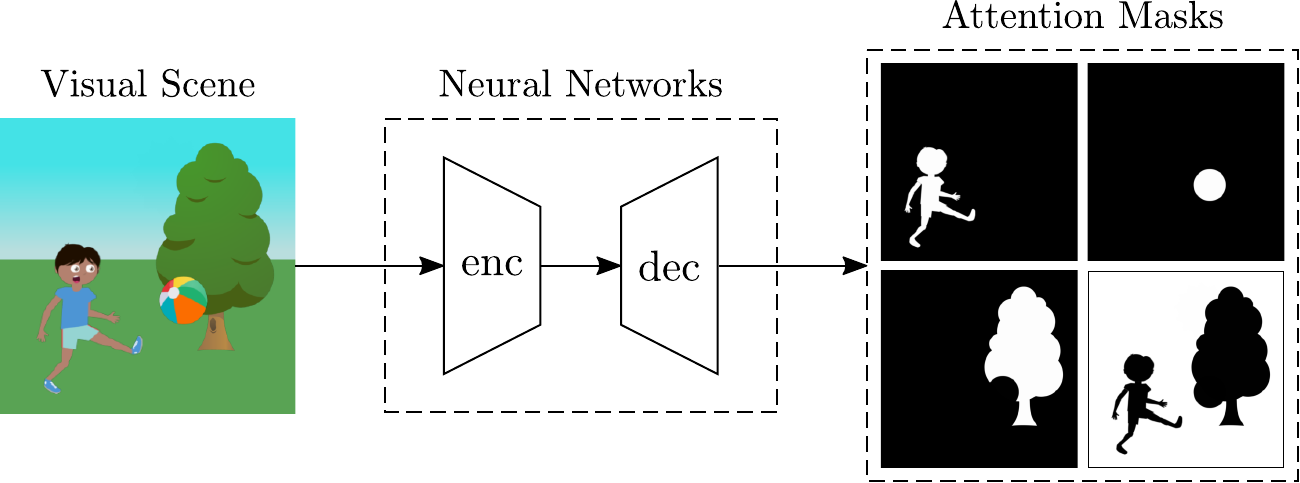}%
		\label{fig:attention_an}}
	\\
	\subfloat[Computing attention masks based on feature similarities]{\includegraphics[width=0.98\columnwidth]{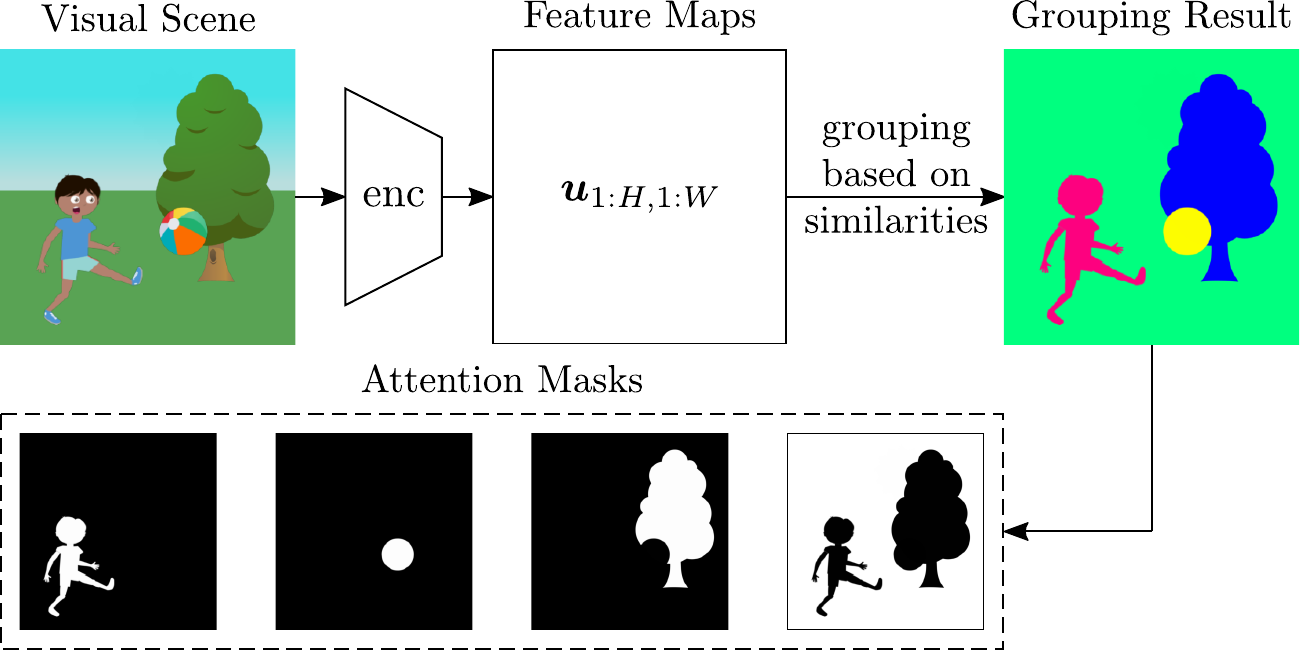}%
		\label{fig:attention_af}}
	\caption{Two types of arbitrary-shaped attention.}
	\label{fig:attention_arbitrary}
\end{figure}

\textbf{Global Features (G):} In methods such as CST-VAE \cite{Huang2016Efficient} and AIR \cite{Eslami2016Attend}, variables $\boldsymbol{z}_{k}^{\text{bbox}}$ and $\boldsymbol{z}_{k}^{\text{attr}}$ of each object are inferred alternately, and the variational distribution is factorized according to the following expression.
\begin{equation}
    \label{equ:infer_bbox_global}
    q(\boldsymbol{z}^{\text{bbox}}, \boldsymbol{z}^{\text{attr}}) \!=\!\! \prod_{k=1}^{K}\!{q(\boldsymbol{z}_{k}^{\text{bbox}}|\boldsymbol{z}_{1:k-1}^{\text{bbox}}, \boldsymbol{z}_{1:k-1}^{\text{attr}}) q(\boldsymbol{z}_{k}^{\text{attr}}|\boldsymbol{z}_{k}^{\text{bbox}})}
\end{equation}
For each object, parameters of $q(\boldsymbol{z}_{k}^{\text{bbox}}|\boldsymbol{z}_{1:k-1}^{\text{bbox}}, \boldsymbol{z}_{1:k-1}^{\text{attr}})$ can be estimated either based on the global features of the image that is initialized as the observed image $\boldsymbol{x}$ and iteratively updated according to the previously inferred variables $\boldsymbol{z}_{1:k-1}^{\text{bbox}}$ and $\boldsymbol{z}_{1:k-1}^{\text{attr}}$ or based on the hidden states of an RNN that iteratively integrate the global features of the observed image $\boldsymbol{x}$ with $\boldsymbol{z}_{1:k-1}^{\text{bbox}}$ and $\boldsymbol{z}_{1:k-1}^{\text{attr}}$.

The main advantage of this attention choice is flexibility. Representations of at most $K$ objects can be inferred regardless of the spatial relationships of these objects, and the computing power is adaptively allocated based on the locations of objects, i.e., regions containing more objects consume more computing power because encoder and decoder networks are executed more times in these regions. Although the usage of global features does not fully exploit the spatial invariance of objects, in cases where objects may be heavily occluded or most regions of the visual scene do not contain any object, it is recommended to estimate bounding boxes based on global features rather than local features because of significantly lower computational complexity.

\textbf{Local Features (L):} In methods such as SPAIR \cite{Crawford2019Spatially} and SPACE \cite{Lin2020SPACE}, each visual scene image is divided into $I \times J$ regions. Variables $\boldsymbol{z}_{1:I,1:J}^{\text{bbox}}$ that parameterize bounding boxes of objects are inferred based on the local features of these regions. After inferring bounding boxes for all regions, variables $\boldsymbol{z}_{1:I,1:J}^{\text{attr}}$ that characterize the intrinsic attributes of objects are extracted. Let $[(i_1, j_1), (i_2, j_2), \dots, (i_{I \times J}, j_{I \times J})]$ denote the predefined index sequence of $I \times J$ regions, and let $\mathcal{S}_k$ denote a subset of $\{1, 2, \dots, k - 1\}$. The factorization of the variational distribution is described below.
\begin{equation}
    \label{equ:infer_bbox_local}
    q(\boldsymbol{z}^{\text{bbox}}, \boldsymbol{z}^{\text{attr}}) = \prod_{k=1}^{I \times J}{q(\boldsymbol{z}_{i_k,j_k}^{\text{bbox}}|\boldsymbol{z}_{(\boldsymbol{i},\boldsymbol{j})_{\mathcal{S}_k}}^{\text{bbox}}) q(\boldsymbol{z}_{i_k,j_k}^{\text{attr}}|\boldsymbol{z}_{i_k,j_k}^{\text{bbox}})}
\end{equation}
In methods like SPAIR \cite{Crawford2019Spatially}, $\mathcal{S}_k = \{1, 2, \dots, k - 1\} \cap \{k': (i_{k'}, j_{k'}) \text{ is a neighbor of } (i_{k}, j_{k})\}$. The inference of $\boldsymbol{z}_{i_k,j_k}^{\text{bbox}}$ depends on the estimations in nearby regions that have already been inferred, and each $\boldsymbol{z}_{i_k,j_k}^{\text{bbox}}$ is inferred sequentially. In methods like SPACE \cite{Lin2020SPACE}, $\mathcal{S}_k = \varnothing$, i.e., the estimations in all regions are independent, and all the variables $\boldsymbol{z}_{i_k,j_k}^{\text{bbox}}$ can be inferred in parallel. It is worth mentioning that each region is usually assigned a binary variable $z_{i,j}^{\text{pres}}$ indicating the presence of an object in that region because some regions may not contain any object.

Methods using local features restrict the search space of bounding boxes and can better model the spatial invariance of objects. Therefore, they are more suitable for learning from high-resolution images containing lots of objects that do not overlap much. There is a simple strategy to combine the advantages of the uses of global and local features. By coarsely partitioning the image into sub-images and modeling each sub-image with Eq. \eqref{equ:infer_bbox_global}, better spatial invariance can be achieved while retaining enough flexibility to allocate more computing power to regions containing more objects.

\subsubsection{Arbitrary-Shaped Attention (A)}
\label{sec:arbitrary_shaped_attention}

Methods using arbitrary-shaped attention can be further classified into two categories based on the computation of attention masks. Methods in one category compute attention masks based on outputs of decoder networks, while methods in the other category compute attention masks based on similarities of local features extracted by encoder networks. The main ideas of these two types are shown in Fig. \ref{fig:attention_arbitrary}.

\textbf{Network Outputs (N):} In methods such as MONet \cite{Burgess2019MONet}, ViMON \cite{Weis2021Benchmarking}, and POD-Net \cite{Du2021Unsupervised}, arbitrary-shaped attention masks are estimated using the combination of encoder and decoder networks like U-Net \cite{Ronneberger2015U}. In methods like GENESIS \cite{Engelcke2020GENESIS}, the representations that characterize the shape and appearance attributes of visual concepts are inferred separately, and the decoded perceived shapes serve as attention masks for inferring appearance representations. In methods like IODINE \cite{Greff2019Multi}, compositional scene representations are iteratively refined. The perceived shapes computed based on the representations inferred in the previous step are used as one of the auxiliary inputs and act as attention masks in the inference of representations in the current step. In methods like N-EM \cite{Greff2017Neural} and LDP \cite{Yuan2019Spatial}, which iteratively infer compositional scene representations based on the EM algorithm, the posteriors $\gamma_{k,n} = p(l_n \!=\! k|\boldsymbol{x}; \boldsymbol{z}^{\text{old}})$ in Eq. \eqref{equ:q_function} can be seen as attention masks of different visual concepts.

The main advantage of this attention choice is simplicity, i.e., attention masks can be computed in the encoding-decoding framework, which is also used to reconstruct visual scene images. The main drawback is caused by the notorious black-box nature of deep neural networks, i.e., the estimation of attention masks in this way lacks interpretability. This attention choice is a reasonable solution if neural networks with suitable inductive biases can be designed to compute attention masks.

\textbf{Feature Similarities (F):} In methods such as TBA \cite{He2019Tracking} and Slot Attention \cite{Locatello2020Object}, representations of visual concepts are estimated based on a weighted average of feature maps. The weights to perform the weighted average are computed based on the similarity among local features on the feature maps and can be seen as attention masks. In methods like GENESIS-V2 \cite{Engelcke2021GENESIS}, attention masks are computed in a way inspired by instance coloring, which has been previously used in supervised instance segmentation. The attention mask of each visual concept is sequentially computed based on similarities between the local feature at a randomly sampled position and all the other local features. In methods like PSGNet \cite{Bear2020Learning}, feature maps are extracted in multiple levels, and label propagation is used to group regions recursively from lower to higher levels. The grouping results are used as attention maps to compute features of the grouped regions.

Compared with network outputs, using feature similarities improves the interpretability of attention masks because this attention choice implies that regions with similar features are more likely to belong to the same visual concept. Furthermore, this attention choice is more related to traditional unsupervised image segmentation, which has been studied extensively for a long time before the era of deep learning and has many successful techniques in it. The major problem in applying traditional techniques is propagating gradients from attention masks to features. If this problem can be solved without significantly increasing the learning difficulty (e.g., introducing many discrete variables that are hard to optimize), computing attention masks based on feature similarities would be the top recommendation because many useful inductive biases can be added.

\section{Benchmarks}

Reconstruction-based compositional scene representation learning with deep neural networks is still in the early stages of research. Different methods usually use different datasets and evaluation metrics to conduct experiments, either due to the lack of consensus benchmarks or because these methods are proposed for different problem settings. This section benchmarks the inference performance of representative methods that learn compositional representations of static visual scenes from a single viewpoint. This problem setting is the most extensively studied one and forms the foundation for more complex problem settings considering object interactions, object motions\footnote{For benchmarks of representative methods that learn from dynamic visual scenes, please refer to Weis et al. \cite{Weis2021Benchmarking} for more details.}, or multiple viewpoints of visual scenes. We reimplement or unify the I/O interfaces of $10$ representative methods in this problem setting. The code for creating datasets and evaluating the performance of the chosen methods is publicly available\footnote{An open-source toolbox is provided to reproduce the benchmark experiments. Code is available at \url{https://tinyurl.com/5bhb6nt3}.}. The complexities of these methods are analyzed in the Supplementary Material. Simply put, the number of network parameters does not have a strong correlation with the computational complexity. Methods employing rectangular attention usually consume less GPU memory and have lower computational complexities. Methods first randomly initializing compositional scene representation and then iteratively refining them using the information in the pixel space are more memory-intensive and computationally intensive. In the following, we first introduce the datasets used for benchmarking, then describe the evaluation metrics, and finally compare the performance of the chosen methods.

\subsection{Datasets}

\input{table_data}

Six datasets are used to evaluate the performance of different methods. These datasets are constructed based on the MNIST \cite{Lecun1998Gradient}, dSprites \cite{Matthey2017dSprites}, Abstract Scene \cite{Zitnick2013Bringing}, CLEVR \cite{Johnson2017CLEVR}, SHOP VRB \cite{Nazarczuk2020SHOP}, and the combination of GSO \cite{Downs2022Google} and HDRI-Haven datasets. For brevity, they are referred to as the \textit{MNIST}, \textit{dSprites}, \textit{AbsScene}, \textit{CLEVR}, \textit{SHOP}, and \textit{GSO} datasets, respectively. These datasets are constructed similarly to the Multi-Objects dataset \cite{Kabra2019Multi} and additionally provide annotations of complete shapes of objects so that it is possible to assess the performance with more evaluation metrics. The configurations used to construct these datasets are described in Table \ref{tab:datasets}. More details are provided in the Supplementary Material.

\subsection{Metrics}

\input{table_test_average}

Six metrics are used to quantitatively assess the performance from four aspects. The segmentation performance is measured by Adjusted Mutual Information (AMI) \cite{Vinh2010Information} and Adjusted Rand Index (ARI) \cite{Hubert1985Comparing}. The amodal segmentation performance is evaluated by Intersection over Union (IoU) and $F_1$ score (F1). The object counting performance and the object ordering performance are assessed by Object Counting Accuracy (OCA) and Object Ordering Accuracy (OOA), respectively. Segmentation is the most general aspect where all the methods can be evaluated and is strongly correlated with the usefulness of compositional scene representations for downstream tasks like property prediction \cite{Papa2022Inductive}. Two variants of AMI and ARI are used to evaluate the segmentation performance more thoroughly. AMI-A and ARI-A are computed using pixels in the entire image and measure how accurately different layers of visual concepts (including both objects and the background) are separated. AMI-O and ARI-O are computed only using pixels in the regions of objects and focus on how accurately different objects are separated. The remaining three aspects, i.e., amodal segmentation, object counting, and object ordering, have more requirements for the inference results and are not applicable to all the methods.
Reconstruction error is not chosen for performance comparison because it has been shown that the performance on downstream tasks is only weakly correlated with reconstruction quality \cite{Papa2022Inductive}.
Detailed descriptions of the metrics are included in the Supplementary Material.

\subsection{Performance}

The performance averaged across the six datasets is shown in Table \ref{tab:test_average}. In general, GMIOO \cite{Yuan2019Generative} and MONet \cite{Burgess2019MONet} perform best and second best, respectively. Detailed results and analyses are included in the Supplementary Material. Because AIR \cite{Eslami2016Attend} does not model the shapes of objects, and N-EM \cite{Greff2017Neural}, IODINE \cite{Greff2019Multi}, MONet \cite{Burgess2019MONet}, Slot Attention \cite{Locatello2020Object}, EfficientMORL \cite{Emami2021Efficient}, and GENESIS-V2 \cite{Engelcke2021GENESIS} do not model the complete shapes of objects, the IoU and F1 scores are not evaluated for them. Because N-EM \cite{Greff2017Neural}, IODINE \cite{Greff2019Multi}, Slot Attention \cite{Locatello2020Object}, and EfficientMORL \cite{Emami2021Efficient} model visual scenes and infer scene representation both in equivariant ways (i.e., visual concepts are unordered), OOA scores are not evaluated for them. Although N-EM \cite{Greff2017Neural}, IODINE \cite{Greff2019Multi}, MONet \cite{Burgess2019MONet}, GENESIS \cite{Engelcke2020GENESIS}, Slot Attention \cite{Locatello2020Object}, EfficientMORL \cite{Emami2021Efficient}, and GENESIS-V2 \cite{Engelcke2021GENESIS} do not model the number of objects, objects can still be heuristically counted based on the segmentation results. The estimated number of objects is considered to be the number of segmented regions for N-EM \cite{Greff2017Neural} (because the background does not have a dedicated segmented region) and the number of segmented regions minus one for the other methods.

\section{Limitations of Existing Methods}
\label{sec:limitations}

With years of research, much progress has been made in reconstruction-based compositional scene representation learning with deep neural networks. However, existing methods still suffer from some limitations that hinder their usefulness in practical applications. The three main limitations are described below.

\textbf{Hierarchies of Compositional Scene Representations:}
A notable characteristic of visual scenes is the rich hierarchical structure. The hierarchy of scene representations not only allows for better exploitation of the compositional nature of visual scenes by compositing at finer-grained levels but also allows rich relationships among visual concepts to be represented in a suitable form. Hierarchies of scene representations have been widely adopted in traditional compositional scene representation learning methods. However, only very few ``deep'' methods (e.g., GSGN \cite{Deng2021Generative}) have considered hierarchical modeling, and the effectiveness is only verified on simple synthetic images where the hierarchical structures of visual concepts are artificial.

\textbf{Ability to Continuously Learn New Visual Concepts:}
The world is ever-changing, and new visual concepts are constantly appearing. It is desirable for compositional scene representation learning methods to support continual learning of new visual concepts and are thus capable of continuously improving the capabilities of the models. However, learning in the continual setting has been little studied in reconstruction-based compositional scene representation learning with deep neural networks. As explained in Section \ref{sec:prototypes}, although using prototype images as representations naturally supports continual learning, this representation choice has drawbacks as it is not suitable for situations where the number of object categories or the dimensionality of images is large. ADI \cite{Yuan2021Knowledge} theoretically does not suffer from this problem, but its effectiveness is only verified in a simple setting involving two learning stages.

\textbf{Applicability to Complex Real-World Visual Scenes:}
Existing methods usually use synthetic images to conduct experiments. Therefore, the effectiveness is mostly verified on relatively simple visual scenes, i.e., under conditions such as visual concepts are not difficult to reconstruct, the diversity of visual concepts is not great, and the lighting is simple and controlled. As for real-world visual scenes where lighting is not controlled and visual concepts are complex and diverse, almost all the existing methods do not perform well without modification \cite{Yafei2022Promising}. Some rare exceptions are the recently proposed BO-QSA \cite{Jia2023Unsupervised} and DINOSAUR \cite{Seitzer2023Bridging}, which have achieved encouraging unsupervised foreground extraction, image segmentation, and object discovery results on real-world images when using Transformer \cite{Vaswani2017Attention} as the decoder network. However, these two methods still suffer from problems such as inaccurate segmentation in the boundary regions of visual concepts. More progress is needed to exploit the potential of compositional scene representations further on complex real-world visual scenes.

\section{Future Directions}

There are three main aspects of future directions. Firstly, progress should be made in the study of representation learning methods themselves, thereby mitigating and eventually addressing the limitations mentioned in the previous section. Secondly, applications of compositional scene representations need to be investigated because an important purpose of learning suitable representations is to achieve better performance on downstream tasks. Last but not least, compositional scene representation learning can be combined with other tasks, thereby learning representations with better properties or higher qualities.

\subsection{Improvements in Representation Learning}

Some possible improvements are briefly described below. The first three correspond to aspects in Sections \ref{sec:modeling} and \ref{sec:inference} that have much room for improvement, and the remaining five relate to the limitations described in Section \ref{sec:limitations}.

\textbf{Class-Aware Representations with Rich Variability:}
As mentioned in Section \ref{sec:object_representations}, compared with using representations in real vector space or a finite number of prototypes alone, the combination of these two allows for better expressiveness because the learned representations can contain categorical information (class ID without semantic label) while retaining enough flexibility to handle complex intra-class variability. One viable way is to describe categories of objects with a dictionary of prototype vectors and encode intra-class variability with variables in real vector space. GSGN \cite{Deng2021Generative} has made a preliminary exploration and verified the effectiveness of this idea on simple synthetic images. More research is needed to make such representations usable in practice. For example, how to learn relationships (e.g., hierarchies) of categories and how to cluster representations accurately when objects may be heavily occluded.

\textbf{Modeling of Lighting Effects:}
Various properties of visual scenes that can be additionally modeled to increase the quality of learned representations have been described in Section \ref{sec:additional_modelings}. Unlike these properties, lighting effects have hardly been studied. In real-world visual scenes, visual perception is greatly influenced by lighting effects. Even for a visual scene where objects are all static, the scene image may vary significantly from morning to night. Therefore, it is desirable to reduce variability in representations of visual concepts by separately modeling lighting effects. Synthesizing realistic lighting effects has been studied in computer graphics for a long time. Inspiration from this field may be drawn to help the research on compositional scene representation learning.

\textbf{Additional Inductive Biases:}
Existing methods already have various inductive biases to help obtain desired decompositions of visual scenes, e.g., the structure of neural networks, the regularization terms in the loss function, and the heuristics used during inference of scene representations. As the complexity of visual scenes increases, more inductive biases are required to learn well \cite{Yafei2022Promising}. Therefore, the research on additional inductive biases is important. As stated in Section \ref{sec:arbitrary_shaped_attention}, inductive biases used in traditional unsupervised segmentation methods may be applied if employing arbitrary-shaped attention masks and computing these masks based on feature similarities. Moreover, one can also design new learning pipelines (e.g., the one used by Tangemann et al. \cite{Tangemann2021Unsupervised}) that include suitable inductive biases to guide the discovery of objects.

\textbf{Hierarchical Compositional Modeling and Inference:}
To learn compositional scene representations with hierarchical structures, both the modeling of visual scenes and the inference of scene representations need to be modified accordingly, and intensive research is required for them. As for hierarchical compositional modeling, one can either use the widely adopted tree structure to model visual scenes from fine-grained visual concepts to the entire scene or use a more complex and powerful structure (e.g., the And-Or graph used by Zhu et al. \cite{Zhu2007Stochastic}) to model richer relationships among visual concepts at different granularities. As for hierarchical inference, in addition to inferring the representations corresponding to nodes of the hierarchical structure, it is necessary to infer the structure itself. Successful techniques in structural learning may be combined with compositional scene representation learning for hierarchical inference.

\textbf{Continual Learning of Object Representations:}
The core of continual learning is to learn new tasks without forgetting the previously learned ones. The new tasks mentioned here are not restricted to new types of tasks. Supporting new categories or handling changes in data distributions is also considered new because the current models do not perform well on these tasks. Besides the problem of catastrophic forgetting commonly considered in the continual learning of neural networks, learning compositional scene representations in the continual setting faces new challenges, such as accurately decomposing visual scenes in the presence of novel objects. Therefore, there are many research opportunities on this problem.

\textbf{Large Benchmark Datasets of Real-World Scenes:}
Most of the existing methods are only evaluated on synthetic datasets, and the number of data used for training is usually around \num[group-separator={,}]{50000} because this much data are enough for models to learn relatively well. Just as ImageNet has contributed to the rapid development of neural networks for image classification (many of them have subsequently been used as backbones in other computer vision tasks), the research on compositional scene representation learning is expected to benefit significantly from large benchmark datasets of real-world visual scenes. Because compositional scene representations can be learned via reconstruction without supervision, expensive and laborious data labeling is not needed for training data. The main difficulty in creating benchmark datasets is selecting suitable training and testing data, making these datasets widely accepted.

\textbf{Combination with Large Models:}
The appearances and shapes of objects in real-world visual scenes are complex and rich in diversity. To encode sufficient details that might be useful for downstream tasks, neural networks with a large number of parameters are needed. In addition, an important advantage of learning via reconstruction is that massive unlabeled data can be used because no annotations are required during training. Employing large models as encoder and decoder networks is more likely to benefit from more training data and longer training time. Furthermore, efficient fine-tuning methods are actively studied in the field of large models. Successful techniques in this field may inspire better applications of compositional scene representation models to downstream tasks.

\textbf{Generalization from Synthetic to Real-World Scenes:}
Learning compositional scene representations directly from complex real-world scenes is difficult without supervision. A promising approach to lower the learning difficulty is to first learn from synthetic visual scenes and then apply the pre-trained models to continue to learn from more complex real-world scenes. The research problems worth exploring include learning knowledge that can be seamlessly transferred from the synthetic domain to the real domain and designing loss functions that can better exploit the free annotations coming with synthetic visual scenes.

\subsection{Applications}

An appealing property of learning via reconstruction is that if the visual scene image can be perfectly reconstructed based on the extracted representations, then no information is lost in these representations. In addition, it has been shown by Dittadi et al. \cite{Dittadi2022Generalization} that compositional scene representations are advantageous in that they generalize relatively well and are robust to distribution shifts of objects in many cases. Therefore, compositional scene representations learned via reconstruction have great practical potential as long as visual scenes are decomposed in a desired way and the reconstruction quality is relatively high. One viable way of applying these representations is to train a large compositional scene representation model and then use it as the pre-trained model for downstream tasks. Some potentially applicable tasks are briefly described below.

\textbf{Image Classification with Better Interpretability:}
A critical shortcoming of deep neural networks is lacking sufficient interpretability. To alleviate this problem, various visualization methods have been proposed to help understand how decisions are made. Compared to performing image classification based on representations of the entire images, using compositional scene representations leads to better interpretability because images have already been abstracted as the composition of more understandable visual concepts (instead of the collection of pixels). Moreover, the visualization of decisions is naturally supported if one distinguishes whether visual concepts are relevant and classifies images based on relevant visual concepts only.

\textbf{Amodal Panoptic Segmentation:}
The segmentation results obtained by all the reconstruction-based compositional scene representation learning methods are panoptic in nature because different instances of objects are distinguished, and the segmentation masks cover the entire image. If complete shapes of objects are modeled, then the segmentation results are both panoptic and amodal, and occluded parts of objects can be imagined by decoding the inferred representations into reconstructions of complete objects. To fine-tune the pre-trained model on the target dataset, annotations can be used in two ways. Firstly, the mapping from compositional scene representations to segmentation masks and categories can be learned with these annotations as supervision. Secondly, by adding an additional supervision term in the loss function as ADI \cite{Yuan2021Knowledge} has done and continuing to train on the target dataset, higher segmentation accuracy could be achieved.

\textbf{Visual Question Answering and Visual Grounding:}
Compositional scene representations are well-suited for visual question answering because visual scenes have already been decomposed into visual concepts (the identities and attributes of different visual concepts have been distinguished), and complex spatial relationships can be encoded in the hierarchical structure of scene representations. In this way, one can focus on reasoning at the level of visual concepts rather than additionally considering learning from pixels or feature maps. The effectiveness of this idea has been demonstrated under the circumstance that the discovery of objects is performed by an object detector pre-trained in the supervised setting \cite{Yi2018Neural,Chen2021Grounding}. Visual grounding, whose goal is to find correspondences between regions in the image and words or phrases in the sentence describing the image, is closely related to visual question answering. It has been shown that this task also benefits from compositional scene representations learned with the help of object detectors \cite{Mao2019Neuro,Han2019Visual}. With the development of reconstruction-based compositional scene representation learning with deep neural networks, it may be possible to reduce the supervision needed for discovering objects accurately in complex real-world visual scenes. Moreover, by fine-tuning the pre-trained models on these datasets, the additional information in the texts may help to learn better scene representations.

\subsection{Combinations with Other Tasks}
Besides the applications to downstream tasks, compositional scene representation learning can also be combined with other tasks to learn representations with better properties. Two representative tasks are compositional 3D reconstruction and compositional embodied vision.

\textbf{Compositional 3D Reconstruction:}
This task is closely related to the problem of learning compositional scene representations from multiple viewpoints. The main differences between these two are that compositional 3D reconstruction focuses more on the reconstruction quality, requires learning 3D models of objects, and does not necessarily learn from multiple viewpoints. If learning compositional scene representations along with this task, the learned representations will contain full 3D information, and images synthesized from different viewpoints will be guaranteed to be consistent (methods like MulMON \cite{Li2020Learning}, DyMON \cite{Nanbo2021Object}, ROOTS \cite{Chen2021ROOTS}, SIMONe \cite{Kabra2021SIMONe}, and OCLOC \cite{Yuan2022Unsupervised} cannot make such a guarantee). Implicit 3D representations, which have recently received more attention with the introduction of NeRF \cite{Mildenhall2020NeRF}, are well-suited for compositional scene representation learning. A series of improvements and extensions of NeRF, such as NeRF{-}{-} \cite{Wang2021NeRF}, KiloNeRF \cite{Reiser2021KiloNeRF}, GRF \cite{Trevithick2021GRF}, and NeRF-VAE \cite{Kosiorek2021NeRF}, lay the foundation for the combination of compositional scene representation learning and implicit 3D representations. ObSuRF \cite{Stelzner2021Decomposing} and uORF \cite{Yu2022Unsupervised} have made preliminary explorations in this promising direction, and more research progress is expected.

\textbf{Compositional Embodied Vision:}
Compared to passively learning compositional scene representations only from observational data, interacting with the environment via embodied agents allows better discovery of the physical properties of objects, which can be included in compositional scene representations and provide additional information for visual scene understanding. In addition, purposefully changing the environment via embodied agents makes it possible to obtain more targeted information from visual scenes, thereby improving learning efficiency. Moreover, embodied agents provide an efficient way to collect training data autonomously and thus reduce human involvement in data preparation. Some methods, such as COBRA \cite{Watters2019COBRA} and Grasp2Vec \cite{Jang2018Grasp2Vec}, have preliminarily explored this direction. Future work may include imitating the development of human infants by combining compositional scene representation learning and embodied vision for learning visual perception and motor skills simultaneously.

\section{Conclusions}

Reconstruction-based compositional scene representation learning with deep neural networks has emerged in the last few years and gradually gained more attention due to its research significance. This survey has introduced the problem setting and development history of this research topic. Existing methods have been categorized in terms of the modeling of visual scenes and the inference of scene representations. Representative methods that consider the most extensively studied problem setting and form the foundation for other methods have been benchmarked on six datasets. An open-source toolbox that includes the code for creating datasets and evaluating the performance of the chosen methods has been provided to reproduce the benchmark experiments. Although existing methods still have limitations in some aspects, reconstruction-based compositional scene representation learning with deep neural networks has broad research prospects. It is believed that this research topic will continue to gain more attention in the coming years, and continuous research progresses in this topic will promote the development of more human-like artificial intelligence systems.

\ifCLASSOPTIONcaptionsoff
  \newpage
\fi



\bibliographystyle{IEEEtran}
\bibliography{reference}
%

%

\vspace{-0.22in}

\begin{IEEEbiography}[{\includegraphics[width=1in,height=1.25in,clip,keepaspectratio]{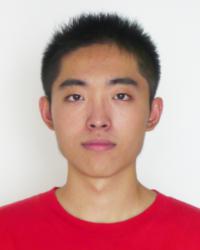}}]{Jinyang Yuan}
received the BS degree in physics from Nanjing University, China, the MS degree in electrical engineering from University of California, San Diego. and the PhD degree in computer science from Fudan University, China. His research interests include computer vision, machine learning, and deep generative models.
\end{IEEEbiography}

\vspace{-0.22in}

\begin{IEEEbiography}[{\includegraphics[width=1in,height=1.25in,clip,keepaspectratio]{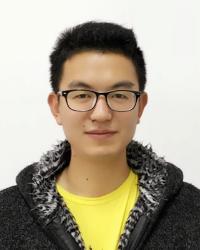}}]{Tonglin Chen}
is currently pursuing the PhD degree in computer science from Fudan University, Shanghai, China. His current research interests include machine learning, deep generative models, and object-centric representation learning.
\end{IEEEbiography}

\vspace{-0.22in}

\begin{IEEEbiography}[{\includegraphics[width=1in,height=1.25in,clip,keepaspectratio]{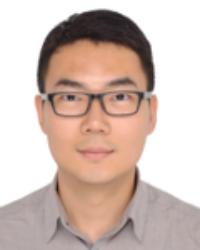}}]{Bin Li}
received the PhD degree in computer science from Fudan University, Shanghai, China. He is an associate professor with the School of Computer Science, Fudan University, Shanghai, China. Before joining Fudan University, Shanghai, China, he was a lecturer with the University of Technology Sydney, Australia and a senior research scientist with Data61 (formerly NICTA), CSIRO, Australia. His current research interests include machine learning and visual intelligence, particularly in compositional scene representation, modeling and inference.
\end{IEEEbiography}

\vspace{-0.22in}

\begin{IEEEbiography}[{\includegraphics[width=1in,height=1.25in,clip,keepaspectratio]{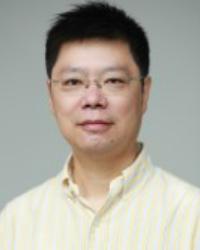}}]{Xiangyang Xue}
received the BS, MS, and PhD
degrees in communication engineering from
Xidian University, Xian, China, in 1989, 1992,
and 1995, respectively. He is currently a professor of computer science with Fudan University,
Shanghai, China. His research interests include
multimedia information processing and machine
learning.
\end{IEEEbiography}





\vfill



\clearpage

\begin{appendices}
	
\section{Advantage in Terms of Informativeness}

\input{table_recon}

As shown by Geman et al. \cite{Geman2002Composition}, compositional scene representations can be learned based on Rissanen's Minimum Description Length (MDL) principle \cite{Rissanen1998Stochastic} because more compact representations can be obtained if visual scenes are correctly decomposed. This finding suggests that, even without considering a large number of potential downstream tasks, compositional scene representations are advantageous in terms of informativeness.

To verify that the advantage of informativeness also applies to reconstruction-based compositional scene representation learning methods that use deep neural networks as encoders and decoders, the reconstruction quality of the methods chosen for benchmarking is compared with their non-compositional versions. The non-compositional versions are modified based on the original compositional versions by changing the number of decomposed layers and the dimensionalities of representations, i.e., these two versions use the same network structure\footnote{Because the quality of reconstructed images is largely affected by network structures, the comparison is made between different versions of the same method rather than between different methods.} and differ only in the modeling of visual scenes. For methods that use the same neural network for the background and objects, the word ``non-compositional'' means that a single representation is learned for the entire visual scene. For methods that use different neural networks for the background and objects, the word ``non-compositional'' means that objects are not modeled compositionally, i.e., each visual scene is only decomposed into two layers, one for the background and the other for all objects in the visual scene. We do not use the fully non-compositional versions (only one layer) because only one of the object and background decoder networks will be used, resulting in an unfair comparison (the non-compositional versions will have significantly fewer network parameters than the compositional versions). For the sake of fairness, the overall lengths of the distributed representations \cite{Hinton1986Distributed} extracted by the compositional and non-compositional versions are identical. The non-distributed representations include variables describing the presence and bounding boxes of objects. The length of non-distributed representations is determined by the modeling of visual scenes and cannot be adjusted freely. The non-compositional versions of methods are obtained in the following way:
\begin{itemize}[leftmargin=*]
	\item For N-EM \cite{Greff2017Neural}, IODINE \cite{Greff2019Multi}, MONet \cite{Burgess2019MONet}, GENESIS \cite{Engelcke2020GENESIS}, Slot Attention \cite{Locatello2020Object}, EfficientMORL \cite{Emami2021Efficient}, and GENESIS-V2 \cite{Engelcke2021GENESIS}, which do not use different neural networks for the background and objects (also do not use non-distributed representations), their non-compositional versions are obtained by \textbf{1)} setting the number of layers to $1$ (non-compositional modeling of visual scenes); and \textbf{2)} multiplying the dimensionality of the representation of each layer by the original number of layers $K + 1$ (maintain the same overall representation length).
	\item For AIR \cite{Eslami2016Attend}, GMIOO \cite{Yuan2019Generative}, and SPACE \cite{Lin2020SPACE}, which use different neural networks for the background and objects (representations of objects contain non-distributed parts, i.e., the variables indicating the presences of objects and the variables parameterizing the bounding boxes of objects), the non-compositional versions are obtained by \textbf{1)} setting the number of object layers from $K$ to $1$ (non-compositional modeling of objects); and \textbf{2)} multiplying the dimensionality of the distributed part of object representations by $K$ (maintain the same overall length of distributed representations).
\end{itemize}

\begin{figure*}[p]
	\centering
	\captionsetup[subfloat]{farskip=2pt,captionskip=2pt}
	\subfloat[The MNIST dataset]{\includegraphics[width=0.23\columnwidth,valign=c]{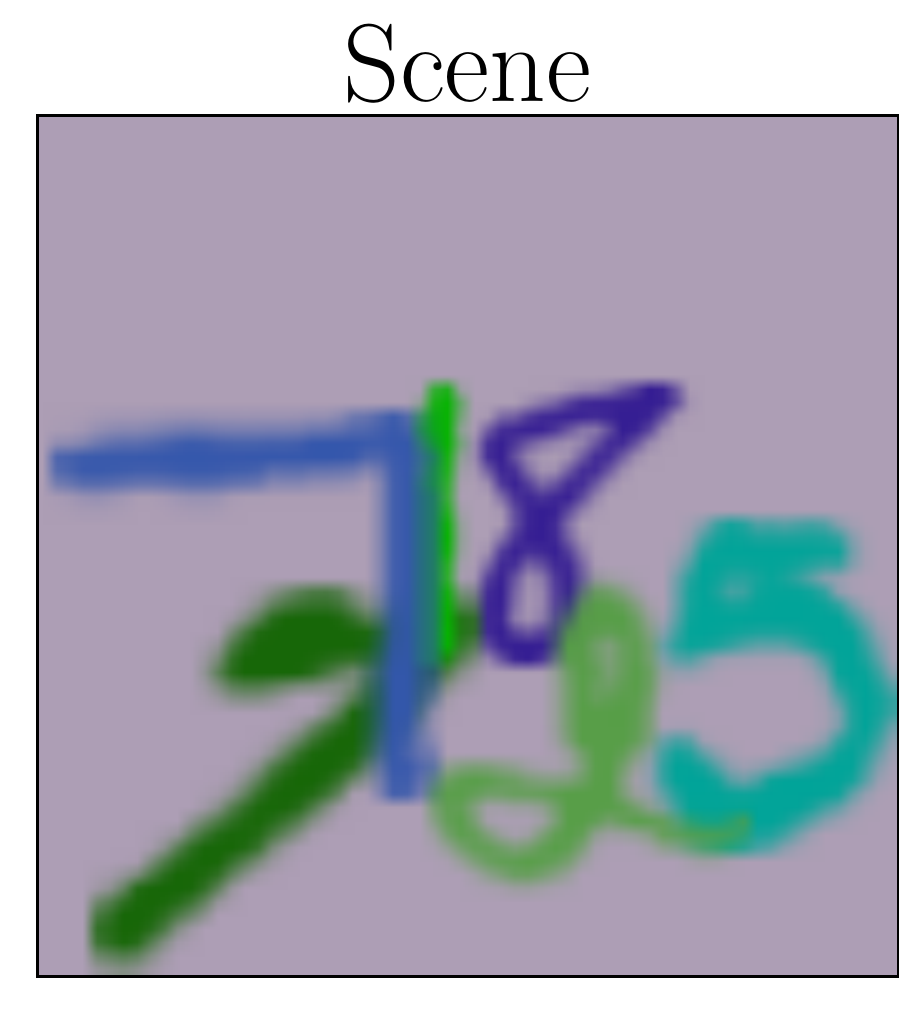}\includegraphics[width=1.8\columnwidth,valign=c]{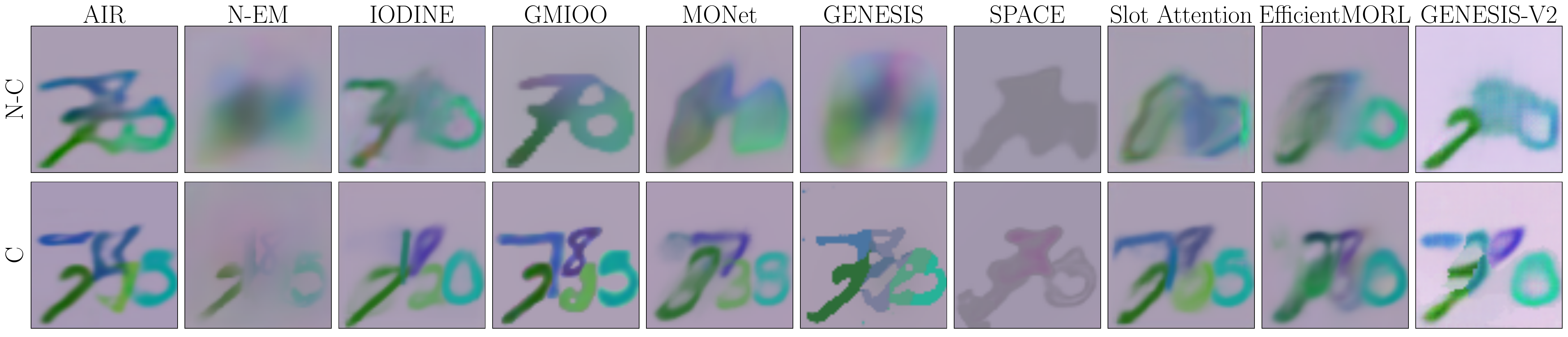}%
		\label{fig:recon_mnist}}
	\\
	\subfloat[The dSprites dataset]{\includegraphics[width=0.23\columnwidth,valign=c]{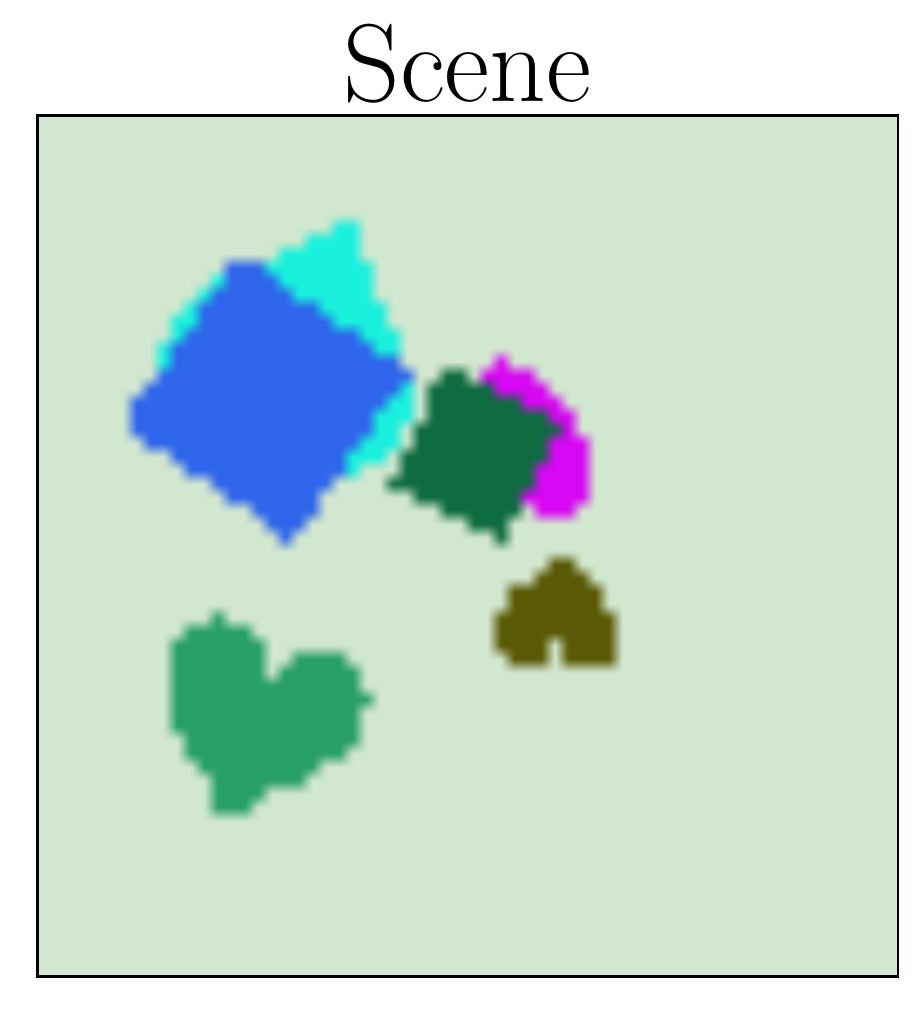}\includegraphics[width=1.8\columnwidth,valign=c]{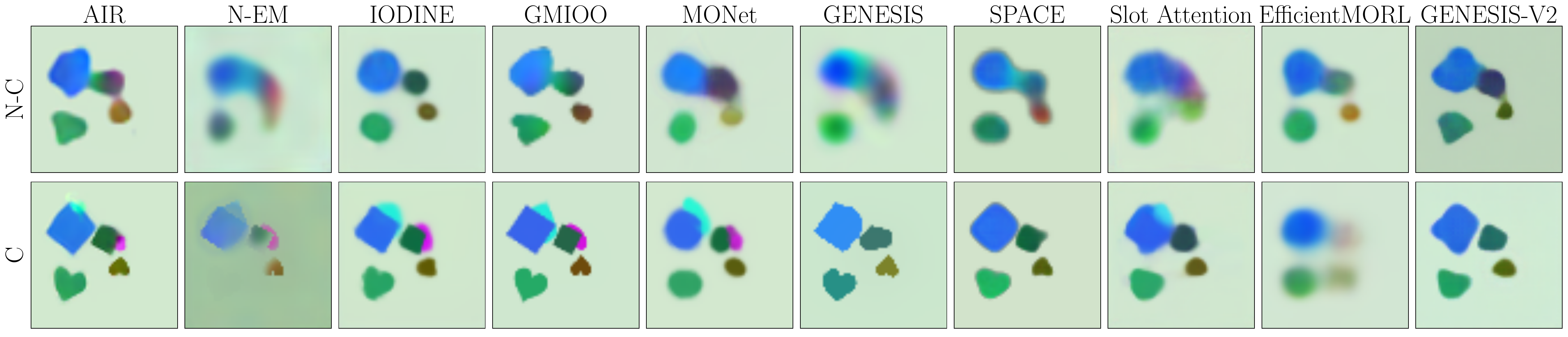}%
		\label{fig:recon_dsprites}}
	\\
	\subfloat[The AbsScene dataset]{\includegraphics[width=0.23\columnwidth,valign=c]{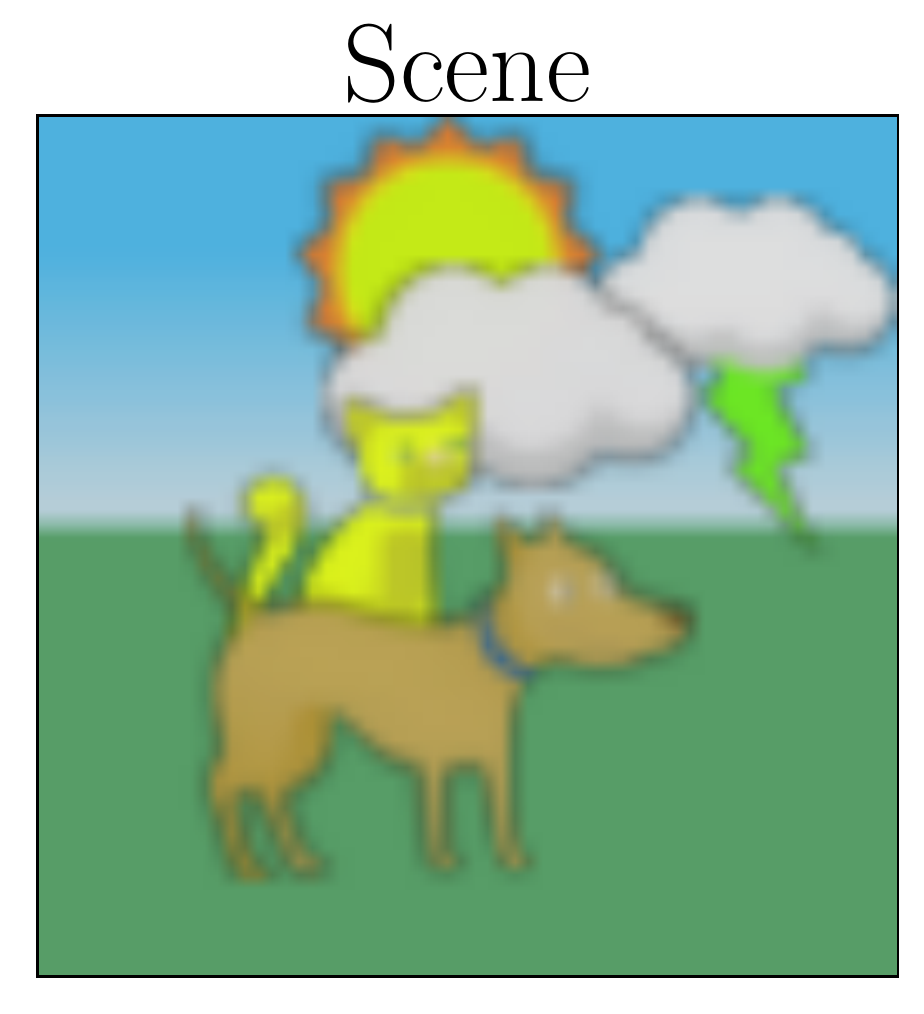}\includegraphics[width=1.8\columnwidth,valign=c]{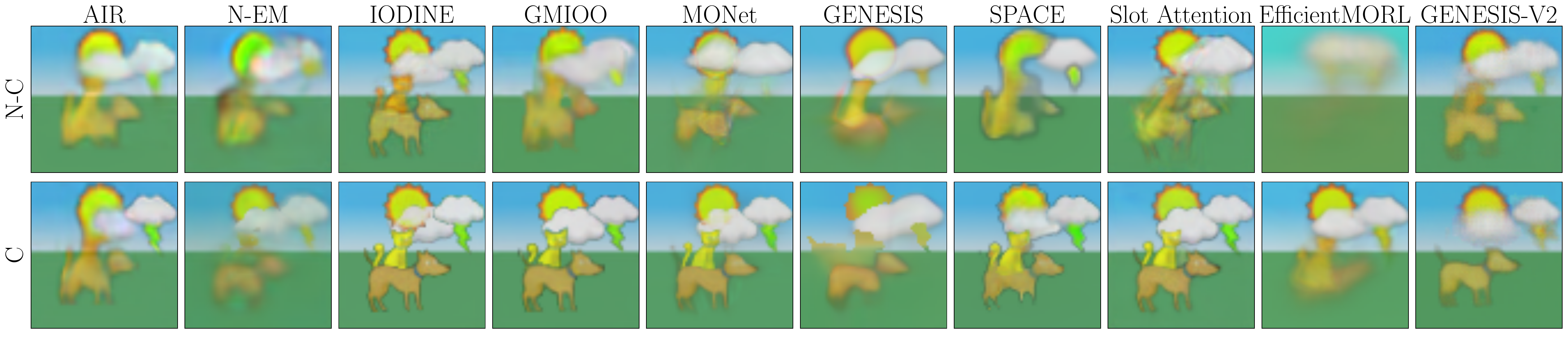}%
		\label{fig:recon_abstract}}
	\\
	\subfloat[The CLEVR dataset]{\includegraphics[width=0.23\columnwidth,valign=c]{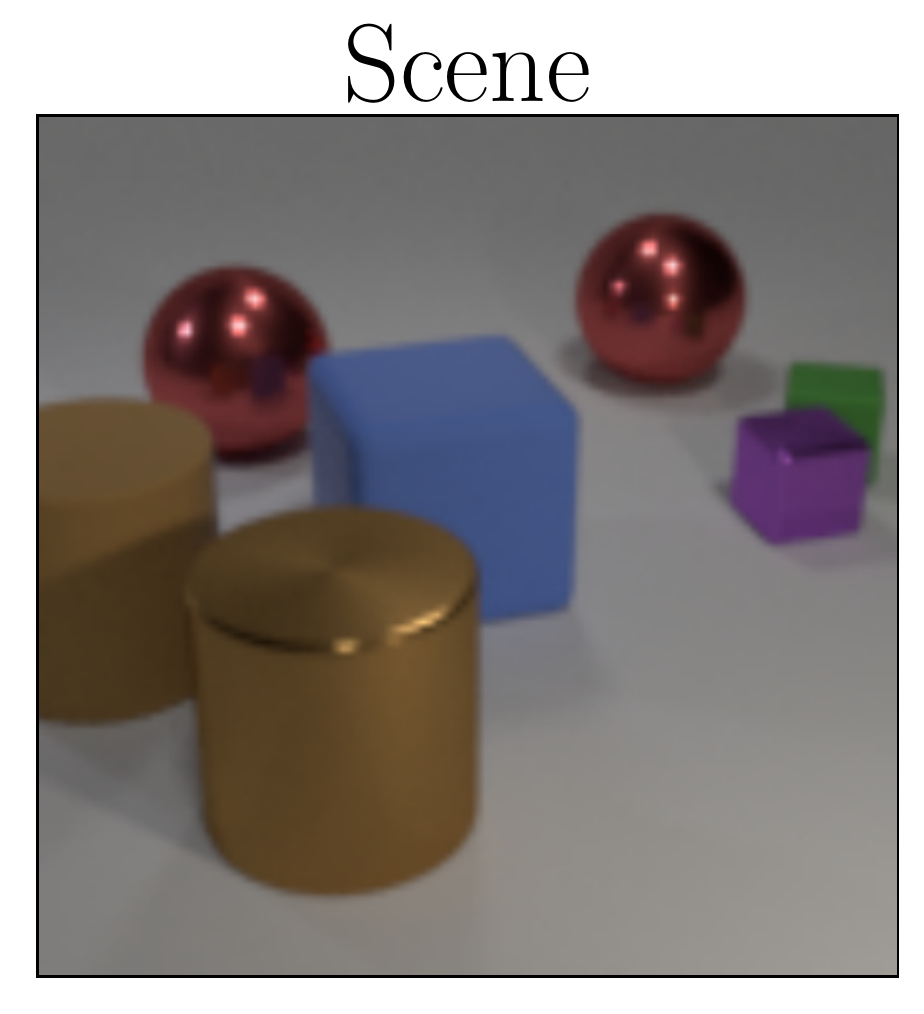}\includegraphics[width=1.8\columnwidth,valign=c]{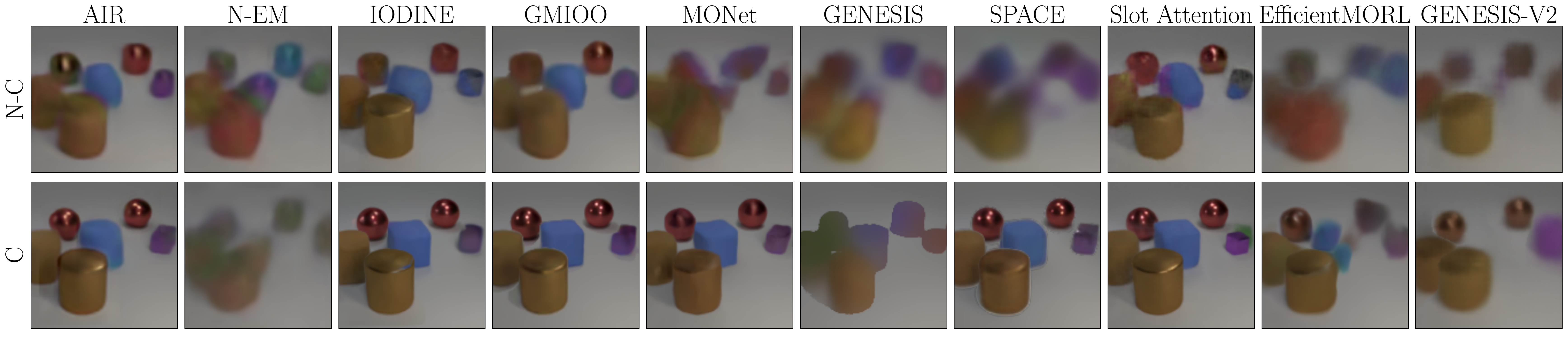}%
		\label{fig:recon_clevr}}
	\\
	\subfloat[The SHOP dataset]{\includegraphics[width=0.23\columnwidth,valign=c]{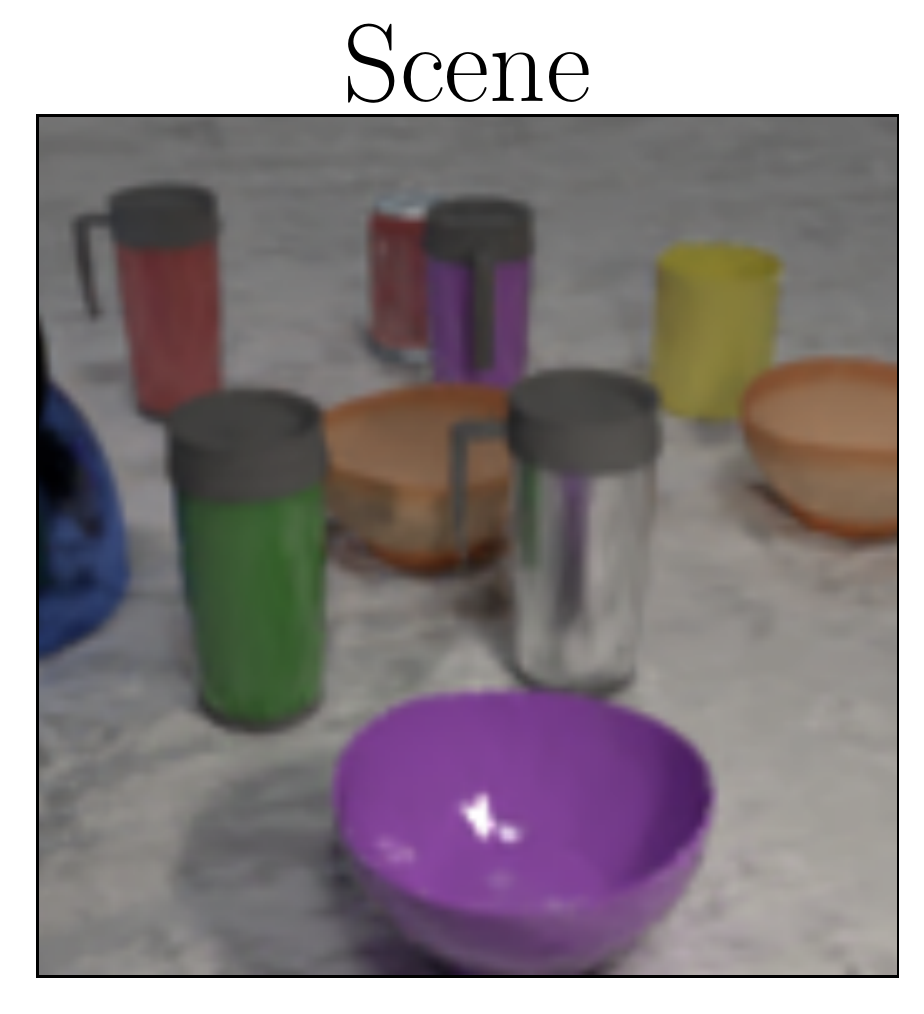}\includegraphics[width=1.8\columnwidth,valign=c]{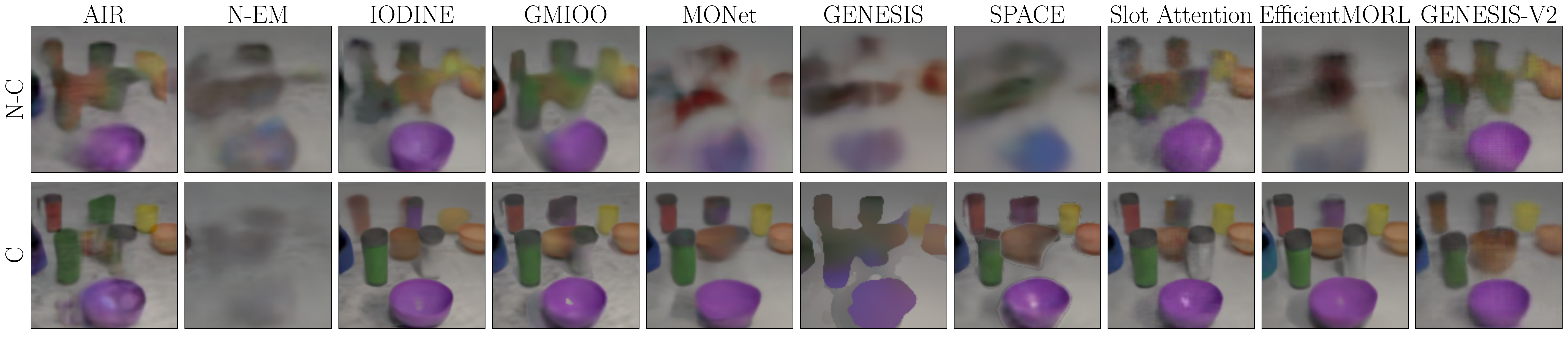}%
		\label{fig:recon_shop}}
	\\
	\subfloat[The GSO dataset]{\includegraphics[width=0.23\columnwidth,valign=c]{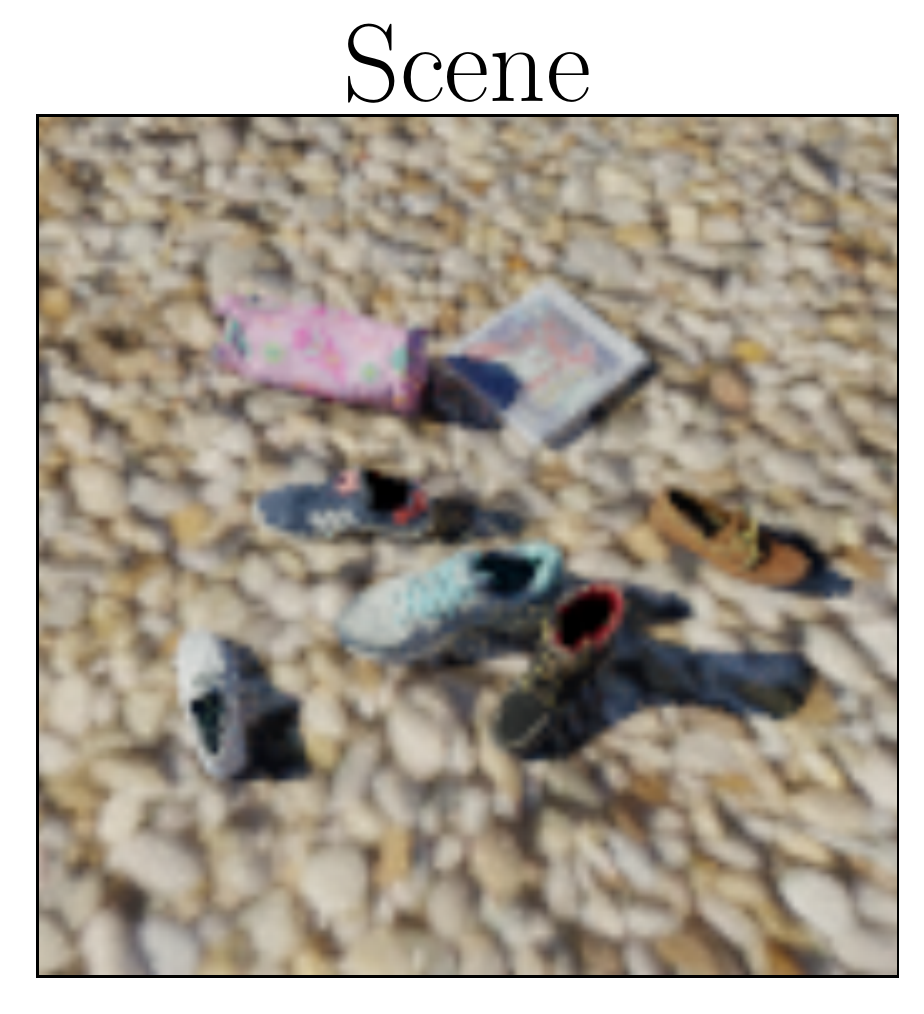}\includegraphics[width=1.8\columnwidth,valign=c]{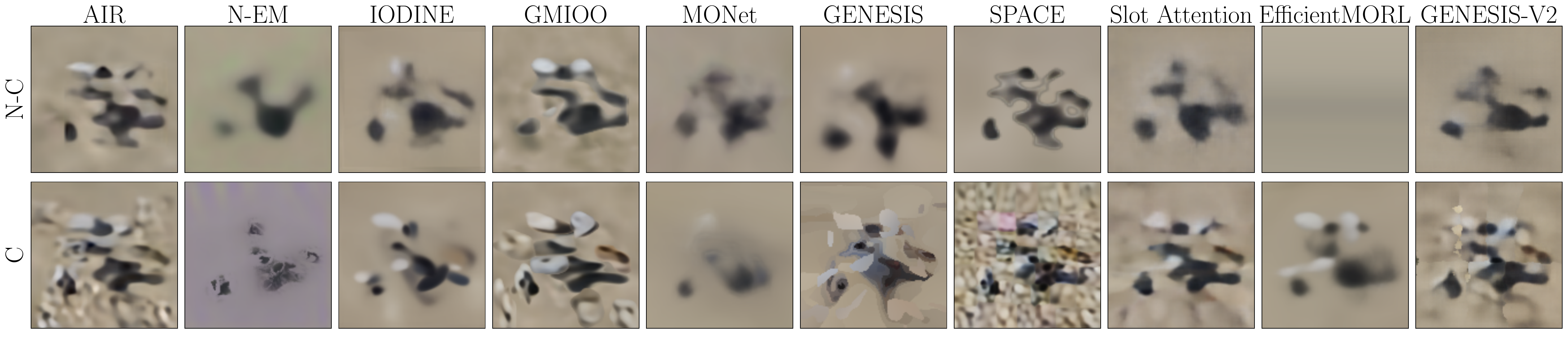}%
		\label{fig:recon_gso}}
	\caption{Comparison of reconstruction qualities of the non-compositional (N-C) and compositional (C) versions of methods on different datasets.}
	\label{fig:recon_compare}
\end{figure*}

The qualitative and quantitative comparisons of reconstruction qualities are shown in Fig. \ref{fig:recon_compare} and Table \ref{tab:recon}, respectively. According to the results, the compositional versions generally outperform the non-compositional ones, which suggests that compositional scene representations are more informative because more details are encoded in the representations, thereby implying that such representations can be learned via reconstruction when information bottlenecks exist. In a few cases, the compositional version does not achieve better reconstruction quality. A possible reason is that visual scenes are not decomposed well, as can be seen from the relatively low segmentation accuracy shown in Tables \ref{tab:test_1} and \ref{tab:test_2}. Another explanation is that the information capacity of compositional scene representations is underutilized, i.e., some bits may contain redundant or even irrelevant information. More specifically, the maximum number of layers $K + 1$ is larger than the number of objects in the visual scene, which makes representations of some layers not contribute to the reconstructed image. If it can be determined which layers can be left unused, the underutilization of information capacity may be an advantage since it enables using adaptive representation length. For methods that model the varying number of objects in a generative way, e.g., AIR \cite{Eslami2016Attend}, GMIOO \cite{Yuan2019Generative}, and SPACE \cite{Lin2020SPACE}, unused layers can be automatically determined by inference. For other methods, heuristic post-processing can be applied based on the segmentation results, e.g., layers that do not contain any segmented regions can be considered unused.

\section{Comparison of Complexities}

The comparison of different methods in terms of model complexities (number of trainable parameters) and training complexities (minimum number of GPUs required, memory footprint of each GPU, number of training steps, time spent on each training step, and total GPU time) is shown in Table \ref{tab:train}. The batch size used during training is also included because the minimum number of GPUs required and the memory footprint of each GPU are directly related to it.

Most neural networks used by existing methods are simple and basic. The most commonly used building blocks are convolution neural networks (using transposed convolution or adding upsampling in decoders), multilayer perceptrons, and recurrent neural networks (including LSTM and GRU). According to Table \ref{tab:train}, the number of network parameters is usually relatively small. It is worth mentioning that the number of network parameters does not have a strong correlation with the computational complexity (can be reflected by the product of the time spent on each training step and the minimum number of GPUs required). For example, although GMIOO \cite{Yuan2019Generative} and GENESIS \cite{Engelcke2020GENESIS} contain much more trainable parameters than the other methods used in the experiments (mainly due to the use of $5 \times 5$ convolution kernels instead of $3 \times 3$ convolution kernels), their computational complexities are significantly lower than IODINE \cite{Greff2019Multi} and EfficientMORL \cite{Emami2021Efficient}, which contain less than one-tenth as many trainable parameters.

As for training complexities, methods employing rectangular attention (e.g., AIR \cite{Eslami2016Attend}, GMIOO \cite{Yuan2019Generative}, and SPACE \cite{Lin2020SPACE}) consume less GPU memory and have lower computational complexities than other methods. The main reason is that rectangular attention is used together with spatial transformation networks, allowing using lower resolutions for decoded images of objects, thereby reducing the amount of computation in decoder networks. Moreover, methods that first randomly initialize compositional scene representations and then iteratively refine them using the information in the pixel space (e.g., N-EM \cite{Greff2017Neural}, IODINE \cite{Greff2019Multi}, and EfficientMORL \cite{Emami2021Efficient}) are more memory-intensive and computationally intensive. This is because the number of iterative steps required is relatively large, and each step involves expensive feature extraction and image generation.

\section{More Details on Datasets}

\begin{figure*}[!t]
	\centering
	\includegraphics[width=2.0\columnwidth]{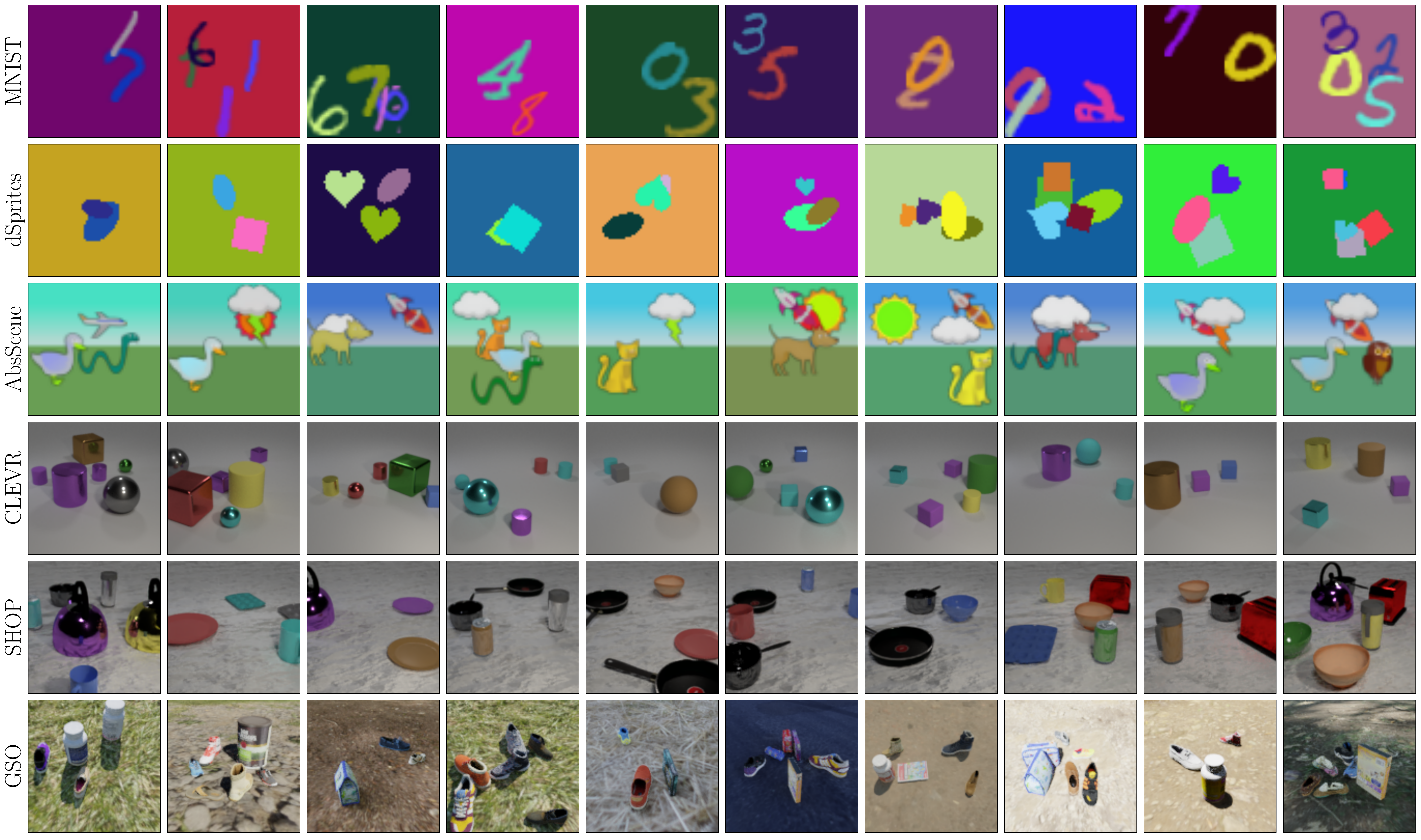}
	\caption{Samples of images in the MNIST, dSprites, AbsScene, CLEVR, SHOP, and GSO datasets.}
	\label{fig:datasets}
\end{figure*}

Samples of images in the datasets are shown in Fig. \ref{fig:datasets}. In images of the MNIST and dSprites datasets, colors of objects and background are all uniformly sampled in the RGB space $[0, 1]^3$, with the constraint that the $l_2$ distance between the colors of the background and any object is at least $0.5$. In images of the AbsScene dataset, the colors of objects and backgrounds are perturbed in the HSV space for richer diversity. Images of the CLEVR and SHOP datasets are generated based on the official code. The original images rendered by the Blender 3D engine are $214 \times 160$. These images are cropped into $128 \times 128$ by removing $19$, $67$, $43$, and $43$ pixels from the top, bottom, left, and right sides, respectively. To ensure that each object is still visible in the cropped image, the official code is modified to check the visibility of objects based on the cropped image. Images of the GSO dataset are created similarly to the CLEVR and SHOP datasets using Kubric \cite{Greff2022Kubric}. When constructing the AbsScene dataset, $10$ objects are selected to limit the complexity and diversity of visual scenes. When constructing the SHOP dataset, $10$ objects and $1$ background from the SHOP VRB dataset are selected for a similar reason. When constructing the GSO dataset that is intentionally more challenging, $373$ objects from the GSO dataset and $52$ backgrounds from the HDRI-Haven dataset are chosen.

\section{Detailed Descriptions of Metrics}

\subsection{Segmentation}

The quality of image segmentation measures how accurately the image is decomposed into different visual concepts and is evaluated using Adjusted Mutual Information (AMI) \cite{Vinh2010Information} and Adjusted Rand Index (ARI) \cite{Hubert1985Comparing}. AMI and ARI are the two most basic metrics, and almost all the compositional scene representation learning methods are evaluated using one or both of them. Two variants of AMI and ARI are used to evaluate the segmentation performance more thoroughly. AMI-A and ARI-A are computed using pixels in the entire image and measure how accurately different layers of visual concepts (including both objects and the background) are separated. AMI-O and ARI-O are computed only using pixels in the regions of objects and focus on how accurately different objects are separated.

Let $I$ denote the number of test images. For each test image, $\hat{K}_{i}$ denotes the actual number of objects in the image. $\hat{\boldsymbol{\rho}}^{i} \!\in\! \{0, 1, \dots, \hat{K}_{i}\}^{N}$ and $\boldsymbol{\rho}^{i} \!\in\! \{0, 1, \dots, K\}^{N}$ denote the actual and the estimated pixel-wise partitions of the image, respectively. $\hat{\boldsymbol{r}}^{i} \!\in\! {\{0, 1\}^{(\hat{K}_{i} + 1) \times N}}$ and $\boldsymbol{r}^{i} \!\in\! {\{0, 1\}^{(K + 1) \times N}}$ are the respective one-hot representations of partitions $\hat{\boldsymbol{\rho}}^{i}$ and $\boldsymbol{\rho}^{i}$. $\mathcal{D}$ denotes the set of pixel indexes used to compute AMI and ARI, i.e., $\mathcal{D} = \{1, 2, \dots, N\}$ when computing AMI-A and ARI-A, and $\mathcal{D} = \{n\!: \boldsymbol{x}_n \text{ is in regions of objects}\}$ when computing AMI-O and ARI-O. The computations of AMI and ARI are described below.
\begin{align}
	\text{AMI} & = \frac{1}{I} \sum_{i=1}^{I}{\frac{\MI(\hat{\boldsymbol{\rho}}_{\mathcal{D}}^{i}, \boldsymbol{\rho}_{\mathcal{D}}^{i}) - \mathbb{E}[\MI(\hat{\boldsymbol{\rho}}_{\mathcal{D}}^{i}, \boldsymbol{\rho}_{\mathcal{D}}^{i})]}{\big(\!\entropy(\hat{\boldsymbol{\rho}}_{\mathcal{D}}^{i}) + \entropy(\boldsymbol{\rho}_{\mathcal{D}}^{i})\big) / 2 - \mathbb{E}[\MI(\hat{\boldsymbol{\rho}}_{\mathcal{D}}^{i}, \boldsymbol{\rho}_{\mathcal{D}}^{i})]}}
	\label{equ:ami} \\
	\text{ARI} & = \frac{1}{I} \sum_{i=1}^{I}{\frac{b_{\text{all}}^{i} - b_{\text{row}}^{i} \cdot b_{\text{col}}^{i} / c^{i}}{\big(b_{\text{row}}^{i} + b_{\text{col}}^{i}\big) / 2 - b_{\text{row}}^{i} \cdot b_{\text{col}}^{i} / c^{i}}}
	\label{equ:ari}
\end{align}
In Eq. \eqref{equ:ami}, $\MI$ and $\entropy$ denote mutual information and entropy, respectively. In Eq. \eqref{equ:ari}, $b_{\text{all}}^{i}$, $b_{\text{row}}^i$, $b_{\text{col}}^i$, and $c^{i}$ are intermediate variables. Let $C(x, y)$ denote the number of combinations $\frac{x!}{(x - y)! \, y!}$, and $v_{\hat{k},k}^{i}$ denote the dot product $\sum\nolimits_{n \in \mathcal{D}}{\big(\hat{r}_{\hat{k},n}^{i} \cdot r_{k,n}^{i}\big)}$. The computations of these intermediate variables are described below.
\begin{align}
	b_{\text{all}}^{i} & = \sum\nolimits_{\hat{k}=0}^{\hat{K}_{i}}{\sum\nolimits_{k=0}^{K}{C\big(v_{\hat{k},k}^{i}, 2\big)}} \\
	b_{\text{row}}^i & = \sum\nolimits_{\hat{k}=0}^{\hat{K}_i}{C\Big(\sum\nolimits_{k=0}^{K}{v_{\hat{k},k}^{i}}, 2\Big)} \\
	b_{\text{col}}^i & = \sum\nolimits_{k=0}^{K}{C\Big(\sum\nolimits_{\hat{k}=0}^{\hat{K}_i}{v_{\hat{k},k}^{i}}, 2\Big)} \\
	c^{i} & = C\Big(\sum\nolimits_{\hat{k}=0}^{\hat{K}_i}{\sum\nolimits_{n \in \mathcal{D}}{\hat{r}_{\hat{k},n}^i}}, 2\Big)
\end{align}

\subsection{Amodal Segmentation}

The amodal segmentation performance measures how accurately the complete shapes of objects are estimated. It provides additional information compared to the segmentation performance because the quality of object shape estimations in occluded regions is considered. Intersection over Union (IoU) and $F_1$ score (F1) are used to evaluate the performance of amodal segmentation. The former tends to reflect the worst performance of complete shape estimations among all the objects in the image, while the latter tends to reflect the average performance. These metrics are only evaluated for methods that can estimate the complete shapes of objects.

Let $\hat{\boldsymbol{s}}_{1:\hat{K}_{i}}^{i} \!\in\! [0, 1]^{\hat{K}_{i} \times N}$ and $\boldsymbol{s}_{1:K}^{i} \!\in\! [0, 1]^{K \times N}$ denote the actual and the estimated complete shapes of objects in the $i$th test image, respectively. To compute IoU and F1, the correspondences between layers in the actual and the estimated complete shapes need to be determined because $\hat{K}_{i}$ and $K$ may not be equal, and the orderings of objects may be different in $\hat{\boldsymbol{s}}_{1:\hat{K}_{i}}^{i}$ and $\boldsymbol{s}_{1:K}^{i}$. $\Xi$ denotes the set containing all the $K!$ possible permutations of the indexes $\{1, 2, \dots, K\}$. The correspondences between $\hat{\boldsymbol{s}}_{1:\hat{K}_{i}}^{i}$ and $\boldsymbol{s}_{1:K}^{i}$ are computed by $\boldsymbol{\xi}^{i} = \argmax_{\boldsymbol{\xi} \in \Xi}{\sum_{k=1}^{\hat{K}_i}{\sum_{n=1}^{N}{\hat{r}_{k,n}^{i} \cdot r_{\xi_k^i,n}^{i}}}}$, which can be efficiently solved based on linear sum assignment. IoU and F1 are computed using the following expressions.
\begin{align}
	\text{IoU} & = \frac{1}{I} \sum_{i=1}^{I}{\frac{1}{\hat{K}_i} \sum_{k=1}^{\hat{K}_i}{\frac{\sum\nolimits_{n=1}^{N}{\min(\hat{s}_{k,n}^{i}, s_{\xi_k^i,n}^{i})}}{\sum\nolimits_{n=1}^{N}{\max(\hat{s}_{k,n}^{i}, s_{\xi_k^i,n}^{i})}}}} \\
	\text{F1} & = \frac{1}{I} \sum_{i=1}^{I}{\frac{1}{\hat{K}_i} \sum_{k=1}^{\hat{K}_i}{\frac{2 \cdot \sum\nolimits_{n=1}^{N}{\min(\hat{s}_{k,n}^{i}, s_{\xi_k^i,n}^{i})}}{\sum\nolimits_{n=1}^{N}{\hat{s}_{k,n}^{i}} + \sum\nolimits_{n=1}^{N}{s_{\xi_k^i,n}^{i}}}}}
\end{align}

\subsection{Object Counting}

For methods that additionally model the number of objects, Object Counting Accuracy (OCA) measures the quality of object number estimation. OCA is computed as the ratio of the number of images in which the number of objects is correctly estimated to the total number of images.

Let $\hat{K}_{i}$ and $\tilde{K}_{i}$ be the actual and estimated numbers of objects in the $i$th test image, respectively, and $\delta$ denote the Kronecker delta function. The computation of OCA is
\begin{equation}
	\text{OCA} = \frac{1}{I} \sum\nolimits_{i=1}^{I}{\delta_{\hat{K}_{i}, \tilde{K}_{i}}}
\end{equation}

\subsection{Object Ordering}

Object Ordering Accuracy (OOA) is used to evaluate the accuracy of depth ordering estimation for methods that additionally model the depth ordering of objects. The computation of OOA is based on the weighted average of pairwise ordering estimations of objects. The weight of each pair of objects is the overlapping area determined by the complete shapes of these objects. Different pairs of objects do not use the same weights because the ordering of two objects with a higher degree of overlap can be more easily estimated.

Similar to amodal segmentation performance, the evaluation of object ordering performance also requires computing correspondences between objects in the ground truth annotations and the estimated results. Let $\boldsymbol{\xi}^{i} = \argmax_{\boldsymbol{\xi} \in \Xi}{\sum_{k=1}^{\hat{K}_i}{\sum_{n=1}^{N}{\hat{r}_{k,n}^{i} \cdot r_{\xi_k^i,n}^{i}}}}$ denote the correspondences computed in the way described in the amodal segmentation part, and $\hat{\eta}_{k_1, k_2}^{i} \!\in\! \{0, 1\}$ and $\eta_{\xi_{k_1}^{i}, \xi_{k_2}^{i}}^{i} \!\in\! \{0, 1\}$ denote the actual and estimated pairwise orderings of the $k_1$th and $k_2$th objects in the $i$th test image, respectively. $\hat{\eta}_{k_1, k_2}^{i}$ and $\eta_{\xi_{k_1}^{i}, \xi_{k_2}^{i}}^{i}$ equal $1$ if the depth of the $k_1$th object is smaller than the $k_2$th object and equal $0$ otherwise. The overlapping area of the $k_1$th and $k_2$th objects is computed based on the actual complete shapes of objects $\hat{\boldsymbol{s}}$, i.e., $w_{k_1,k_2}^{i} = \sum\nolimits_{n=1}^{N}{\hat{s}_{k_1,n}^{i} \cdot \hat{s}_{k_2,n}^{i}}$, and OOA is computed using the following expression.
\begin{equation}
	\text{OOA} = \frac{1}{I} \sum_{i=1}^{I}{\frac{\sum\nolimits_{k_1=1}^{\hat{K}_i - 1}{\!\sum\nolimits_{k_2=k_1 + 1}^{\hat{K}_i}{w_{k_1\!,k_2}^{i} \!\cdot \delta_{\hat{\eta}_{k_1\!,k_2}^{i},\, \eta_{\xi_{k_1}^{i}\!\!,\xi_{k_2}^{i}}}}}}{\sum\nolimits_{k_1=1}^{\hat{K}_i - 1}{\!\sum\nolimits_{k_2=k_1 + 1}^{\hat{K}_i}{w_{k_1\!,k_2}^{i}}}}}
\end{equation}

\section{Detailed Benchmark Results}

The performance comparison of different methods on each dataset is presented in Tables \ref{tab:test_1} and \ref{tab:test_2}. All the models are trained once and tested for five runs because all the methods have more or less randomness:
\begin{itemize}[leftmargin=*]
	\item AIR \cite{Eslami2016Attend}, IODINE \cite{Greff2019Multi}, GMIOO \cite{Yuan2019Generative}, MONet \cite{Burgess2019MONet}, GENESIS \cite{Engelcke2020GENESIS}, SPACE \cite{Lin2020SPACE}, EfficientMORL \cite{Emami2021Efficient}, and GENESIS-V2 \cite{Engelcke2021GENESIS} define the prior distribution of compositional scene representations and are probabilistic in nature. The variational inference employed by these methods involves a sampling operation that introduces randomness.
	\item N-EM \cite{Greff2017Neural}, IODINE \cite{Greff2019Multi}, Slot Attention \cite{Locatello2020Object}, and EfficientMORL \cite{Emami2021Efficient} infer compositional scene representations via random initialization and iterative refinement. Different initializations will lead to different inference results.
	\item GENESIS-V2 \cite{Engelcke2021GENESIS} computes attention masks in a stochastic way. Changes in attention masks will cause changes in the inferred compositional scene representations.
\end{itemize}
The mean and standard deviation of five runs are both included in the reported results. It can be seen that:
\begin{itemize}[leftmargin=*]
	\item The randomness in the methods does not lead to much uncertainty in performance.
	\item On the MNIST and dSprites datasets, where images are synthesized based on layer compositions and appearances of visual concepts are solid colors, GMIOO \cite{Yuan2019Generative} performs best in almost all aspects. Possible reasons include modeling visual scenes similarly to the synthesis of images, utilizing prior knowledge of the distribution of positions and scales of objects, and adopting iterative inference that is beneficial for handling object occlusions.
	\item On the AbsScene dataset, where appearances of objects in images synthesized based on layer compositions are more complex than the MNIST and dSprites datasets, IODINE \cite{Greff2019Multi} performs best in terms of AMI and ARI. There are mainly two possible reasons. Firstly, the perceived shapes are directly modeled using the softmax function, avoiding the need to apply additional steps to distinguish background from objects and determine the depth ordering of objects. Secondly, compositional scene representations are iteratively refined based on gradient information, leading to easier separation of objects in boundary regions.
	\item On the CLEVR dataset, the performance of various methods, i.e., AIR \cite{Eslami2016Attend}, IODINE \cite{Greff2019Multi}, GMIOO \cite{Yuan2019Generative}, MONet \cite{Burgess2019MONet}, SPACE \cite{Lin2020SPACE}, and Slot Attention \cite{Locatello2020Object}, is close in terms of AMI-O and ARI-O, which indicates that these methods can all separate different objects relatively well if the background region is excluded. In general, MONet performs best, especially in AMI-A and ARI-A. A possible reason is that MONet applies U-Net to compute attention masks based on hierarchical features. The structure of U-Net includes inductive biases that make better use of edge information to generate attention masks, resulting in more accurate discrimination between the background and objects in the presence of shadows.
	\item On the SHOP dataset, MONet \cite{Burgess2019MONet} and GMIOO \cite{Yuan2019Generative} perform best in general. It is hard to determine which one is better. Except for AIR \cite{Eslami2016Attend}, which does not model the shapes of objects, methods that distinguish background from objects perform noticeably better than other methods in terms of AMI-A and ARI-A. This finding also applies to the CLEVR dataset. It can be seen that distinguishing background from objects is beneficial for the segmentation performance, especially when shadows exist.
	\item On the GSO dataset, GMIOO \cite{Yuan2019Generative} performs best in most aspects. Although images in the GSO dataset are synthesized using a 3D engine, they are largely photorealistic because all the 3D models are scanned from real-world objects, and all the backgrounds are real images. The encouraging performance on this challenging dataset shows the potential of applying reconstruction-based compositional scene representation learning with neural networks to more complex real-world visual scenes.
	\item For all the methods, the performance usually does not degrade much when test images contain more objects than those used for training, which validates the generalizability of compositional scene representation learning.
\end{itemize}

\input{table_train}
\input{table_test1}
\input{table_test2}
	
\end{appendices}

\end{document}

%% file: table_notation.tex
\begin{table}[!t]
	\addtolength{\tabcolsep}{-3pt}
	\renewcommand{\arraystretch}{1.35}
    \caption{Notations used throughout the paper.}
    \label{tab:notation}
    \centering
    \begin{tabular}{|c|l|}
        \hline
         Notation & Meaning  \\ \hline
         $N \in \mathbb{Z}_{+}$ & The number of pixels in each image \\ \hline
         $C \in \mathbb{Z}_{+}$ & The number of image channels \\ \hline
         $K \in \mathbb{Z}_{+}$ & The number of layers modeling objects \\ \hline
         $\boldsymbol{x} \in \mathbb{R}^{N \times C}$ & The observed image \\ \hline
         $\tilde{\boldsymbol{x}} \in \mathbb{R}^{N \times C}$ & The reconstructed image \\ \hline
         $\boldsymbol{a}_{k} \in \mathbb{R}^{N \times C}$ & The appearance of the $k$th visual concept \\ \hline
         $\boldsymbol{s}_{k} \in [0, 1]^{N}$ & The complete shape (or logit of perceived shape if not \\[-0.03in]
         (or $\mathbb{R}^{N}$) & modeling complete shape) of the $k$th visual concept \\ \hline
         \multirow{1.8}{*}{$\boldsymbol{\pi}_{k} \in [0, 1]^{N}$} & The perceived shape of the $k$th visual concept \\[-0.03in]
         & (may be incomplete due to occlusion)  \\ \hline
         \multirow{1.8}{*}{$o_k \in \mathbb{R}_{+}$} & The optional variable describing the depth of the $k$th \\[-0.05in]
         & visual concept \\ \hline
         \multirow{1.8}{*}{$\boldsymbol{z}_{k}$} & The representation or the collection of representations \\[-0.03in]
         & of the $k$th visual concept \\
         \hline
    \end{tabular}
\end{table}

%% file: fig_history.tex
\begin{tikzpicture}[
	node distance=0.5cm and 1.4cm,
	mynode/.style={
		draw,rounded rectangle,fill=color_feat,minimum width=2.4cm,minimum height=0.7cm,inner sep=0cm,outer sep=0cm,align=center,font=\footnotesize
	},
	mynoderound/.style={
		draw,rounded rectangle,fill=color_sync,minimum width=2.4cm,minimum height=0.7cm,inner sep=0cm,outer sep=0cm,align=center,font=\footnotesize
	},
	mydot/.style={
		circle,minimum size=0cm,inner sep=0cm,outer sep=0cm,align=center
	},
]

\node[mynoderound] (rc) {\textbf{RC}\\[-0.6mu]2016 \cite{Greff2016Binding}};
\node[mynoderound,above=of rc] (tagger) {\textbf{Tagger}\\[-0.6mu]2016 \cite{Greff2016Tagger}};
\node[mynoderound,above=of tagger] (rtagger) {\textbf{RTagger}\\[-0.6mu]2017 \cite{PremontSchwarz2017Recurrent}};
\node[mynoderound,below=of rc] (nem) {\textbf{N-EM}\\[-0.6mu]2017 \cite{Greff2017Neural}};
\node[mynoderound,below=of nem] (rnem) {\textbf{Relational N-EM}\\[-0.6mu]2018 \cite{Steenkiste2018Relational}};

\node[mynoderound,right=of tagger] (iodine) {\textbf{IODINE}\\[-0.6mu]2019 \cite{Greff2019Multi}};
\node[mynoderound,below=of iodine] (ldp) {\textbf{LDP}\\[-0.6mu]2019 \cite{Yuan2019Spatial}};

\node[mynode,below=of rnem] (cstvae) {\textbf{CST-VAE}\\[-0.6mu]2016 \cite{Huang2016Efficient}};
\node[mynode,below=of cstvae] (air) {\textbf{AIR}\\[-0.6mu]2016 \cite{Eslami2016Attend}};
\node[mynode,below=of air] (sqair) {\textbf{SQAIR}\\[-0.6mu]2018 \cite{Kosiorek2018Sequential}};
\node[mynode,below=of sqair] (rsqair) {\textbf{R-SQAIR}\\[-0.6mu]2019 \cite{Stanic2019R}};

\node[mynode,below=of ldp] (gmioo) {\textbf{GMIOO}\\[-0.6mu]2019 \cite{Yuan2019Generative}};
\node[mynode,below=of gmioo] (monet) {\textbf{MONet}\\[-0.6mu]2019 \cite{Burgess2019MONet}};
\node[mynode,below=of monet] (supair) {\textbf{SuPAIR}\\[-0.6mu]2019 \cite{Stelzner2019Faster}};
\node[mynode,below=of supair] (asr) {\textbf{ASR}\\[-0.6mu]2019 \cite{Xu2019Multi}};
\node[mynode,below=of asr] (spair) {\textbf{SPAIR}\\[-0.6mu]2019 \cite{Crawford2019Spatially}};
\node[mynode,below=of spair] (silot) {\textbf{SILOT}\\[-0.6mu]2020 \cite{Crawford2020Exploiting}};

\node[mynoderound,right=of iodine] (mulmon) {\textbf{MulMON}\\[-0.6mu]2020 \cite{Li2020Learning}};
\node[mynoderound,below=of mulmon] (slotattn) {\textbf{Slot Attention}\\[-0.6mu]2020 \cite{Locatello2020Object}};
\node[mynode,below=of slotattn] (yangetal) {\textbf{Yang et al.}\\[-0.6mu]2020 \cite{Yang2020Learning}};
\node[mynode,below=of yangetal] (genesis) {\textbf{GENESIS}\\[-0.6mu]2020 \cite{Engelcke2020GENESIS}};
\node[mynode,below=of genesis] (gnm) {\textbf{GNM}\\[-0.6mu]2020 \cite{Jiang2020Generative}};
\node[mynode,below=of gnm] (scalor) {\textbf{SCALOR}\\[-0.6mu]2020 \cite{Jiang2020SCALOR}};
\node[mynode,below=of scalor] (space) {\textbf{SPACE}\\[-0.6mu]2020 \cite{Lin2020SPACE}};
\node[mynode,below=of space] (gswm) {\textbf{G-SWM}\\[-0.6mu]2020 \cite{Lin2020Improving}};

\node[mynoderound,right=of mulmon] (dymon) {\textbf{DyMON}\\[-0.6mu]2021 \cite{Nanbo2021Object}};
\node[mynoderound,above=of dymon] (provide) {\textbf{PROVIDE}\\[-0.6mu]2021 \cite{Zablotskaia2021PROVIDE}};
\node[mynoderound,below=of dymon] (efficientmorl) {\textbf{EfficientMORL}\\[-0.6mu]2021 \cite{Emami2021Efficient}};
\node[mynoderound,below=of efficientmorl] (simone) {\textbf{SIMONe}\\[-0.6mu]2021 \cite{Kabra2021SIMONe}};
\node[mynoderound,below=of simone] (ocloc) {\textbf{OCLOC}\\[-0.6mu]2022 \cite{Yuan2022Unsupervised}};
\node[mynode,below=of ocloc] (genesisv2) {\textbf{GENESIS-V2}\\[-0.6mu]2021 \cite{Engelcke2021GENESIS}};
\node[mynode,below=of genesisv2] (vimon) {\textbf{ViMON}\\[-0.6mu]2021 \cite{Weis2021Benchmarking}};
\node[mynode,below=of vimon] (gsgn) {\textbf{GSGN}\\[-0.6mu]2021 \cite{Deng2021Generative}};
\node[mynode,below=of gsgn] (roots) {\textbf{ROOTS}\\[-0.6mu]2021 \cite{Chen2021ROOTS}};

\node[mynoderound,right=of dymon] (vikstroemetal) {\textbf{Vikström et al.}\\[-0.6mu]2022 \cite{Vikstroem2022Learning}};
\node[mynoderound,below=of vikstroemetal] (savi) {\textbf{SAVi}\\[-0.6mu]2022 \cite{Kipf2022Conditional}};
\node[mynoderound,below=of savi] (savipp) {\textbf{SAVi++}\\[-0.6mu]2022 \cite{Elsayed2022SAVi}};
\node[mynoderound,below=of savipp] (slotformer) {\textbf{SlotFormer}\\[-0.6mu]2023 \cite{Wu2023SlotFormer}};
\node[mynoderound,below=of slotformer] (boqsa) {\textbf{BO-QSA}\\[-0.6mu]2023 \cite{Jia2023Unsupervised}};
\node[mynoderound,below=of boqsa] (dinosaur) {\textbf{DINOSAUR}\\[-0.6mu]2023 \cite{Seitzer2023Bridging}};

\node[mydot,below=0.25cm of genesis] (dot1) {};
\node[mydot,left=0.8cm of dot1] (dot2) {};
\node[mydot,right=0.8cm of dot1] (dot3) {};

\node[mydot,below=0.25cm of efficientmorl] (dot4) {};
\node[mydot,left=0.8cm of dot4] (dot5) {};
\node[mydot,right=0.8cm of dot4] (dot6) {};

\node[mydot,right=0.4cm of ocloc] (dot7) {};

\path
(tagger) edge[-latex] (rtagger)
(rc) edge[-latex] (tagger)
(rc) edge[-latex] (nem)
(cstvae) edge[-latex,out=30,in=-150] (ldp)
(air) edge[-latex] (sqair)
(air) edge[-latex,out=30,in=-150] (gmioo)
(air) edge[-latex,out=25,in=-155] (monet)
(air) edge[-latex] (supair)
(air) edge[-latex] (asr)
(air) edge[-latex] (spair)
(nem) edge[-latex] (rnem)
(nem) edge[-latex] (iodine)
(nem) edge[-latex] (ldp)
(sqair) edge[-latex] (rsqair)
(iodine) edge[-latex] (mulmon)
(iodine) edge[-latex] (slotattn)
(iodine) edge[-latex] (provide)
(iodine) edge[-latex] (efficientmorl)
(ldp) edge[-latex] (gmioo)
(monet) edge[-latex] (yangetal)
(monet) edge[-latex] (genesis)
(monet) edge[-,out=-20,in=180] (dot2)
(dot2) edge[-] (dot3)
(dot3) edge[-latex,out=0,in=150] (vimon)
(spair) edge[-latex] (space)
(spair) edge[-latex] (silot)
(mulmon) edge[-latex] (dymon)
(slotattn) edge[-latex] (efficientmorl)
(slotattn) edge[-latex] (simone)
(slotattn) edge[-latex,out=-25,in=150] (ocloc)
(slotattn) edge[-latex,out=-30,in=150] (genesisv2)
(genesis) edge[-latex] (genesisv2)
(space) edge[-latex,out=20,in=-20] (gnm)
(space) edge[-latex] (scalor)
(space) edge[-latex] (gswm)
(space) edge[-latex] (gsgn)
(space) edge[-latex] (roots)
(slotattn) edge[-,out=-10,in=180] (dot5)
(dot5) edge[-] (dot6)
(dot6) edge[-latex,out=0,in=180] (vikstroemetal)
(dot6) edge[-latex,out=0,in=180] (savi)
(savi) edge[-latex] (savipp)
(dot6) edge[-,out=-10,in=90] (dot7)
(dot6) edge[-latex,out=0,in=180] (slotformer)
(dot7) edge[-latex,out=-90,in=175] (boqsa)
(dot7) edge[-latex,out=-90,in=170] (dinosaur)
;

\end{tikzpicture}

%% file: fig_compare.tex
\begin{figure}[!t]
    \small
	\centering
	\addtolength{\tabcolsep}{-5.2pt}
	\begin{tabular}{r|C{0.26in}|C{0.21in}|C{0.25in}|C{0.21in}|C{0.21in}|C{0.21in}|C{0.21in}|C{0.23in}|C{0.25in}}
		\toprule
		Methods & [C] & [S] & [R] & [N] & [L] & [V] & [M] & [I] & [A] \\ \midrule
		RC \cite{Greff2016Binding} & M & N & R-E & & & & & RE & A-N \\
		Tagger \cite{Greff2016Tagger} & M & N & R-E & & & & & RE & A-N \\
		RTagger \cite{PremontSchwarz2017Recurrent} & M & N & R-E & & & & \checkmark & RE & A-N \\
		N-EM \cite{Greff2017Neural} & M & N & R-E & & & & & EM & A-N \\
		Relational N-EM \cite{Steenkiste2018Relational} & M & N & R-E & & \checkmark & & & EM & A-N \\
		LDP \cite{Yuan2019Spatial} & M & S & R-E & & & & & EM & A-N \\
		IODINE \cite{Greff2019Multi} & M & N & R-L & & & & & VI & A-N \\
		MulMON \cite{Li2020Learning} & M & N & R-L & & & \checkmark & & VI & A-N \\
		DyMON \cite{Nanbo2021Object} & M & N & R-L & & & \checkmark & \checkmark & VI & A-N \\
		PROVIDE \cite{Zablotskaia2021PROVIDE} & M & N & R-L & & & & \checkmark & VI & A-N \\
		PSGNet \cite{Bear2020Learning} & S & N & R-E & & & & \checkmark & RE & A-F \\
		Slot Attention \cite{Locatello2020Object} & S & N & R-E & & & & & RE & A-F \\
		EfficientMORL \cite{Emami2021Efficient} & M/S & N & R-L & & & & & VI & A-F \\
		SIMONe \cite{Kabra2021SIMONe} & M & N & R-L & & & \checkmark & & VI & A-F \\
		OCLOC \cite{Yuan2022Unsupervised} & M & O & R-L & \checkmark & & \checkmark & & VI & A-F \\
		Vikström et al. \cite{Vikstroem2022Learning} & S & N & R-E & & & & & RE & A-F \\
		SAVi \cite{Kipf2022Conditional} & S & N & R-E & & \checkmark & & \checkmark & RE & A-F \\
		SAVi++ \cite{Elsayed2022SAVi} & S & N & R-E & & \checkmark & & \checkmark & RE & A-F \\
		SlotFormer \cite{Wu2023SlotFormer} & S & N & R-E & & \checkmark & & \checkmark & RE & A-F \\
		BO-QSA \cite{Jia2023Unsupervised} & S & N & R-E & & & & & RE & A-F \\
		DINOSAUR \cite{Seitzer2023Bridging} & S & N & R-E & & & & & RE & A-F \\
		CST-VAE \cite{Huang2016Efficient} & S & S & R-L & & & & & VI & R-G \\
		AIR \cite{Eslami2016Attend} & S & & R-L & \checkmark & & & & VI & R-G \\
		SQAIR \cite{Kosiorek2018Sequential} & S & & R-L & \checkmark & & & \checkmark & VI & R-G \\
		R-SQAIR \cite{Stanic2019R} & S & & R-L & \checkmark & \checkmark & & \checkmark & VI & R-G \\
		ASR \cite{Xu2019Multi} & S & & R-L & \checkmark & \checkmark & & & VI & R-G \\
		GMIOO \cite{Yuan2019Generative} & M & S & R-L & \checkmark & & & & VI & R-G \\
		SuPAIR \cite{Stelzner2019Faster} & S & S & R-L & \checkmark & & & & VI & R-G \\
		TBA \cite{He2019Tracking} & S & S & R-E & \checkmark & & & \checkmark & RE & A-F \\
		SPAIR \cite{Crawford2019Spatially} & S & O & R-L & \checkmark & & & & VI & R-L \\
		SILOT \cite{Crawford2020Exploiting} & S & O & R-L & \checkmark & & & \checkmark & VI & R-L \\
		SPACE \cite{Lin2020SPACE} & S & O & R-L & \checkmark & & & & VI & R-L \\
		SCALOR \cite{Jiang2020SCALOR} & S & O & R-L & \checkmark & & & \checkmark & VI & R-L \\
		GNM \cite{Jiang2020Generative} & S & O & R-L & \checkmark & \checkmark & & & VI & R-L \\
		G-SWM \cite{Lin2020Improving} & S & O & R-L & \checkmark & \checkmark & & \checkmark & VI & R-L \\
		GSGN \cite{Deng2021Generative} & S & O & P-V & \checkmark & \checkmark & & & VI & R-L \\
		ROOTS \cite{Chen2021ROOTS} & S & O & R-L & \checkmark & \checkmark & \checkmark & & VI & R-L \\
		MONet \cite{Burgess2019MONet} & M & N & R-L & & & & & VI & A-N \\
		Yang et al. \cite{Yang2020Learning} & M & N & R-L & & & & & VI & A-N \\
		ViMON \cite{Weis2021Benchmarking} & M & N & R-L & & & & \checkmark & VI & A-N \\
		POD-Net \cite{Du2021Unsupervised} & M & N & R-L & & & & \checkmark & VI & A-N \\
		GENESIS \cite{Engelcke2020GENESIS} & M & S & R-L & & \checkmark & & & VI & A-N \\
		GENESIS-V2 \cite{Engelcke2021GENESIS} & M & N & R-L & & \checkmark & & & VI & A-F \\
		DTI-Sprites \cite{Monnier2021Unsupervised} & S & S & P-I & \checkmark & & & & RE & R-G \\
		PCDNet \cite{VillarCorrales2022Unsupervised} & S & S & P-I & & & & & RE & R-G \\
		MarioNette \cite{Smirnov2021MarioNette} & S & S & P-V & & & & & RE & R-L \\
		Loci \cite{Traub2023Learning} & M & O & R-L & & \checkmark & & \checkmark & VI & A-N \\
		Gao \& Li \cite{Gao2023Time} & S & O & R-L & \checkmark & & \checkmark & & VI & A-F \\
		 \bottomrule
	\end{tabular}
	\caption{Categorization of existing methods in terms of: composition of layers [C] (\textit{M: spatial mixture models, S: weighted summations}); modeling of shapes [S] (\textit{N: direct normalization of network outputs, O: normalization of complete shapes and ordering, S: stick-breaking composition of complete shapes, blank: shape is not modeled}); representation of objects [R] (\textit{R-E: embeddings in real vector space, R-L: latent variables in real vector space, P-I: a finite number of prototype images, P-V: a finite number of prototype vectors}); whether modeling the number of objects [N], layouts of scenes [L], multiple viewpoints of scenes [V], and motions of objects [M]; inference frameworks [I] (\textit{VI: amortized variational inference, EM: expectation maximization, RE: reconstruction error}); and attention mechanisms [A] (\textit{R-G: rectangular attention based on global features, R-L: rectangular attention based on local features, A-N: arbitrary-shaped attention based on network outputs, A-F: arbitrary-shaped attention based on feature similarities}).
	}
	\label{fig:compare}
\end{figure}

%% file: table_data.tex
\begin{table*}[!t]
	\small
	\centering
	\addtolength{\tabcolsep}{-2pt}
	\renewcommand{\arraystretch}{1.2}
	\caption{Configurations of the MNIST, dSprites, AbsScene, CLEVR, SHOP, and GSO datasets.}
	\label{tab:datasets}
	\begin{tabular}{|C{0.65in}|C{0.4in}|C{0.4in}|C{0.4in}|C{0.4in}|C{0.4in}|C{0.4in}|C{0.4in}|C{0.4in}|C{0.4in}|C{0.4in}|C{0.4in}|C{0.4in}|}
		\hline
		Datasets & \multicolumn{4}{c|}{MNIST / AbsScene} & \multicolumn{4}{c|}{dSprites} & \multicolumn{4}{c|}{CLEVR / SHOP / GSO} \\ \hline
		Splits &   Train    &   Valid    &   Test 1   &   Test 2   &   Train    &   Valid    &   Test 1   &   Test 2 &   Train    &   Valid    &   Test 1   &   Test 2   \\ \hline
		\# of Images &   50000    &    1000    &    1000    &    1000    &   50000    &    1000    &    1000    &    1000    &   50000    &    1000    &    1000    &    1000    \\ \hline
		\# of Objects & 2 $\sim$ 4 & 2 $\sim$ 4 & 2 $\sim$ 4 & 5 $\sim$ 6 & 2 $\sim$ 5 & 2 $\sim$ 5 & 2 $\sim$ 5 & 6 $\sim$ 8 & 3 $\sim$ 6 & 3 $\sim$ 6 & 3 $\sim$ 6 & 7 $\sim$ 10 \\ \hline
		Image Size & \multicolumn{4}{c|}{64 $\times$ 64} & \multicolumn{4}{c|}{64 $\times$ 64} & \multicolumn{4}{c|}{128 $\times$ 128} \\ \hline
		Min Visible & \multicolumn{4}{c|}{25\%} & \multicolumn{4}{c|}{25\%} & \multicolumn{4}{c|}{128 pixels} \\ \hline
	\end{tabular}
\end{table*}

%% file: table_test_average.tex
\begin{table*}[!t]
    \small
	\centering
    \addtolength{\tabcolsep}{-1.2pt}
	\caption{The performance averaged across six datasets. The top-2 scores are underlined, with the best in bold and the second best in italics.}
	\label{tab:test_average}
	\begin{tabular}{c|r|C{0.55in}C{0.55in}C{0.55in}C{0.55in}C{0.55in}C{0.55in}C{0.55in}C{0.55in}}
		\toprule
		&  Method   &          AMI-A          &          ARI-A          &          AMI-O          &          ARI-O          &           IoU           &           F1            &           OCA           &           OOA           \\ \midrule
		\multirow{10}{*}{Test 1} & AIR \cite{Eslami2016Attend} & 0.380 & 0.397 & 0.845 & 0.827 & N/A & N/A & 0.549 & \underline{\textit{0.709}} \\
		& N-EM \cite{Greff2017Neural} & 0.208 & 0.233 & 0.341 & 0.282 & N/A & N/A & 0.013 & N/A \\
		& IODINE \cite{Greff2019Multi} & 0.638 & \underline{\textit{0.700}} & 0.772 & 0.752 & N/A & N/A & 0.487 & N/A \\
		& GMIOO \cite{Yuan2019Generative} & \underline{\textbf{0.738}} & \underline{\textbf{0.811}} & \underline{\textbf{0.916}} & \underline{\textbf{0.914}} & \underline{\textbf{0.708}} & \underline{\textbf{0.808}} & \underline{\textbf{0.772}} & \underline{\textbf{0.846}} \\
		& MONet \cite{Burgess2019MONet} & \underline{\textit{0.657}} & 0.699 & \underline{\textit{0.863}} & \underline{\textit{0.857}} & N/A & N/A & \underline{\textit{0.663}} & 0.583 \\
		& GENESIS \cite{Engelcke2020GENESIS} & 0.411 & 0.412 & 0.420 & 0.382 & 0.105 & 0.170 & 0.213 & 0.603 \\
		& SPACE \cite{Lin2020SPACE} & 0.640 & 0.678 & 0.817 & 0.765 & \underline{\textit{0.630}} & \underline{\textit{0.739}} & 0.436 & 0.666 \\
		& Slot Attention \cite{Locatello2020Object} & 0.393 & 0.321 & 0.758 & 0.711 & N/A & N/A & 0.028 & N/A \\
		& EfficientMORL \cite{Emami2021Efficient} & 0.341 & 0.279 & 0.673 & 0.621 & N/A & N/A & 0.107 & N/A \\
		& GENESIS-V2 \cite{Engelcke2021GENESIS} & 0.304 & 0.206 & 0.728 & 0.693 & N/A & N/A & 0.153 & 0.574 \\
		\midrule
		\multirow{10}{*}{Test 2} & AIR \cite{Eslami2016Attend} & 0.410 & 0.402 & 0.802 & 0.740 & N/A & N/A & 0.327 & \underline{\textit{0.689}} \\
		& N-EM \cite{Greff2017Neural} & 0.256 & 0.268 & 0.354 & 0.261 & N/A & N/A & 0.017 & N/A \\
		& IODINE \cite{Greff2019Multi} & 0.633 & 0.652 & 0.781 & 0.731 & N/A & N/A & 0.387 & N/A \\
		& GMIOO \cite{Yuan2019Generative} & \underline{\textbf{0.732}} & \underline{\textbf{0.781}} & \underline{\textbf{0.891}} & \underline{\textbf{0.868}} & \underline{\textbf{0.647}} & \underline{\textbf{0.746}} & \underline{\textbf{0.534}} & \underline{\textbf{0.823}} \\
		& MONet \cite{Burgess2019MONet} & \underline{\textit{0.635}} & \underline{\textit{0.665}} & \underline{\textit{0.820}} & \underline{\textit{0.785}} & N/A & N/A & \underline{\textit{0.446}} & 0.619 \\
		& GENESIS \cite{Engelcke2020GENESIS} & 0.380 & 0.378 & 0.415 & 0.315 & 0.076 & 0.132 & 0.160 & 0.584 \\
		& SPACE \cite{Lin2020SPACE} & 0.628 & 0.639 & 0.802 & 0.717 & \underline{\textit{0.543}} & \underline{\textit{0.654}} & 0.265 & 0.650 \\
		& Slot Attention \cite{Locatello2020Object} & 0.447 & 0.330 & 0.761 & 0.696 & N/A & N/A & 0.029 & N/A \\
		& EfficientMORL \cite{Emami2021Efficient} & 0.366 & 0.236 & 0.662 & 0.562 & N/A & N/A & 0.085 & N/A \\
		& GENESIS-V2 \cite{Engelcke2021GENESIS} & 0.378 & 0.235 & 0.723 & 0.655 & N/A & N/A & 0.189 & 0.617 \\
		\bottomrule
	\end{tabular}
\end{table*}

%% file: table_recon.tex
\begin{table*}[!t]
    \small
	\centering
    \addtolength{\tabcolsep}{-3.5pt}
    \renewcommand{\arraystretch}{0.985}
	\caption{Comparison of reconstruction errors (MSE) of the non-compositional (N-C) and compositional (C) versions of each method. The overall lengths of the distributed parts of representations extracted by two versions of the same method are identical.}
	\label{tab:recon}
	\begin{tabular}{r|c|C{0.78in}C{0.78in}C{0.78in}C{0.78in}C{0.78in}C{0.78in}|C{0.5in}}
		\toprule
		\multicolumn{9}{c}{Test 1} \\ \midrule
		Method & Ver. & MNIST & dSprites & AbsScene & CLEVR & SHOP & GSO & Average \\ \midrule
		\multirow{2}{*}{AIR \cite{Eslami2016Attend}} & N-C & 3.95e-3$\pm$6e-6 & 3.35e-3$\pm$7e-6 & 2.47e-3$\pm$7e-6 & 1.06e-3$\pm$2e-6 & 2.97e-3$\pm$4e-6 & 5.80e-3$\pm$1e-6 & 3.27e-3 \\
		& C & \textbf{1.98e-3}$\pm$9e-6 & \textbf{1.71e-3}$\pm$5e-6 & \textbf{1.52e-3}$\pm$6e-6 & \textbf{0.47e-3}$\pm$3e-6 & \textbf{1.39e-3}$\pm$5e-6 & \textbf{4.16e-3}$\pm$2e-6 & \textbf{1.87e-3} \\ \midrule
		\multirow{2}{*}{N-EM \cite{Greff2017Neural}} & N-C & 11.14e-3$\pm$5e-6 & 8.62e-3$\pm$1e-5 & \textbf{2.84e-3}$\pm$3e-6 & \textbf{1.95e-3}$\pm$2e-6 & \textbf{5.31e-3}$\pm$2e-6 & 9.57e-3$\pm$3e-6 & 6.57e-3 \\
		& C & \textbf{5.85e-3}$\pm$2e-5 & \textbf{2.31e-3}$\pm$5e-6 & 5.42e-3$\pm$7e-6 & 3.72e-3$\pm$2e-6 & 6.61e-3$\pm$7e-6 & \textbf{7.17e-3}$\pm$6e-6 & \textbf{5.18e-3} \\ \midrule
		\multirow{2}{*}{IODINE \cite{Greff2019Multi}} & N-C & 6.91e-3$\pm$3e-6 & 4.94e-3$\pm$3e-6 & 1.13e-3$\pm$2e-6 & 0.75e-3$\pm$2e-6 & 3.13e-3$\pm$2e-6 & 7.19e-3$\pm$1e-6 & 4.01e-3 \\
		& C & \textbf{4.23e-3}$\pm$4e-5 & \textbf{2.42e-3}$\pm$4e-5 & \textbf{0.39e-3}$\pm$4e-6 & \textbf{0.36e-3}$\pm$3e-6 & \textbf{1.99e-3}$\pm$2e-5 & \textbf{6.25e-3}$\pm$2e-5 & \textbf{2.61e-3} \\ \midrule
		\multirow{2}{*}{GMIOO \cite{Yuan2019Generative}} & N-C & 6.76e-3$\pm$1e-5 & 3.39e-3$\pm$2e-6 & 2.37e-3$\pm$5e-6 & 1.04e-3$\pm$3e-6 & 3.02e-3$\pm$2e-6 & 6.01e-3$\pm$2e-6 & 3.76e-3 \\
		& C & \textbf{1.82e-3}$\pm$8e-6 & \textbf{1.14e-3}$\pm$6e-6 & \textbf{0.49e-3}$\pm$6e-6 & \textbf{0.35e-3}$\pm$2e-6 & \textbf{1.04e-3}$\pm$2e-6 & \textbf{4.11e-3}$\pm$5e-6 & \textbf{1.49e-3} \\ \midrule
		\multirow{2}{*}{MONet \cite{Burgess2019MONet}} & N-C & 10.79e-3$\pm$4e-6 & 8.13e-3$\pm$3e-6 & 3.14e-3$\pm$3e-6 & 2.70e-3$\pm$8e-7 & 7.42e-3$\pm$2e-6 & 10.68e-3$\pm$2e-6 & 7.14e-3 \\
		& C & \textbf{5.13e-3}$\pm$3e-6 & \textbf{3.30e-3}$\pm$5e-5 & \textbf{1.08e-3}$\pm$1e-6 & \textbf{0.59e-3}$\pm$6e-7 & \textbf{3.45e-3}$\pm$2e-6 & \textbf{8.88e-3}$\pm$2e-6 & \textbf{3.74e-3} \\ \midrule
		\multirow{2}{*}{GENESIS \cite{Engelcke2020GENESIS}} & N-C & 10.53e-3$\pm$2e-8 & 8.01e-3$\pm$1e-8 & \textbf{3.42e-3}$\pm$2e-8 & \textbf{3.27e-3}$\pm$5e-6 & 7.32e-3$\pm$2e-8 & 11.18e-3$\pm$6e-9 & 7.29e-3 \\
		& C & \textbf{5.16e-3}$\pm$1e-5 & \textbf{3.69e-3}$\pm$9e-6 & \textbf{3.42e-3}$\pm$1e-5 & 3.30e-3$\pm$6e-6 & \textbf{3.85e-3}$\pm$5e-6 & \textbf{5.45e-3}$\pm$6e-8 & \textbf{4.15e-3} \\ \midrule
		\multirow{2}{*}{SPACE \cite{Lin2020SPACE}} & N-C & 11.77e-3$\pm$8e-6 & 6.36e-3$\pm$1e-5 & 3.96e-3$\pm$3e-6 & 3.03e-3$\pm$8e-7 & 6.97e-3$\pm$2e-6 & 6.80e-3$\pm$4e-6 & 6.48e-3 \\
		& C & \textbf{8.52e-3}$\pm$1e-5 & \textbf{2.31e-3}$\pm$2e-5 & \textbf{1.34e-3}$\pm$1e-5 & \textbf{0.41e-3}$\pm$3e-6 & \textbf{1.49e-3}$\pm$4e-6 & \textbf{2.61e-3}$\pm$1e-6 & \textbf{2.78e-3} \\ \midrule
		\multirow{2}{*}{Slot Attention \cite{Locatello2020Object}} & N-C & 10.62e-3$\pm$9e-9 & 8.98e-3$\pm$3e-8 & 2.93e-3$\pm$2e-9 & 1.04e-3$\pm$1e-9 & 3.56e-3$\pm$3e-8 & 8.23e-3$\pm$4e-8 & 5.89e-3 \\
		& C & \textbf{3.97e-3}$\pm$2e-5 & \textbf{3.39e-3}$\pm$2e-5 & \textbf{0.57e-3}$\pm$6e-6 & \textbf{0.26e-3}$\pm$5e-6 & \textbf{1.48e-3}$\pm$5e-6 & \textbf{4.68e-3}$\pm$9e-6 & \textbf{2.39e-3} \\ \midrule
		\multirow{2}{*}{EfficientMORL \cite{Emami2021Efficient}} & N-C & 8.64e-3$\pm$5e-6 & \textbf{5.77e-3}$\pm$5e-6 & 10.75e-3$\pm$5e-5 & 3.30e-3$\pm$4e-6 & 7.30e-3$\pm$2e-5 & 17.16e-3$\pm$3e-5 & 8.82e-3 \\
		& C & \textbf{6.32e-3}$\pm$2e-5 & 9.70e-3$\pm$2e-5 & \textbf{1.75e-3}$\pm$4e-5 & \textbf{0.88e-3}$\pm$1e-5 & \textbf{1.64e-3}$\pm$1e-5 & \textbf{6.93e-3}$\pm$2e-5 & \textbf{4.54e-3} \\ \midrule
		\multirow{2}{*}{GENESIS-V2 \cite{Engelcke2021GENESIS}} & N-C & 9.68e-3$\pm$1e-7 & 6.01e-3$\pm$2e-8 & 3.53e-3$\pm$1e-5 & \textbf{3.26e-3}$\pm$9e-6 & \textbf{3.19e-3}$\pm$2e-8 & 8.04e-3$\pm$2e-8 & 5.62e-3 \\
		& C & \textbf{7.90e-3}$\pm$3e-6 & \textbf{5.15e-3}$\pm$2e-5 & \textbf{3.27e-3}$\pm$3e-5 & \textbf{3.26e-3}$\pm$8e-6 & 3.32e-3$\pm$1e-5 & \textbf{4.30e-3}$\pm$2e-6 & \textbf{4.53e-3} \\ \bottomrule
	\end{tabular}  \\[0.02in]
	\begin{tabular}{r|c|C{0.78in}C{0.78in}C{0.78in}C{0.78in}C{0.78in}C{0.78in}|C{0.5in}}
		\toprule
		\multicolumn{9}{c}{Test 2} \\ \midrule
		Method & Ver. & MNIST & dSprites & AbsScene & CLEVR & SHOP & GSO & Average \\ \midrule
		\multirow{2}{*}{AIR \cite{Eslami2016Attend}} & N-C & 10.27e-3$\pm$7e-6 & 8.08e-3$\pm$5e-6 & 5.55e-3$\pm$8e-6 & 2.63e-3$\pm$3e-6 & 4.81e-3$\pm$4e-6 & 8.95e-3$\pm$3e-6 & 6.71e-3 \\
		& C & \textbf{4.29e-3}$\pm$2e-5 & \textbf{3.76e-3}$\pm$1e-5 & \textbf{3.74e-3}$\pm$1e-5 & \textbf{1.10e-3}$\pm$7e-6 & \textbf{2.23e-3}$\pm$4e-6 & \textbf{5.34e-3}$\pm$5e-6 & \textbf{3.41e-3} \\ \midrule
		\multirow{2}{*}{N-EM \cite{Greff2017Neural}} & N-C & \textbf{19.96e-3}$\pm$7e-6 & 18.63e-3$\pm$9e-6 & \textbf{6.82e-3}$\pm$3e-6 & \textbf{5.21e-3}$\pm$3e-6 & \textbf{8.58e-3}$\pm$3e-6 & \textbf{15.25e-3}$\pm$4e-6 & \textbf{12.41e-3} \\
		& C & 24.35e-3$\pm$9e-5 & \textbf{15.01e-3}$\pm$4e-5 & 9.97e-3$\pm$8e-6 & 7.60e-3$\pm$8e-6 & 10.49e-3$\pm$7e-6 & 15.64e-3$\pm$1e-5 & 13.84e-3 \\ \midrule
		\multirow{2}{*}{IODINE \cite{Greff2019Multi}} & N-C & 15.98e-3$\pm$8e-6 & 12.95e-3$\pm$4e-6 & 3.89e-3$\pm$4e-6 & 2.45e-3$\pm$8e-6 & 5.17e-3$\pm$3e-6 & 11.16e-3$\pm$2e-6 & 8.60e-3 \\
		& C & \textbf{8.72e-3}$\pm$5e-5 & \textbf{5.05e-3}$\pm$3e-5 & \textbf{1.06e-3}$\pm$1e-5 & \textbf{0.78e-3}$\pm$1e-5 & \textbf{2.78e-3}$\pm$1e-5 & \textbf{8.91e-3}$\pm$2e-5 & \textbf{4.55e-3} \\ \midrule
		\multirow{2}{*}{GMIOO \cite{Yuan2019Generative}} & N-C & 15.20e-3$\pm$2e-5 & 8.79e-3$\pm$5e-6 & 6.11e-3$\pm$7e-6 & 2.94e-3$\pm$4e-6 & 5.10e-3$\pm$2e-6 & 9.56e-3$\pm$4e-6 & 7.95e-3 \\
		& C & \textbf{3.65e-3}$\pm$1e-5 & \textbf{2.44e-3}$\pm$1e-5 & \textbf{1.10e-3}$\pm$1e-5 & \textbf{0.78e-3}$\pm$3e-6 & \textbf{1.76e-3}$\pm$4e-6 & \textbf{5.49e-3}$\pm$4e-6 & \textbf{2.54e-3} \\ \midrule
		\multirow{2}{*}{MONet \cite{Burgess2019MONet}} & N-C & 20.14e-3$\pm$3e-6 & 17.75e-3$\pm$5e-6 & 8.18e-3$\pm$2e-6 & 5.96e-3$\pm$2e-6 & 10.63e-3$\pm$3e-6 & 16.42e-3$\pm$6e-7 & 13.18e-3 \\
		& C & \textbf{10.81e-3}$\pm$4e-5 & \textbf{6.46e-3}$\pm$9e-5 & \textbf{3.92e-3}$\pm$8e-6 & \textbf{1.56e-3}$\pm$6e-5 & \textbf{4.78e-3}$\pm$2e-6 & \textbf{14.23e-3}$\pm$5e-5 & \textbf{6.96e-3} \\ \midrule
		\multirow{2}{*}{GENESIS \cite{Engelcke2020GENESIS}} & N-C & 18.49e-3$\pm$1e-8 & 16.07e-3$\pm$1e-8 & 7.79e-3$\pm$2e-8 & \textbf{6.10e-3}$\pm$8e-6 & 10.86e-3$\pm$2e-9 & 17.59e-3$\pm$9e-9 & 12.82e-3 \\
		& C & \textbf{12.68e-3}$\pm$1e-5 & \textbf{9.36e-3}$\pm$3e-5 & \textbf{6.86e-3}$\pm$2e-5 & 6.13e-3$\pm$4e-6 & \textbf{5.84e-3}$\pm$9e-6 & \textbf{8.60e-3}$\pm$8e-8 & \textbf{8.25e-3} \\ \midrule
		\multirow{2}{*}{SPACE \cite{Lin2020SPACE}} & N-C & 20.85e-3$\pm$1e-5 & 13.41e-3$\pm$3e-5 & 8.29e-3$\pm$6e-6 & 5.74e-3$\pm$1e-6 & 9.84e-3$\pm$1e-6 & 9.90e-3$\pm$6e-6 & 11.34e-3 \\
		& C & \textbf{15.12e-3}$\pm$2e-5 & \textbf{5.44e-3}$\pm$3e-5 & \textbf{3.23e-3}$\pm$1e-5 & \textbf{1.02e-3}$\pm$6e-6 & \textbf{2.33e-3}$\pm$7e-6 & \textbf{3.31e-3}$\pm$2e-6 & \textbf{5.08e-3} \\ \midrule
		\multirow{2}{*}{Slot Attention \cite{Locatello2020Object}} & N-C & 20.71e-3$\pm$6e-9 & 20.10e-3$\pm$4e-9 & 7.94e-3$\pm$4e-9 & 3.44e-3$\pm$7e-10 & 6.47e-3$\pm$4e-8 & 12.96e-3$\pm$5e-8 & 11.94e-3 \\
		& C & \textbf{9.62e-3}$\pm$8e-5 & \textbf{7.84e-3}$\pm$2e-5 & \textbf{1.99e-3}$\pm$2e-5 & \textbf{1.20e-3}$\pm$2e-5 & \textbf{2.63e-3}$\pm$1e-5 & \textbf{6.30e-3}$\pm$5e-6 & \textbf{4.93e-3} \\ \midrule
		\multirow{2}{*}{EfficientMORL \cite{Emami2021Efficient}} & N-C & 17.99e-3$\pm$6e-6 & \textbf{14.00e-3}$\pm$6e-6 & 17.29e-3$\pm$7e-5 & 6.18e-3$\pm$2e-5 & 10.44e-3$\pm$2e-5 & 24.17e-3$\pm$2e-5 & 15.01e-3 \\
		& C & \textbf{13.63e-3}$\pm$2e-5 & 18.76e-3$\pm$2e-5 & \textbf{5.22e-3}$\pm$4e-5 & \textbf{2.77e-3}$\pm$2e-5 & \textbf{3.13e-3}$\pm$3e-5 & \textbf{10.91e-3}$\pm$4e-5 & \textbf{9.07e-3} \\ \midrule
		\multirow{2}{*}{GENESIS-V2 \cite{Engelcke2021GENESIS}} & N-C & 17.89e-3$\pm$7e-8 & 11.71e-3$\pm$3e-8 & \textbf{7.07e-3}$\pm$3e-5 & 6.16e-3$\pm$1e-5 & 5.33e-3$\pm$3e-8 & 12.75e-3$\pm$4e-8 & 10.15e-3 \\
		& C & \textbf{13.63e-3}$\pm$1e-5 & \textbf{7.42e-3}$\pm$2e-5 & 7.12e-3$\pm$3e-5 & \textbf{5.94e-3}$\pm$2e-5 & \textbf{4.88e-3}$\pm$2e-5 & \textbf{7.21e-3}$\pm$4e-5 & \textbf{7.70e-3} \\ \bottomrule
	\end{tabular}
\end{table*}

%% file: table_train.tex
\begin{table*}[!t]
    \small
	\centering
	\addtolength{\tabcolsep}{-3.4pt}
	\renewcommand{\arraystretch}{0.985}
	\caption{Comparison of model complexities and training complexities. All the methods are tested on NVIDIA GeForce RTX 3090 (24G GPU Memory).}
	\label{tab:train}
	\begin{tabular}{c|r|C{0.7in}C{0.7in}C{0.7in}C{0.7in}C{0.7in}C{0.7in}C{0.7in}}
		\toprule
		         Dataset           &                                    Method & Parameters & Batch Size & Min GPUs & Mem / GPU & Train Steps & Time / Step & GPU Time  \\ \midrule
		 \multirow{10}{*}{MNIST}   &               AIR \cite{Eslami2016Attend} &  5.109 M   &     64     &    1     &  2.84 G   &    200 K    &   0.145 s   &  8.08 h   \\
		                           &               N-EM \cite{Greff2017Neural} &  0.882 M   &     64     &    1     &  11.60 G  &    50 K     &   0.686 s   &  9.53 h   \\
		                           &              IODINE \cite{Greff2019Multi} &  0.332 M   &     32     &    1     &  8.47 G   &   1000 K    &   0.955 s   & 265.29 h  \\
		                           &           GMIOO \cite{Yuan2019Generative} &  12.306 M  &     64     &    1     &  5.12 G   &    200 K    &   0.309 s   &  17.18 h  \\
		                           &             MONet \cite{Burgess2019MONet} &  1.020 M   &     64     &    1     &  5.37 G   &   1000 K    &   0.122 s   &  33.88 h  \\
		                           &        GENESIS \cite{Engelcke2020GENESIS} &  12.621 M  &     32     &    1     &  6.34 G   &    500 K    &   0.147 s   &  20.45 h  \\
		                           &                 SPACE \cite{Lin2020SPACE} &  4.902 M   &     16     &    1     &  4.92 G   &    200 K    &   0.172 s   &  9.58 h   \\
		                           & Slot Attention \cite{Locatello2020Object} &  0.236 M   &     64     &    1     &  9.85 G   &    500 K    &   0.107 s   &  14.88 h  \\
		                           &   EfficientMORL \cite{Emami2021Efficient} &  0.666 M   &     32     &    1     &  12.09 G  &    300 K    &   0.572 s   &  47.70 h  \\
		                           &     GENESIS-V2 \cite{Engelcke2021GENESIS} &  2.729 M   &     32     &    1     &  4.23 G   &    500 K    &   0.128 s   &  17.84 h  \\ \midrule
		\multirow{10}{*}{dSprites} &               AIR \cite{Eslami2016Attend} &  5.109 M   &     64     &    1     &  2.99 G   &    200 K    &   0.161 s   &  8.93 h   \\
		                           &               N-EM \cite{Greff2017Neural} &  0.882 M   &     64     &    1     &  13.14 G  &    50 K     &   0.804 s   &  11.16 h  \\
		                           &              IODINE \cite{Greff2019Multi} &  0.332 M   &     32     &    1     &  12.57 G  &   1000 K    &   1.125 s   & 312.39 h  \\
		                           &           GMIOO \cite{Yuan2019Generative} &  12.306 M  &     64     &    1     &  5.85 G   &    200 K    &   0.352 s   &  19.53 h  \\
		                           &             MONet \cite{Burgess2019MONet} &  1.020 M   &     64     &    1     &  9.47 G   &   1000 K    &   0.144 s   &  39.95 h  \\
		                           &        GENESIS \cite{Engelcke2020GENESIS} &  12.621 M  &     32     &    1     &  7.16 G   &    500 K    &   0.167 s   &  23.24 h  \\
		                           &                 SPACE \cite{Lin2020SPACE} &  4.902 M   &     16     &    1     &  4.93 G   &    200 K    &   0.172 s   &  9.58 h   \\
		                           & Slot Attention \cite{Locatello2020Object} &  0.236 M   &     64     &    1     &  9.85 G   &    500 K    &   0.116 s   &  16.13 h  \\
		                           &   EfficientMORL \cite{Emami2021Efficient} &  0.666 M   &     32     &    1     &  13.69 G  &    300 K    &   0.656 s   &  54.70 h  \\
		                           &     GENESIS-V2 \cite{Engelcke2021GENESIS} &  2.729 M   &     32     &    1     &  4.47 G   &    500 K    &   0.146 s   &  20.23 h  \\ \midrule
		\multirow{10}{*}{AbsScene} &               AIR \cite{Eslami2016Attend} &  5.109 M   &     64     &    1     &  2.84 G   &    200 K    &   0.144 s   &  8.01 h   \\
		                           &               N-EM \cite{Greff2017Neural} &  0.882 M   &     64     &    1     &  11.60 G  &    50 K     &   0.687 s   &  9.55 h   \\
		                           &              IODINE \cite{Greff2019Multi} &  0.332 M   &     32     &    1     &  8.47 G   &   1000 K    &   0.953 s   & 264.78 h  \\
		                           &           GMIOO \cite{Yuan2019Generative} &  12.818 M  &     64     &    1     &  5.38 G   &    200 K    &   0.321 s   &  17.85 h  \\
		                           &             MONet \cite{Burgess2019MONet} &  1.020 M   &     64     &    1     &  5.37 G   &   1000 K    &   0.122 s   &  33.90 h  \\
		                           &        GENESIS \cite{Engelcke2020GENESIS} &  12.621 M  &     32     &    1     &  6.34 G   &    500 K    &   0.147 s   &  20.46 h  \\
		                           &                 SPACE \cite{Lin2020SPACE} &  4.833 M   &     16     &    1     &  4.91 G   &    200 K    &   0.173 s   &  9.61 h   \\
		                           & Slot Attention \cite{Locatello2020Object} &  0.236 M   &     64     &    1     &  9.85 G   &    500 K    &   0.107 s   &  14.82 h  \\
		                           &   EfficientMORL \cite{Emami2021Efficient} &  0.666 M   &     32     &    1     &  12.09 G  &    300 K    &   0.574 s   &  47.84 h  \\
		                           &     GENESIS-V2 \cite{Engelcke2021GENESIS} &  2.729 M   &     32     &    1     &  4.23 G   &    500 K    &   0.128 s   &  17.72 h  \\ \midrule
		 \multirow{10}{*}{CLEVR}   &               AIR \cite{Eslami2016Attend} &  5.117 M   &     64     &    1     &  5.54 G   &    200 K    &   0.220 s   &  12.20 h  \\
		                           &               N-EM \cite{Greff2017Neural} &  2.529 M   &     64     &    4     &  14.29 G  &    50 K     &   0.621 s   &  34.51 h  \\
		                           &              IODINE \cite{Greff2019Multi} &  1.109 M   &     32     &    4     &  18.00 G  &   1000 K    &   1.254 s   & 1392.85 h \\
		                           &           GMIOO \cite{Yuan2019Generative} &  12.830 M  &     64     &    1     &  15.34 G  &    200 K    &   0.618 s   &  34.33 h  \\
		                           &             MONet \cite{Burgess2019MONet} &  1.834 M   &     64     &    1     &  20.73 G  &   1000 K    &   0.516 s   & 143.26 h  \\
		                           &        GENESIS \cite{Engelcke2020GENESIS} &  13.407 M  &     32     &    1     &  17.31 G  &    500 K    &   0.316 s   &  43.85 h  \\
		                           &                 SPACE \cite{Lin2020SPACE} &  7.275 M   &     16     &    1     &  10.70 G  &    200 K    &   0.384 s   &  21.32 h  \\
		                           & Slot Attention \cite{Locatello2020Object} &  0.890 M   &     64     &    2     &  18.07 G  &    500 K    &   0.278 s   &  77.09 h  \\
		                           &   EfficientMORL \cite{Emami2021Efficient} &  0.666 M   &     32     &    4     &  13.90 G  &    300 K    &   0.766 s   & 255.43 h  \\
		                           &     GENESIS-V2 \cite{Engelcke2021GENESIS} &  2.840 M   &     32     &    1     &  12.30 G  &    500 K    &   0.329 s   &  45.67 h  \\ \midrule
		  \multirow{10}{*}{SHOP}   &               AIR \cite{Eslami2016Attend} &  7.256 M   &     64     &    1     &  5.68 G   &    200 K    &   0.219 s   &  12.16 h  \\
		                           &               N-EM \cite{Greff2017Neural} &  2.529 M   &     64     &    4     &  14.29 G  &    50 K     &   0.623 s   &  34.61 h  \\
		                           &              IODINE \cite{Greff2019Multi} &  1.109 M   &     32     &    4     &  18.00 G  &   1000 K    &   1.253 s   & 1391.76 h \\
		                           &           GMIOO \cite{Yuan2019Generative} &  14.969 M  &     64     &    1     &  15.74 G  &    200 K    &   0.628 s   &  34.91 h  \\
		                           &             MONet \cite{Burgess2019MONet} &  1.834 M   &     64     &    1     &  20.73 G  &   1000 K    &   0.516 s   & 143.26 h  \\
		                           &        GENESIS \cite{Engelcke2020GENESIS} &  13.407 M  &     32     &    1     &  17.31 G  &    500 K    &   0.315 s   &  43.77 h  \\
		                           &                 SPACE \cite{Lin2020SPACE} &  7.303 M   &     16     &    1     &  11.06 G  &    200 K    &   0.392 s   &  21.77 h  \\
		                           & Slot Attention \cite{Locatello2020Object} &  0.890 M   &     64     &    2     &  18.07 G  &    500 K    &   0.278 s   &  77.24 h  \\
		                           &   EfficientMORL \cite{Emami2021Efficient} &  0.666 M   &     32     &    4     &  13.90 G  &    300 K    &   0.767 s   & 255.53 h  \\
		                           &     GENESIS-V2 \cite{Engelcke2021GENESIS} &  2.840 M   &     32     &    1     &  12.30 G  &    500 K    &   0.332 s   &  46.11 h  \\ \midrule
		  \multirow{10}{*}{GSO}    &               AIR \cite{Eslami2016Attend} &  7.256 M   &     64     &    1     &  5.68 G   &    200 K    &   0.221 s   &  12.29 h  \\
		                           &               N-EM \cite{Greff2017Neural} &  2.529 M   &     64     &    4     &  14.29 G  &    50 K     &   0.622 s   &  34.57 h  \\
		                           &              IODINE \cite{Greff2019Multi} &  1.109 M   &     32     &    4     &  18.00 G  &   1000 K    &   1.247 s   & 1385.19 h \\
		                           &           GMIOO \cite{Yuan2019Generative} &  14.969 M  &     64     &    1     &  15.74 G  &    200 K    &   0.623 s   &  34.63 h  \\
		                           &             MONet \cite{Burgess2019MONet} &  1.834 M   &     64     &    1     &  20.73 G  &   1000 K    &   0.516 s   & 143.27 h  \\
		                           &        GENESIS \cite{Engelcke2020GENESIS} &  13.407 M  &     32     &    1     &  17.31 G  &    500 K    &   0.316 s   &  43.92 h  \\
		                           &                 SPACE \cite{Lin2020SPACE} &  7.303 M   &     16     &    1     &  11.06 G  &    200 K    &   0.394 s   &  21.88 h  \\
		                           & Slot Attention \cite{Locatello2020Object} &  0.890 M   &     64     &    2     &  18.07 G  &    500 K    &   0.276 s   &  76.77 h  \\
		                           &   EfficientMORL \cite{Emami2021Efficient} &  0.666 M   &     32     &    4     &  13.90 G  &    300 K    &   0.767 s   & 255.56 h  \\
		                           &     GENESIS-V2 \cite{Engelcke2021GENESIS} &  2.840 M   &     32     &    1     &  12.30 G  &    500 K    &   0.329 s   &  45.74 h  \\ \bottomrule
	\end{tabular}
\end{table*}

%% file: table_test1.tex
\begin{table*}[!t]
    \small
	\centering
    \addtolength{\tabcolsep}{-3.9pt}
    \renewcommand{\arraystretch}{0.985}
	\caption{Performance comparison on the Test 1 split. The top-2 scores are underlined, with the best in bold and the second best in italics.}
	\label{tab:test_1}
	\begin{tabular}{c|r|C{0.62in}C{0.62in}C{0.62in}C{0.62in}C{0.62in}C{0.62in}C{0.62in}C{0.62in}}
		\toprule
		Dataset          &  Method   &          AMI-A          &          ARI-A          &          AMI-O          &          ARI-O          &           IoU           &           F1            &           OCA           &           OOA           \\ \midrule
		\multirow{10}{*}{MNIST} & AIR \cite{Eslami2016Attend} & 0.439$\pm$6e-4 & 0.496$\pm$7e-4 & \underline{\textit{0.871}}$\pm$7e-4 & \underline{\textit{0.868}}$\pm$1e-3 & N/A & N/A & 0.796$\pm$5e-3 & \underline{\textit{0.563}}$\pm$4e-3 \\
		& N-EM \cite{Greff2017Neural} & 0.328$\pm$4e-3 & 0.307$\pm$6e-3 & 0.720$\pm$1e-3 & 0.684$\pm$2e-3 & N/A & N/A & 0.000$\pm$0e-0 & N/A \\
		& IODINE \cite{Greff2019Multi} & 0.564$\pm$1e-3 & 0.714$\pm$1e-3 & 0.628$\pm$2e-3 & 0.608$\pm$3e-3 & N/A & N/A & 0.286$\pm$1e-2 & N/A \\
		& GMIOO \cite{Yuan2019Generative} & \underline{\textbf{0.673}}$\pm$6e-4 & \underline{\textbf{0.789}}$\pm$6e-4 & \underline{\textbf{0.878}}$\pm$2e-3 & \underline{\textbf{0.889}}$\pm$2e-3 & \underline{\textbf{0.644}}$\pm$1e-3 & \underline{\textbf{0.775}}$\pm$1e-3 & \underline{\textit{0.857}}$\pm$8e-3 & \underline{\textbf{0.727}}$\pm$7e-3 \\
		& MONet \cite{Burgess2019MONet} & 0.516$\pm$2e-4 & 0.672$\pm$2e-4 & 0.830$\pm$5e-4 & 0.849$\pm$4e-4 & N/A & N/A & \underline{\textbf{0.912}}$\pm$7e-4 & 0.481$\pm$2e-3 \\
		& GENESIS \cite{Engelcke2020GENESIS} & \underline{\textit{0.578}}$\pm$5e-4 & \underline{\textit{0.749}}$\pm$3e-4 & 0.597$\pm$2e-3 & 0.593$\pm$2e-3 & 0.077$\pm$3e-4 & 0.138$\pm$4e-4 & 0.365$\pm$6e-3 & 0.448$\pm$5e-3 \\
		& SPACE \cite{Lin2020SPACE} & 0.487$\pm$1e-3 & 0.597$\pm$2e-3 & 0.712$\pm$8e-4 & 0.687$\pm$7e-4 & \underline{\textit{0.463}}$\pm$5e-4 & \underline{\textit{0.604}}$\pm$5e-4 & 0.552$\pm$3e-3 & 0.531$\pm$1e-2 \\
		& Slot Attention \cite{Locatello2020Object} & 0.561$\pm$7e-4 & 0.736$\pm$9e-4 & 0.528$\pm$1e-3 & 0.418$\pm$3e-3 & N/A & N/A & 0.019$\pm$2e-3 & N/A \\
		& EfficientMORL \cite{Emami2021Efficient} & 0.093$\pm$2e-4 & 0.002$\pm$2e-4 & 0.535$\pm$9e-4 & 0.465$\pm$2e-3 & N/A & N/A & 0.000$\pm$0e-0 & N/A \\
		& GENESIS-V2 \cite{Engelcke2021GENESIS} & 0.150$\pm$8e-5 & 0.127$\pm$2e-4 & 0.417$\pm$2e-4 & 0.330$\pm$2e-4 & N/A & N/A & 0.019$\pm$2e-3 & 0.498$\pm$2e-2 \\
		\midrule
		\multirow{10}{*}{dSprites} & AIR \cite{Eslami2016Attend} & 0.601$\pm$3e-4 & 0.599$\pm$4e-4 & \underline{\textit{0.908}}$\pm$5e-4 & 0.899$\pm$7e-4 & N/A & N/A & 0.732$\pm$1e-2 & \underline{\textit{0.726}}$\pm$2e-3 \\
		& N-EM \cite{Greff2017Neural} & 0.559$\pm$6e-3 & 0.592$\pm$9e-3 & 0.754$\pm$2e-3 & 0.642$\pm$2e-3 & N/A & N/A & 0.000$\pm$0e-0 & N/A \\
		& IODINE \cite{Greff2019Multi} & 0.850$\pm$1e-3 & 0.930$\pm$7e-4 & 0.834$\pm$3e-3 & 0.875$\pm$3e-3 & N/A & N/A & 0.837$\pm$9e-3 & N/A \\
		& GMIOO \cite{Yuan2019Generative} & \underline{\textbf{0.923}}$\pm$6e-4 & \underline{\textbf{0.969}}$\pm$3e-4 & \underline{\textbf{0.953}}$\pm$1e-3 & \underline{\textbf{0.962}}$\pm$2e-3 & \underline{\textbf{0.868}}$\pm$1e-3 & \underline{\textbf{0.918}}$\pm$1e-3 & \underline{\textbf{0.899}}$\pm$4e-3 & \underline{\textbf{0.907}}$\pm$1e-2 \\
		& MONet \cite{Burgess2019MONet} & 0.798$\pm$3e-4 & 0.900$\pm$3e-4 & 0.907$\pm$2e-4 & \underline{\textit{0.925}}$\pm$2e-4 & N/A & N/A & \underline{\textit{0.890}}$\pm$3e-3 & 0.630$\pm$1e-3 \\
		& GENESIS \cite{Engelcke2020GENESIS} & 0.818$\pm$6e-4 & 0.918$\pm$4e-4 & 0.761$\pm$1e-3 & 0.798$\pm$1e-3 & 0.078$\pm$2e-4 & 0.141$\pm$3e-4 & 0.502$\pm$7e-3 & 0.427$\pm$3e-3 \\
		& SPACE \cite{Lin2020SPACE} & \underline{\textit{0.865}}$\pm$5e-4 & \underline{\textit{0.941}}$\pm$4e-4 & 0.869$\pm$4e-4 & 0.860$\pm$1e-3 & \underline{\textit{0.745}}$\pm$5e-4 & \underline{\textit{0.827}}$\pm$5e-4 & 0.492$\pm$6e-3 & 0.569$\pm$1e-2 \\
		& Slot Attention \cite{Locatello2020Object} & 0.275$\pm$7e-4 & 0.152$\pm$5e-4 & 0.717$\pm$1e-3 & 0.694$\pm$2e-3 & N/A & N/A & 0.000$\pm$0e-0 & N/A \\
		& EfficientMORL \cite{Emami2021Efficient} & 0.134$\pm$5e-4 & 0.061$\pm$1e-3 & 0.451$\pm$3e-3 & 0.375$\pm$4e-3 & N/A & N/A & 0.000$\pm$1e-3 & N/A \\
		& GENESIS-V2 \cite{Engelcke2021GENESIS} & 0.545$\pm$3e-3 & 0.583$\pm$4e-3 & 0.858$\pm$1e-3 & 0.868$\pm$2e-3 & N/A & N/A & 0.308$\pm$1e-2 & 0.565$\pm$8e-3 \\
		\midrule
		\multirow{10}{*}{AbsScene} & AIR \cite{Eslami2016Attend} & 0.404$\pm$4e-4 & 0.453$\pm$5e-4 & 0.676$\pm$1e-3 & 0.636$\pm$1e-3 & N/A & N/A & 0.415$\pm$3e-3 & 0.639$\pm$3e-3 \\
		& N-EM \cite{Greff2017Neural} & 0.061$\pm$1e-3 & 0.095$\pm$2e-3 & 0.068$\pm$5e-4 & 0.051$\pm$1e-3 & N/A & N/A & 0.025$\pm$4e-3 & N/A \\
		& IODINE \cite{Greff2019Multi} & \underline{\textbf{0.799}}$\pm$3e-4 & \underline{\textbf{0.872}}$\pm$2e-4 & \underline{\textbf{0.956}}$\pm$4e-4 & \underline{\textbf{0.971}}$\pm$3e-4 & N/A & N/A & \underline{\textit{0.940}}$\pm$5e-3 & N/A \\
		& GMIOO \cite{Yuan2019Generative} & \underline{\textit{0.766}}$\pm$4e-4 & \underline{\textit{0.845}}$\pm$4e-4 & \underline{\textit{0.932}}$\pm$6e-4 & \underline{\textit{0.946}}$\pm$1e-3 & \underline{\textbf{0.765}}$\pm$1e-3 & \underline{\textbf{0.858}}$\pm$1e-3 & \underline{\textbf{0.957}}$\pm$2e-3 & \underline{\textbf{0.961}}$\pm$1e-3 \\
		& MONet \cite{Burgess2019MONet} & 0.570$\pm$1e-4 & 0.346$\pm$6e-5 & 0.864$\pm$4e-4 & 0.896$\pm$3e-4 & N/A & N/A & 0.086$\pm$1e-3 & 0.358$\pm$2e-3 \\
		& GENESIS \cite{Engelcke2020GENESIS} & 0.155$\pm$3e-4 & 0.033$\pm$1e-4 & 0.398$\pm$1e-3 & 0.350$\pm$9e-4 & 0.094$\pm$1e-4 & 0.168$\pm$2e-4 & 0.267$\pm$6e-3 & 0.616$\pm$5e-3 \\
		& SPACE \cite{Lin2020SPACE} & 0.751$\pm$4e-4 & 0.834$\pm$5e-4 & 0.800$\pm$2e-3 & 0.795$\pm$3e-3 & \underline{\textit{0.642}}$\pm$6e-4 & \underline{\textit{0.727}}$\pm$6e-4 & 0.564$\pm$3e-3 & \underline{\textit{0.741}}$\pm$8e-3 \\
		& Slot Attention \cite{Locatello2020Object} & 0.758$\pm$3e-3 & 0.807$\pm$6e-3 & 0.900$\pm$7e-4 & 0.893$\pm$8e-4 & N/A & N/A & 0.146$\pm$1e-2 & N/A \\
		& EfficientMORL \cite{Emami2021Efficient} & 0.473$\pm$1e-3 & 0.495$\pm$1e-3 & 0.752$\pm$3e-3 & 0.727$\pm$4e-3 & N/A & N/A & 0.231$\pm$2e-2 & N/A \\
		& GENESIS-V2 \cite{Engelcke2021GENESIS} & 0.472$\pm$2e-3 & 0.417$\pm$4e-3 & 0.838$\pm$1e-3 & 0.855$\pm$1e-3 & N/A & N/A & 0.131$\pm$6e-3 & 0.571$\pm$1e-2 \\
		\midrule
		\multirow{10}{*}{CLEVR} & AIR \cite{Eslami2016Attend} & 0.082$\pm$3e-4 & 0.080$\pm$2e-4 & 0.979$\pm$4e-4 & 0.983$\pm$5e-4 & N/A & N/A & \underline{\textit{0.924}}$\pm$4e-3 & 0.920$\pm$4e-3 \\
		& N-EM \cite{Greff2017Neural} & 0.095$\pm$2e-3 & 0.130$\pm$3e-3 & 0.105$\pm$1e-3 & 0.055$\pm$2e-3 & N/A & N/A & 0.035$\pm$4e-3 & N/A \\
		& IODINE \cite{Greff2019Multi} & 0.540$\pm$1e-3 & 0.536$\pm$2e-3 & 0.970$\pm$9e-4 & 0.973$\pm$8e-4 & N/A & N/A & 0.499$\pm$1e-2 & N/A \\
		& GMIOO \cite{Yuan2019Generative} & 0.716$\pm$6e-4 & 0.776$\pm$7e-4 & 0.977$\pm$5e-4 & 0.979$\pm$9e-4 & \underline{\textit{0.696}}$\pm$2e-3 & \underline{\textit{0.809}}$\pm$2e-3 & 0.891$\pm$3e-3 & \underline{\textit{0.932}}$\pm$4e-3 \\
		& MONet \cite{Burgess2019MONet} & \underline{\textbf{0.882}}$\pm$9e-5 & \underline{\textbf{0.934}}$\pm$6e-5 & \underline{\textbf{0.985}}$\pm$4e-5 & \underline{\textbf{0.989}}$\pm$3e-5 & N/A & N/A & \underline{\textbf{0.966}}$\pm$8e-4 & 0.712$\pm$5e-4 \\
		& GENESIS \cite{Engelcke2020GENESIS} & 0.391$\pm$9e-4 & 0.502$\pm$8e-4 & 0.263$\pm$2e-3 & 0.218$\pm$2e-3 & 0.117$\pm$5e-4 & 0.174$\pm$4e-4 & 0.143$\pm$5e-3 & 0.737$\pm$1e-2 \\
		& SPACE \cite{Lin2020SPACE} & \underline{\textit{0.795}}$\pm$3e-4 & \underline{\textit{0.859}}$\pm$3e-4 & 0.972$\pm$1e-4 & 0.975$\pm$3e-4 & \underline{\textbf{0.775}}$\pm$7e-4 & \underline{\textbf{0.862}}$\pm$7e-4 & 0.710$\pm$2e-3 & \underline{\textbf{0.936}}$\pm$7e-3 \\
		& Slot Attention \cite{Locatello2020Object} & 0.240$\pm$3e-4 & 0.026$\pm$2e-4 & \underline{\textit{0.982}}$\pm$3e-4 & \underline{\textit{0.984}}$\pm$6e-4 & N/A & N/A & 0.002$\pm$1e-3 & N/A \\
		& EfficientMORL \cite{Emami2021Efficient} & 0.565$\pm$2e-3 & 0.616$\pm$3e-3 & 0.826$\pm$4e-3 & 0.779$\pm$6e-3 & N/A & N/A & 0.284$\pm$1e-2 & N/A \\
		& GENESIS-V2 \cite{Engelcke2021GENESIS} & 0.190$\pm$6e-4 & 0.023$\pm$8e-4 & 0.639$\pm$2e-3 & 0.585$\pm$2e-3 & N/A & N/A & 0.206$\pm$8e-3 & 0.712$\pm$2e-2 \\
		\midrule
		\multirow{10}{*}{SHOP} & AIR \cite{Eslami2016Attend} & 0.511$\pm$2e-4 & 0.467$\pm$3e-4 & 0.896$\pm$3e-4 & 0.904$\pm$4e-4 & N/A & N/A & 0.308$\pm$1e-2 & 0.672$\pm$2e-3 \\
		& N-EM \cite{Greff2017Neural} & 0.116$\pm$9e-4 & 0.184$\pm$2e-3 & 0.145$\pm$7e-4 & 0.082$\pm$1e-3 & N/A & N/A & 0.009$\pm$3e-3 & N/A \\
		& IODINE \cite{Greff2019Multi} & 0.714$\pm$1e-3 & 0.723$\pm$3e-3 & 0.837$\pm$1e-3 & 0.795$\pm$2e-3 & N/A & N/A & 0.344$\pm$6e-3 & N/A \\
		& GMIOO \cite{Yuan2019Generative} & \underline{\textit{0.776}}$\pm$2e-4 & \underline{\textit{0.829}}$\pm$2e-4 & 0.958$\pm$2e-4 & 0.962$\pm$4e-4 & \underline{\textbf{0.754}}$\pm$6e-4 & \underline{\textbf{0.843}}$\pm$7e-4 & \underline{\textit{0.669}}$\pm$5e-3 & \underline{\textbf{0.940}}$\pm$6e-3 \\
		& MONet \cite{Burgess2019MONet} & \underline{\textbf{0.784}}$\pm$2e-4 & \underline{\textbf{0.854}}$\pm$3e-4 & 0.941$\pm$1e-4 & 0.946$\pm$2e-4 & N/A & N/A & \underline{\textbf{0.861}}$\pm$2e-3 & \underline{\textit{0.881}}$\pm$2e-3 \\
		& GENESIS \cite{Engelcke2020GENESIS} & 0.316$\pm$3e-4 & 0.102$\pm$2e-4 & 0.265$\pm$2e-4 & 0.158$\pm$5e-4 & 0.169$\pm$5e-4 & 0.250$\pm$7e-4 & 0.000$\pm$0e-0 & 0.749$\pm$8e-3 \\
		& SPACE \cite{Lin2020SPACE} & 0.757$\pm$3e-4 & 0.827$\pm$3e-4 & 0.902$\pm$5e-4 & 0.897$\pm$6e-4 & \underline{\textit{0.742}}$\pm$4e-4 & \underline{\textit{0.840}}$\pm$4e-4 & 0.297$\pm$2e-3 & 0.759$\pm$4e-3 \\
		& Slot Attention \cite{Locatello2020Object} & 0.388$\pm$6e-4 & 0.159$\pm$1e-3 & \underline{\textit{0.962}}$\pm$3e-4 & \underline{\textit{0.969}}$\pm$4e-4 & N/A & N/A & 0.000$\pm$5e-4 & N/A \\
		& EfficientMORL \cite{Emami2021Efficient} & 0.502$\pm$1e-3 & 0.319$\pm$3e-3 & \underline{\textbf{0.964}}$\pm$7e-4 & \underline{\textbf{0.973}}$\pm$1e-3 & N/A & N/A & 0.040$\pm$3e-3 & N/A \\
		& GENESIS-V2 \cite{Engelcke2021GENESIS} & 0.314$\pm$3e-4 & 0.067$\pm$3e-4 & 0.930$\pm$9e-4 & 0.937$\pm$9e-4 & N/A & N/A & 0.001$\pm$6e-4 & 0.641$\pm$2e-2 \\
		\midrule
		\multirow{10}{*}{GSO} & AIR \cite{Eslami2016Attend} & 0.243$\pm$3e-4 & 0.285$\pm$4e-4 & \underline{\textit{0.738}}$\pm$6e-4 & \underline{\textit{0.672}}$\pm$1e-3 & N/A & N/A & 0.120$\pm$4e-3 & \underline{\textbf{0.735}}$\pm$6e-3 \\
		& N-EM \cite{Greff2017Neural} & 0.091$\pm$6e-4 & 0.087$\pm$1e-3 & 0.256$\pm$1e-3 & 0.180$\pm$2e-3 & N/A & N/A & 0.006$\pm$2e-3 & N/A \\
		& IODINE \cite{Greff2019Multi} & 0.360$\pm$2e-3 & 0.423$\pm$4e-3 & 0.406$\pm$3e-3 & 0.291$\pm$3e-3 & N/A & N/A & 0.015$\pm$4e-3 & N/A \\
		& GMIOO \cite{Yuan2019Generative} & \underline{\textbf{0.576}}$\pm$5e-4 & \underline{\textbf{0.658}}$\pm$8e-4 & \underline{\textbf{0.798}}$\pm$1e-3 & \underline{\textbf{0.745}}$\pm$2e-3 & \underline{\textbf{0.524}}$\pm$8e-4 & \underline{\textbf{0.647}}$\pm$9e-4 & \underline{\textbf{0.358}}$\pm$9e-3 & 0.607$\pm$2e-2 \\
		& MONet \cite{Burgess2019MONet} & \underline{\textit{0.392}}$\pm$3e-4 & \underline{\textit{0.489}}$\pm$3e-4 & 0.651$\pm$5e-4 & 0.540$\pm$7e-4 & N/A & N/A & \underline{\textit{0.262}}$\pm$2e-3 & 0.437$\pm$7e-3 \\
		& GENESIS \cite{Engelcke2020GENESIS} & 0.208$\pm$2e-6 & 0.166$\pm$6e-6 & 0.234$\pm$1e-5 & 0.176$\pm$2e-5 & 0.098$\pm$7e-6 & 0.152$\pm$1e-5 & 0.000$\pm$0e-0 & \underline{\textit{0.639}}$\pm$3e-4 \\
		& SPACE \cite{Lin2020SPACE} & 0.184$\pm$5e-5 & 0.012$\pm$1e-5 & 0.645$\pm$1e-4 & 0.377$\pm$3e-4 & \underline{\textit{0.411}}$\pm$8e-4 & \underline{\textit{0.574}}$\pm$8e-4 & 0.000$\pm$0e-0 & 0.460$\pm$2e-2 \\
		& Slot Attention \cite{Locatello2020Object} & 0.134$\pm$9e-5 & 0.047$\pm$2e-4 & 0.462$\pm$2e-4 & 0.307$\pm$5e-4 & N/A & N/A & 0.000$\pm$5e-4 & N/A \\
		& EfficientMORL \cite{Emami2021Efficient} & 0.276$\pm$1e-3 & 0.179$\pm$2e-3 & 0.509$\pm$3e-3 & 0.409$\pm$4e-3 & N/A & N/A & 0.088$\pm$7e-3 & N/A \\
		& GENESIS-V2 \cite{Engelcke2021GENESIS} & 0.152$\pm$1e-5 & 0.020$\pm$2e-5 & 0.689$\pm$8e-5 & 0.582$\pm$2e-4 & N/A & N/A & 0.255$\pm$4e-3 & 0.457$\pm$4e-2 \\
		\bottomrule
	\end{tabular}
\end{table*}

%% file: table_test2.tex
\begin{table*}[!t]
    \small
	\centering
	\addtolength{\tabcolsep}{-3.9pt}
	\renewcommand{\arraystretch}{0.985}
	\caption{Performance comparison on the Test 2 split. The top-2 scores are underlined, with the best in bold and the second best in italics.}
	\label{tab:test_2}
	\begin{tabular}{c|r|C{0.62in}C{0.62in}C{0.62in}C{0.62in}C{0.62in}C{0.62in}C{0.62in}C{0.62in}}
		\toprule
		Dataset          &  Method   &          AMI-A          &          ARI-A          &          AMI-O          &          ARI-O          &           IoU           &           F1            &           OCA           &           OOA           \\ \midrule
		\multirow{10}{*}{MNIST} & AIR \cite{Eslami2016Attend} & \underline{\textit{0.526}}$\pm$3e-4 & 0.572$\pm$6e-4 & \underline{\textit{0.827}}$\pm$8e-4 & \underline{\textit{0.781}}$\pm$1e-3 & N/A & N/A & 0.526$\pm$8e-3 & 0.537$\pm$4e-3 \\
		& N-EM \cite{Greff2017Neural} & 0.317$\pm$2e-3 & 0.271$\pm$3e-3 & 0.553$\pm$2e-3 & 0.457$\pm$2e-3 & N/A & N/A & 0.006$\pm$2e-3 & N/A \\
		& IODINE \cite{Greff2019Multi} & 0.518$\pm$2e-3 & \underline{\textit{0.633}}$\pm$2e-3 & 0.650$\pm$2e-3 & 0.562$\pm$4e-3 & N/A & N/A & 0.365$\pm$9e-3 & N/A \\
		& GMIOO \cite{Yuan2019Generative} & \underline{\textbf{0.661}}$\pm$5e-4 & \underline{\textbf{0.761}}$\pm$5e-4 & \underline{\textbf{0.851}}$\pm$3e-4 & \underline{\textbf{0.833}}$\pm$2e-4 & \underline{\textbf{0.607}}$\pm$8e-4 & \underline{\textbf{0.739}}$\pm$8e-4 & \underline{\textbf{0.686}}$\pm$7e-3 & \underline{\textbf{0.701}}$\pm$5e-3 \\
		& MONet \cite{Burgess2019MONet} & 0.451$\pm$2e-4 & 0.575$\pm$2e-4 & 0.785$\pm$2e-4 & 0.755$\pm$2e-4 & N/A & N/A & \underline{\textit{0.535}}$\pm$2e-3 & 0.473$\pm$1e-3 \\
		& GENESIS \cite{Engelcke2020GENESIS} & 0.458$\pm$4e-4 & 0.628$\pm$3e-4 & 0.536$\pm$8e-4 & 0.431$\pm$1e-3 & 0.076$\pm$1e-4 & 0.135$\pm$2e-4 & 0.298$\pm$8e-3 & 0.464$\pm$3e-3 \\
		& SPACE \cite{Lin2020SPACE} & 0.429$\pm$1e-3 & 0.519$\pm$1e-3 & 0.684$\pm$1e-3 & 0.599$\pm$1e-3 & \underline{\textit{0.374}}$\pm$3e-4 & \underline{\textit{0.505}}$\pm$3e-4 & 0.311$\pm$7e-3 & \underline{\textit{0.551}}$\pm$9e-3 \\
		& Slot Attention \cite{Locatello2020Object} & 0.493$\pm$1e-3 & 0.630$\pm$1e-3 & 0.605$\pm$1e-3 & 0.467$\pm$2e-3 & N/A & N/A & 0.039$\pm$4e-3 & N/A \\
		& EfficientMORL \cite{Emami2021Efficient} & 0.156$\pm$3e-4 & 0.011$\pm$2e-4 & 0.551$\pm$7e-4 & 0.423$\pm$1e-3 & N/A & N/A & 0.001$\pm$1e-3 & N/A \\
		& GENESIS-V2 \cite{Engelcke2021GENESIS} & 0.205$\pm$1e-4 & 0.122$\pm$5e-5 & 0.509$\pm$2e-4 & 0.393$\pm$3e-4 & N/A & N/A & 0.210$\pm$1e-2 & 0.529$\pm$9e-3 \\
		\midrule
		\multirow{10}{*}{dSprites} & AIR \cite{Eslami2016Attend} & 0.758$\pm$3e-4 & 0.746$\pm$4e-4 & \underline{\textit{0.880}}$\pm$3e-4 & 0.832$\pm$4e-4 & N/A & N/A & 0.358$\pm$6e-3 & \underline{\textit{0.726}}$\pm$2e-3 \\
		& N-EM \cite{Greff2017Neural} & 0.558$\pm$1e-3 & 0.488$\pm$3e-3 & 0.749$\pm$1e-3 & 0.629$\pm$2e-3 & N/A & N/A & 0.027$\pm$3e-3 & N/A \\
		& IODINE \cite{Greff2019Multi} & \underline{\textit{0.821}}$\pm$3e-4 & \underline{\textit{0.906}}$\pm$4e-4 & 0.840$\pm$3e-4 & 0.853$\pm$6e-4 & N/A & N/A & \underline{\textit{0.602}}$\pm$7e-3 & N/A \\
		& GMIOO \cite{Yuan2019Generative} & \underline{\textbf{0.903}}$\pm$5e-4 & \underline{\textbf{0.958}}$\pm$3e-4 & \underline{\textbf{0.931}}$\pm$6e-4 & \underline{\textbf{0.929}}$\pm$1e-3 & \underline{\textbf{0.794}}$\pm$2e-3 & \underline{\textbf{0.857}}$\pm$2e-3 & \underline{\textbf{0.632}}$\pm$9e-3 & \underline{\textbf{0.879}}$\pm$3e-3 \\
		& MONet \cite{Burgess2019MONet} & 0.776$\pm$2e-4 & 0.883$\pm$2e-4 & 0.867$\pm$3e-4 & \underline{\textit{0.858}}$\pm$3e-4 & N/A & N/A & 0.592$\pm$5e-3 & 0.674$\pm$3e-4 \\
		& GENESIS \cite{Engelcke2020GENESIS} & 0.748$\pm$5e-4 & 0.881$\pm$4e-4 & 0.712$\pm$6e-4 & 0.680$\pm$1e-3 & 0.067$\pm$2e-4 & 0.121$\pm$3e-4 & 0.290$\pm$8e-3 & 0.420$\pm$2e-3 \\
		& SPACE \cite{Lin2020SPACE} & 0.813$\pm$5e-4 & \underline{\textit{0.906}}$\pm$6e-4 & 0.835$\pm$7e-4 & 0.798$\pm$1e-3 & \underline{\textit{0.599}}$\pm$9e-4 & \underline{\textit{0.694}}$\pm$8e-4 & 0.289$\pm$8e-3 & 0.619$\pm$7e-3 \\
		& Slot Attention \cite{Locatello2020Object} & 0.362$\pm$4e-4 & 0.196$\pm$2e-4 & 0.749$\pm$6e-4 & 0.705$\pm$9e-4 & N/A & N/A & 0.000$\pm$8e-4 & N/A \\
		& EfficientMORL \cite{Emami2021Efficient} & 0.197$\pm$5e-4 & 0.059$\pm$9e-4 & 0.488$\pm$1e-3 & 0.353$\pm$2e-3 & N/A & N/A & 0.011$\pm$2e-3 & N/A \\
		& GENESIS-V2 \cite{Engelcke2021GENESIS} & 0.616$\pm$2e-3 & 0.646$\pm$3e-3 & 0.826$\pm$6e-4 & 0.800$\pm$9e-4 & N/A & N/A & 0.295$\pm$2e-2 & 0.580$\pm$3e-3 \\
		\midrule
		\multirow{10}{*}{AbsScene} & AIR \cite{Eslami2016Attend} & 0.300$\pm$3e-4 & 0.294$\pm$3e-4 & 0.608$\pm$5e-4 & 0.478$\pm$6e-4 & N/A & N/A & 0.025$\pm$8e-3 & 0.624$\pm$3e-3 \\
		& N-EM \cite{Greff2017Neural} & 0.144$\pm$4e-4 & 0.208$\pm$2e-3 & 0.174$\pm$1e-3 & 0.116$\pm$1e-3 & N/A & N/A & 0.018$\pm$3e-3 & N/A \\
		& IODINE \cite{Greff2019Multi} & \underline{\textbf{0.801}}$\pm$4e-4 & \underline{\textbf{0.857}}$\pm$4e-4 & \underline{\textbf{0.931}}$\pm$6e-4 & \underline{\textbf{0.945}}$\pm$5e-4 & N/A & N/A & \underline{\textit{0.738}}$\pm$7e-3 & N/A \\
		& GMIOO \cite{Yuan2019Generative} & \underline{\textit{0.771}}$\pm$5e-4 & \underline{\textit{0.830}}$\pm$5e-4 & \underline{\textit{0.908}}$\pm$8e-4 & \underline{\textit{0.909}}$\pm$1e-3 & \underline{\textbf{0.747}}$\pm$2e-3 & \underline{\textbf{0.835}}$\pm$2e-3 & \underline{\textbf{0.782}}$\pm$8e-3 & \underline{\textbf{0.947}}$\pm$2e-3 \\
		& MONet \cite{Burgess2019MONet} & 0.620$\pm$1e-4 & 0.427$\pm$1e-4 & 0.814$\pm$1e-4 & 0.824$\pm$2e-4 & N/A & N/A & 0.493$\pm$3e-3 & 0.418$\pm$2e-3 \\
		& GENESIS \cite{Engelcke2020GENESIS} & 0.192$\pm$4e-4 & 0.048$\pm$2e-4 & 0.403$\pm$8e-4 & 0.286$\pm$4e-4 & 0.085$\pm$3e-4 & 0.153$\pm$4e-4 & 0.141$\pm$1e-2 & \underline{\textit{0.735}}$\pm$4e-3 \\
		& SPACE \cite{Lin2020SPACE} & 0.715$\pm$3e-4 & 0.794$\pm$5e-4 & 0.773$\pm$9e-4 & 0.722$\pm$1e-3 & \underline{\textit{0.539}}$\pm$2e-4 & \underline{\textit{0.623}}$\pm$2e-4 & 0.303$\pm$4e-3 & 0.707$\pm$4e-3 \\
		& Slot Attention \cite{Locatello2020Object} & 0.759$\pm$3e-3 & 0.794$\pm$5e-3 & 0.870$\pm$4e-4 & 0.855$\pm$5e-4 & N/A & N/A & 0.119$\pm$1e-2 & N/A \\
		& EfficientMORL \cite{Emami2021Efficient} & 0.459$\pm$1e-3 & 0.429$\pm$2e-3 & 0.683$\pm$1e-3 & 0.583$\pm$2e-3 & N/A & N/A & 0.303$\pm$1e-2 & N/A \\
		& GENESIS-V2 \cite{Engelcke2021GENESIS} & 0.501$\pm$1e-3 & 0.422$\pm$4e-3 & 0.766$\pm$6e-4 & 0.757$\pm$1e-3 & N/A & N/A & 0.164$\pm$1e-2 & 0.639$\pm$9e-3 \\
		\midrule
		\multirow{10}{*}{CLEVR} & AIR \cite{Eslami2016Attend} & 0.044$\pm$1e-4 & 0.034$\pm$1e-4 & 0.954$\pm$4e-4 & \underline{\textit{0.952}}$\pm$8e-4 & N/A & N/A & \underline{\textbf{0.645}}$\pm$5e-3 & 0.880$\pm$1e-3 \\
		& N-EM \cite{Greff2017Neural} & 0.128$\pm$4e-4 & 0.178$\pm$7e-4 & 0.143$\pm$7e-4 & 0.063$\pm$6e-4 & N/A & N/A & 0.013$\pm$3e-3 & N/A \\
		& IODINE \cite{Greff2019Multi} & 0.576$\pm$3e-4 & 0.478$\pm$1e-3 & 0.954$\pm$1e-3 & \underline{\textit{0.952}}$\pm$2e-3 & N/A & N/A & 0.339$\pm$1e-2 & N/A \\
		& GMIOO \cite{Yuan2019Generative} & 0.727$\pm$3e-4 & 0.741$\pm$5e-4 & \underline{\textit{0.956}}$\pm$3e-4 & 0.949$\pm$5e-4 & \underline{\textit{0.649}}$\pm$1e-3 & \underline{\textit{0.757}}$\pm$9e-4 & 0.525$\pm$6e-3 & \underline{\textbf{0.907}}$\pm$3e-3 \\
		& MONet \cite{Burgess2019MONet} & \underline{\textbf{0.850}}$\pm$3e-4 & \underline{\textbf{0.896}}$\pm$8e-4 & \underline{\textbf{0.957}}$\pm$3e-4 & \underline{\textbf{0.955}}$\pm$3e-4 & N/A & N/A & \underline{\textit{0.596}}$\pm$4e-3 & 0.737$\pm$4e-4 \\
		& GENESIS \cite{Engelcke2020GENESIS} & 0.324$\pm$6e-4 & 0.411$\pm$6e-4 & 0.230$\pm$2e-3 & 0.150$\pm$2e-3 & 0.063$\pm$1e-4 & 0.106$\pm$1e-4 & 0.221$\pm$1e-2 & 0.577$\pm$5e-3 \\
		& SPACE \cite{Lin2020SPACE} & \underline{\textit{0.781}}$\pm$2e-4 & \underline{\textit{0.822}}$\pm$3e-4 & 0.939$\pm$4e-4 & 0.931$\pm$5e-4 & \underline{\textbf{0.683}}$\pm$4e-4 & \underline{\textbf{0.773}}$\pm$5e-4 & 0.416$\pm$3e-3 & \underline{\textit{0.894}}$\pm$4e-3 \\
		& Slot Attention \cite{Locatello2020Object} & 0.378$\pm$2e-4 & 0.078$\pm$8e-5 & 0.945$\pm$5e-4 & 0.939$\pm$8e-4 & N/A & N/A & 0.011$\pm$2e-3 & N/A \\
		& EfficientMORL \cite{Emami2021Efficient} & 0.535$\pm$1e-3 & 0.494$\pm$3e-3 & 0.776$\pm$2e-3 & 0.696$\pm$3e-3 & N/A & N/A & 0.054$\pm$5e-3 & N/A \\
		& GENESIS-V2 \cite{Engelcke2021GENESIS} & 0.289$\pm$1e-3 & 0.066$\pm$1e-3 & 0.665$\pm$2e-3 & 0.551$\pm$2e-3 & N/A & N/A & 0.218$\pm$7e-3 & 0.730$\pm$8e-3 \\
		\midrule
		\multirow{10}{*}{SHOP} & AIR \cite{Eslami2016Attend} & 0.636$\pm$1e-4 & 0.563$\pm$3e-4 & 0.857$\pm$1e-4 & 0.831$\pm$4e-4 & N/A & N/A & 0.264$\pm$1e-2 & 0.650$\pm$1e-3 \\
		& N-EM \cite{Greff2017Neural} & 0.234$\pm$1e-3 & 0.302$\pm$3e-3 & 0.273$\pm$8e-4 & 0.160$\pm$1e-3 & N/A & N/A & 0.002$\pm$1e-3 & N/A \\
		& IODINE \cite{Greff2019Multi} & 0.703$\pm$2e-3 & 0.638$\pm$6e-3 & 0.846$\pm$9e-4 & 0.789$\pm$7e-4 & N/A & N/A & 0.263$\pm$8e-3 & N/A \\
		& GMIOO \cite{Yuan2019Generative} & \underline{\textbf{0.759}}$\pm$3e-4 & \underline{\textit{0.773}}$\pm$2e-4 & 0.926$\pm$3e-4 & 0.917$\pm$9e-4 & \underline{\textit{0.652}}$\pm$1e-3 & \underline{\textit{0.743}}$\pm$1e-3 & \underline{\textit{0.341}}$\pm$8e-3 & \underline{\textbf{0.876}}$\pm$1e-3 \\
		& MONet \cite{Burgess2019MONet} & \underline{\textit{0.752}}$\pm$1e-4 & \underline{\textbf{0.796}}$\pm$2e-4 & 0.900$\pm$1e-4 & 0.885$\pm$2e-4 & N/A & N/A & \underline{\textbf{0.375}}$\pm$2e-3 & \underline{\textit{0.809}}$\pm$1e-3 \\
		& GENESIS \cite{Engelcke2020GENESIS} & 0.326$\pm$2e-4 & 0.087$\pm$2e-4 & 0.360$\pm$4e-4 & 0.188$\pm$3e-4 & 0.105$\pm$1e-4 & 0.171$\pm$2e-4 & 0.000$\pm$0e-0 & 0.681$\pm$3e-3 \\
		& SPACE \cite{Lin2020SPACE} & 0.736$\pm$2e-4 & 0.771$\pm$2e-4 & 0.874$\pm$4e-4 & 0.857$\pm$9e-4 & \underline{\textbf{0.663}}$\pm$6e-4 & \underline{\textbf{0.767}}$\pm$7e-4 & 0.269$\pm$6e-3 & 0.669$\pm$9e-3 \\
		& Slot Attention \cite{Locatello2020Object} & 0.487$\pm$1e-3 & 0.194$\pm$3e-3 & \underline{\textbf{0.942}}$\pm$6e-4 & \underline{\textbf{0.950}}$\pm$4e-4 & N/A & N/A & 0.000$\pm$0e-0 & N/A \\
		& EfficientMORL \cite{Emami2021Efficient} & 0.545$\pm$6e-4 & 0.292$\pm$1e-3 & \underline{\textit{0.934}}$\pm$8e-4 & \underline{\textit{0.941}}$\pm$1e-3 & N/A & N/A & 0.042$\pm$5e-3 & N/A \\
		& GENESIS-V2 \cite{Engelcke2021GENESIS} & 0.426$\pm$3e-4 & 0.119$\pm$6e-4 & 0.900$\pm$9e-4 & 0.903$\pm$1e-3 & N/A & N/A & 0.003$\pm$1e-3 & 0.613$\pm$1e-2 \\
		\midrule
		\multirow{10}{*}{GSO} & AIR \cite{Eslami2016Attend} & 0.196$\pm$1e-4 & 0.200$\pm$1e-4 & 0.685$\pm$5e-4 & \underline{\textit{0.564}}$\pm$9e-4 & N/A & N/A & 0.142$\pm$7e-3 & \underline{\textbf{0.718}}$\pm$4e-3 \\
		& N-EM \cite{Greff2017Neural} & 0.153$\pm$1e-3 & 0.162$\pm$3e-3 & 0.232$\pm$5e-4 & 0.140$\pm$9e-4 & N/A & N/A & 0.034$\pm$3e-3 & N/A \\
		& IODINE \cite{Greff2019Multi} & \underline{\textit{0.379}}$\pm$7e-4 & 0.400$\pm$3e-3 & 0.466$\pm$1e-3 & 0.285$\pm$2e-3 & N/A & N/A & 0.017$\pm$2e-3 & N/A \\
		& GMIOO \cite{Yuan2019Generative} & \underline{\textbf{0.572}}$\pm$3e-4 & \underline{\textbf{0.621}}$\pm$2e-4 & \underline{\textbf{0.774}}$\pm$7e-4 & \underline{\textbf{0.670}}$\pm$2e-3 & \underline{\textbf{0.431}}$\pm$1e-3 & \underline{\textit{0.547}}$\pm$1e-3 & \underline{\textit{0.240}}$\pm$6e-3 & 0.626$\pm$9e-3 \\
		& MONet \cite{Burgess2019MONet} & 0.360$\pm$3e-4 & \underline{\textit{0.414}}$\pm$4e-4 & 0.599$\pm$1e-4 & 0.432$\pm$3e-4 & N/A & N/A & 0.083$\pm$6e-3 & 0.601$\pm$5e-3 \\
		& GENESIS \cite{Engelcke2020GENESIS} & 0.233$\pm$7e-6 & 0.213$\pm$5e-6 & 0.250$\pm$1e-5 & 0.156$\pm$2e-5 & 0.062$\pm$2e-5 & 0.104$\pm$3e-5 & 0.008$\pm$1e-3 & \underline{\textit{0.630}}$\pm$4e-4 \\
		& SPACE \cite{Lin2020SPACE} & 0.294$\pm$7e-5 & 0.023$\pm$4e-5 & \underline{\textit{0.705}}$\pm$2e-4 & 0.398$\pm$2e-4 & \underline{\textit{0.400}}$\pm$2e-4 & \underline{\textbf{0.562}}$\pm$2e-4 & 0.000$\pm$0e-0 & 0.463$\pm$2e-2 \\
		& Slot Attention \cite{Locatello2020Object} & 0.205$\pm$6e-5 & 0.089$\pm$2e-4 & 0.455$\pm$2e-4 & 0.257$\pm$6e-4 & N/A & N/A & 0.004$\pm$2e-3 & N/A \\
		& EfficientMORL \cite{Emami2021Efficient} & 0.302$\pm$8e-4 & 0.130$\pm$9e-4 & 0.538$\pm$2e-3 & 0.375$\pm$3e-3 & N/A & N/A & 0.100$\pm$9e-3 & N/A \\
		& GENESIS-V2 \cite{Engelcke2021GENESIS} & 0.232$\pm$1e-4 & 0.034$\pm$1e-4 & 0.672$\pm$2e-4 & 0.525$\pm$8e-4 & N/A & N/A & \underline{\textbf{0.242}}$\pm$8e-3 & 0.614$\pm$7e-3 \\
		\bottomrule
	\end{tabular}
\end{table*}